\documentclass{article}

\usepackage{arxiv}

\usepackage[utf8]{inputenc} 
\usepackage[T1]{fontenc}    
\usepackage{hyperref}       
\usepackage{url}            
\usepackage{booktabs}       
\usepackage{amsfonts}       
\usepackage{nicefrac}       
\usepackage{microtype}      
\usepackage{lipsum}
\usepackage[T1]{fontenc}
\usepackage{array}
\usepackage{multirow}
\usepackage{amsthm}
\usepackage{amsmath}
\usepackage{amssymb}
\usepackage{graphicx}
\usepackage{epstopdf}
\usepackage{subfig}
\usepackage[table]{xcolor}
\newtheorem{thm}{Theorem}

\title{PAC: A Novel Self-Adaptive Neuro-Fuzzy Controller for Micro Aerial Vehicles
\thanks{This paper has been accepted for publication in Information Science Journal 2019. The source code is available in \url{https://www.researchgate.net/publication/336241858_Code_hexacopter_PAC_latest}.}}

\author{
  Md Meftahul Ferdaus \\
  School of Engineering and Information Technology\\
  UNSW, Canberra, ACT\\
  Australia, 2612 \\
  \texttt{m.ferdaus@student.unsw.edu.au} \\
  \And
  Mahardhika Pratama \\
  School of Computer Science and Engineering\\
  Nanyang Technological University\\
  Singapore, 639798 \\
  \texttt{mpratama@ntu.edu.sg} \\
   \And
  Sreenatha G. Anavatti \\
 School of Engineering and Information Technology\\
  UNSW, Canberra, ACT\\
  Australia, 2612 \\
  \texttt{s.anavatti@adfa.edu.au} \\
   \And
  Matthew A. Garratt \\
 School of Engineering and Information Technology\\
  UNSW, Canberra, ACT\\
  Australia, 2612 \\
  \texttt{M.Garratt@adfa.edu.au} \\
  \And
  Edwin Lughofer \\
 Department of Knowledge Based Mathematical Systems\\
  Johannes Kepler University, Linz, Austria\\
  \texttt{edwin.lughofer@jku.at} \\
}

\begin{document}
\maketitle

\begin{abstract}
There exists an increasing demand for a flexible and computationally efficient controller for micro aerial vehicles (MAVs) due to a high degree of environmental perturbations. In this work, an evolving neuro-fuzzy controller, namely Parsimonious Controller (PAC) is proposed. It features fewer network parameters than conventional approaches due to the absence of rule premise parameters. PAC is built upon a recently developed evolving neuro-fuzzy system known as parsimonious learning machine (PALM) and adopts new rule growing and pruning modules derived from the approximation of bias and variance. These rule adaptation methods have no reliance on user-defined thresholds, thereby increasing the PAC's autonomy for real-time deployment. PAC adapts the consequent parameters with the sliding mode control (SMC) theory in the single-pass fashion. The boundedness and convergence of the closed-loop control system's tracking error and the controller's consequent parameters are confirmed by utilizing the LaSalle-Yoshizawa theorem. Lastly, the controller's efficacy is evaluated by observing various trajectory tracking performance from a bio-inspired flapping wing micro aerial vehicle (BI-FWMAV) and a rotary wing micro aerial vehicle called hexacopter. Furthermore, it is compared to three distinctive controllers. Our PAC outperforms the linear PID controller and feed-forward neural network (FFNN) based nonlinear adaptive controller. Compared to its predecessor, G-controller, the tracking accuracy is comparable, but the PAC incurs significantly fewer parameters to attain similar or better performance than the G-controller.   
\end{abstract}

\keywords{Micro aerial vehicle \and Neuro-fuzzy \and Parsimonious learning machine \and Self-adaptive}

\section{Introduction}\label{Intro}
Advancements in portable electronic technology over the past few years encourage researchers to work on Micro Aerial Vehicles (MAVs). Attempts to humanize these MAVs, such as with the capacity of autonomous flying in a confined space is challenging. A positive step to such achievement is the low cost of miniature electronic components like sensors, actuators, microprocessors, batteries, etc. Another point is their immense applicability in both civilian and military sectors. Usage of MAVs in the military sector has intensified in the last decades. Participation of MAVs in civilian applications also has a lot of socio-economic benefits. To transmute all these potentialities of MAVs into reality, a major concern is to pursue the preferable control autonomy.

First principle techniques (FPTs) are prevalent in stabilizing and controlling MAVs, but, within a specific range around the equilibrium. Among numerous FPTs, linear controllers like Proportional Integral Derivative (PID), Linear Quadratic (LQ) methods are employed successfully in different MAVs due to straightforward structure. Nonetheless, linear controllers performance degrade abruptly to deal with environmental disruption due to MAV's inherent non-linearity and coupled dynamics. In rejecting disturbance of nonlinear MAV dynamics, nonlinear control techniques namely Backstepping, Sliding Mode techniques, Feedback Linearization (FBL), H$\infty$ perform diligently. A shortfall of all these linear and nonlinear controllers is their foremost dependency on the precise plant dynamics to be controlled. In MAVs, noisy data may obtain from various sensors and microelectronic components. Besides, encountering environmental disruptions like wind gust in open space, actuator degradation is usual in MAVs. Integration of such unexpected facts in MAV's dynamics is laborious or inconceivable. These deficiencies of FPT based controllers tempts research towards mathematical model-free methods.

Among numerous model-free approaches, the Fuzzy Logic systems (FLS), and Neural Networks (NN) are the most commonly used methods in various control applications \cite{ferdaus2018towards}, since they can effectively deal with complex nonlinear and chaotic systems. When superiority's of both FLS and NN based controllers are joined into a single structure, they are known as Neuro-Fuzzy (NF) controllers, which are also employed in diverse engineering industries. However, development of a NN, FLS or NF-based high-performance controllers is tough. Among numerous existing methods, a simple way to construct NN, FLS or NF-based controller is to train them offline using the known knowledge about the plant dynamics. Furthermore, it can be executed by mimicking an FPT controller like PID through offline training. These trained controllers possess several deficiencies. Firstly, they consist of a fixed architecture with an exact number of neurons or membership functions, rules or layers. With their static configuration, they can not adapt with time-varying non-stationary plants. Secondly, the controllers have a dependency on experts knowledge or training data. Therefore, control performance relies on the accuracy of that knowledge or data. To summarize, these static NN, FLS or NF-based controllers do not have any direct association with the plant models; rather, there exists an indirect reliance. To mitigate limitations of these static controllers, researchers have tried to utilize an adaptive neuro-fuzzy inference system (ANFIS) in various control applications \cite{melin2007hybrid,melin2003adaptive,melin2007intelligent,melin2004new}. To improve the performance of the type-1 FLS, or type-1 ANFIS-based controllers, and to handle uncertainties, research has also been observed to develop type-2 FLS-based controllers \cite{ontiveros2018comparative,castillo2016generalized,castillo2016comparative,cervantes2015type,sanchez2015generalized}. 

To improve the performance of model-free static controllers, or ANFIS-based controllers, whether using type-1 or type-2 mechanism, researchers have developed adaptive controllers by combining conventional nonlinear control techniques such as backstepping, sliding mode techniques \cite{chang2018sliding}, feedback linearization (FBL), H$\infty$, etc. with FLS, NN, or NF structure, which can tune their network parameters to alleviate the adverse influence of uncertainties and disturbances. In these control schemes, only the network parameters are altered, where they are maintaining a fixed structure. It enforces us to specify the number of nodes, layers, or rules beforehand. It is hard to know the exact number of rules a priori to attain the desired control performance. Controllers with only a few rules may fail to produce the desired performance, whereas too many of them may originate an over-complex structure of controllers to actualize in real-time. To circumvent such shortcomings, NN or NF controllers with flexible architecture can be employed. These flexible controllers are not only able to tune parameters, but also to evolve the structure by adding or deleting layers or rules in self-adaptive fashion \cite{angelov2004fuzzy}. In addition, the self-adaptive trait is capable of adapting to changing  plant dynamics without re-tuning or retraining phase from scratch. 

Research on evolving intelligent controllers has started at the beginning of the twenty-first century. In 2003, a self-organizing NF controller was proposed in \cite{gao2003online}, where they applied system error, $\epsilon-$ completeness, and error reduction ratio in their rule evolution scheme. Their controller needed to store all preceding input-output data, which forced to compute a large matrix in each step and yields a high computation cost. It makes them impractical to implement in systems like MAVs, where a fast response is expected from the controller to emulate the desired commands. Another evolving controller was developed in \cite{angelov2004fuzzy} by utilizing the evolving Takagi Sugeno (eTS) model \cite{angelov2004approach}. Though their controller was evolving in nature, it suffers from several imperfections. First, in their structural evolution mechanism, fuzzy rules were added or replaced only, pruning of rules was missing. Besides, their eTS fuzzy controller needed to memorize the previous data obtained from both the plant and controller. Such limitation led to a computationally costly control mechanism and made them unrealistic in swift reaction-based control application. A hybrid evolving controller was developed in \cite{de2007evolving} by mixing an evolving and a static TS-fuzzy system. Despite their design simplicity, they required to know some parameters of the plant to be controlled. Such parameters may not be available during control operation. An evolving controller was also actualized using model predictive control technique \cite{han2013real}. However, their dependency on plant's dynamic model restricted their application in complex nonlinear systems, where dynamic models may not be known.

Above mentioned evolving NN, FLS or NF-based controller are equipped with hyper-sphere-shaped clustering techniques, where mostly univariate Gaussian membership functions are used. As a consequence, those evolving controllers are associated with several rule premise parameters like mean and width. These parameters need to be updated continuously, which rises computational complexities. Recently, hyper-ellipsoid-shaped clustering-based evolving NF controller namely G-controller was developed by \cite{ferdaus2019generic}. Multivariate Gaussian function was used due to its scale-invariant behavior and ability to handle both axis parallel and non-axis parallel data distribution. Nevertheless, like the hyper-sphere-shaped clustering-based counterparts, the G-controller involved numerous network parameters. A bottleneck of evolving controllers is the engagement of manifold parameters, which strikes adversely in furnishing a fast response. To lessen the parameters of evolving controllers, \cite{vskrjanc2014robust} used evolving data-cloud-based clustering technique to develop an evolving TS-fuzzy controller. Though they reduced network parameters, still bound to calculate accumulated distance of a particular point to all other points. The obligation of such calculation can not effectively diminish the memory demand of their evolving controller. 

To mitigate the computational complexity of the above discussed evolving controllers, hyper-plane-shaped clustering technique-based evolving controller could be a promising avenue, since they are free from premise parameters. Lately, a hyper-plane-shaped clustering technique-based evolving NF architecture, namely parsimonious learning machine (PALM) was proposed in \cite{ferdaus2019palm}. In this work, the PALM architecture has been utilized to develop an evolving controller namely PAC. Unlike the PALM, the rule adaptation module in PAC is replaced to originate a simplified evolving controller. In addition, the self-constructing clustering and similarity analysis-based rule-evolution mechanism of PALM have a dependency on user-defined parameters. To mitigate such dependency, in PAC, the rule evolution mechanism of PALM has been swapped with the so-called network significance method, which is deduced from the idea of bias-variance trade-off \cite{pratama2018atl}. In PAC's rule evolution technique, the statistical contribution of each rule can be estimated to add or prune them on demand, which aids to achieve the desired control performance. 

\subsection{Main features}
Main features of our proposed evolving controller, namely PAC can be uttered as follows:
\begin{enumerate}
\item \textbf{\textit{Premise-free fuzzy rule-based system:~}}Usually, fuzzy logic controllers are rule-based. They consist of antecedent and consequent parts, where both parts are associated with several parameters. Unlike the conventional fuzzy logic controller, the fuzzy rules in PAC are portrayed by hyper-planes which underpins both the rule premise and consequent parts. As a result, it does not have any premise parameter. Such scheme trims the rule-based parameters to the level of $R\times(N+1)$, where $N$ denotes the number of input dimension and $R$ is the number of fuzzy rules. The number of network parameters of the proposed controller is considerably lower than the conventional evolving fuzzy controllers discussed in the literature.
\item \textbf{\textit{New rule growing and pruning mechanism based on bias and variance:~}} In PALM \cite{ferdaus2019palm}, self-constructing clustering approach was employed to generate rules, which faces computational complexities to calculate variance and covariance among different variables. Besides, similarity analysis among hyper-planes in terms of distance and orientation between hyper-planes were measured to merge rules in PALM. To eliminate such a complex calculation of growing and pruning rules, bias-variance concept-based simplified method, namely network significance \cite{pratama2018atl} is proposed in PAC. A new rule is added in the underfitting situation, while the pruning process is triggered by the overfitting condition. The key difference of this paper from \cite{pratama2018atl} lies in the estimation of bias and variance for the hyperplane-shaped clustering-based rules.
\item \textbf{\textit{New evolving fuzzy controller:~}}In general, the evolving fuzzy controllers' rule evolution methods have a reliance on a number of predefined thresholds. To eliminate such dependency on user-defined thresholds, the bias-variance concept based network significance method is exercised in PAC, where we do not need any user-defined problem-specific thresholds. Besides, the proposed controller has no premise parameters, its only consequent parameter, i.e., weights are adapted by using SMC learning theory to confirm a stable closed-loop system. To evaluate the controller's stable and precise tracking performance, it has been implemented into a simulated BI-FWMAV and hexacopter plant.
\end{enumerate}

Aforementioned features of the PAC are desiring to achieve the desired control autonomy in MAVs. Therefore, the proposed controller is applied to control a BI-FWMAV and a hexacopter plant in this work. In addition, the whole code of the PAC is written in C programming language. It is compatible with the majority of the MAVs hardware, where its implementation is made accessible publicly in \cite{PAC_Code} to assure reproducible research.         
   
\section{Related work\label{sec:Related work}}
Recent advent in portable microelectronic technology is encouraging to develop MAVs. Their autonomous flying ability in both natural and human-made environment is aspiring for numerous defense applications. Besides, they can play a vital role in civilian ventures like disaster alleviation, communication, environment conservation, etc. Until now, majority of the commercially available MAVs are controlled by FPT-based linear controllers like PIDs because of their design simplicity and easy employment. In this work, we are dealing with the control performance of a hexacopter and BI-FWMAV. In regulating rotary wing and flapping wing MAVs, implementations of PID controllers have witnessed profusely in last decades. Though PIDs can control MAVs with satisfactory accuracy in certain environment, due to their fixed gain, they are not robust against real-world uncertainties like a wind gust. Another prominent linear control method for MAVs is Linear Quadratic (LQ) technique \cite{bouabdallah2004pid}. These linear controllers suffer from attaining desired performance due to the complex nonlinearity in system dynamics. Besides, various un-modeled uncertainties and perturbations have a detrimental influence on these precise-model-based linear controllers. Afterward, to control MAVs researchers had focused on various nonlinear controllers namely back-stepping, sliding mode control \cite{busarakum2014design}, feedback linearization (FBL), H$\infty$ robust control techniques by the virtue of their disturbance rejection. However, like their linear counterparts, nonlinear controllers have a direct association with the precise dynamic model to be controlled. Such dependency limits their application where the plant dynamics are unknown. In our work, the entire dynamics of the BI-FWMAV is unidentified, where exercising the model-based nonlinear controllers is impractical. Besides, BI-FWMAV's lightweight architecture is naive to manifolds un-modeled outdoor uncertainties. Model-free control approaches are an effective solution to deal with un-modeled uncertainties. 

Model-free control approaches for MAVs have commenced with FLS \cite{bacik2015design}, NN \cite{kim1997nonlinear}, or NF-based controllers \cite{kurnaz2010adaptive}. However, the requirement of offline training, static structure, dependency on experts' knowledge confine their applicability in complex systems with the uncertain environment. Also, they experienced a spillover effect. To mitigate such shortcomings, researchers have tried to develop FLS, NN, and NF-based intelligent adaptive controllers \cite{melin2003adaptive,melin2004new,cervantes2015type}. However, these controllers are not able to alter their structure.  

The evolving intelligent controllers have the promising quality to handle complex dynamic systems with uncertainties. Nonetheless, they have not yet been employed profusely in regulating MAVs due to the complex framework and challenging execution. Apart from that, MAVs require a control method with prompt response, which sometimes difficult to acquire from evolving controllers due to their high memory demand. Lately, an evolving controller using interval type-2 (IT2) NF structure was proposed in \cite{chen2011robust} to control a quadcopter MAV. Nonetheless, their IT2NF system was functioning as an uncertainty and perturbation observer, and a PD controller was used to control the attitude and position of the quadcopter. In \cite{dong2016self}, an evolving NF controller was proposed for a simulated quadcopter MAV plant. Though their controller successfully generated and pruned fuzzy rules on the fly with a satisfied tracking accuracy, they only had considered a simplified dynamic model of the quadcopter to be controlled. One of the latest research on evolving controller in regulating MAVs attitude and altitude was reported in \cite{ferdaus2019generic}. Unlike the evolving controllers discussed in the literature, they effectually exercised a NF evolving controller in a high dimension nonlinear complex MAV plant. Besides, their controller was evaluated by incorporating uncertain wind gust effect in their MAVs plant. A comprehensive survey of evolving fuzzy system for regression, classification and control tasks is provided in \cite{EFSsurvey}. 

In evolving controllers, the adaptation of consequent parameters plays a crucial role to attain the desired accuracy. To adapt the evolving controller's rule consequent parameters, gradient-based methods are typically used \cite{lin2009self}. However, the gradient-based controllers perform well only when they were used to control plants with a slow variation in dynamics. Furthermore, gradient-based methods like dynamic back-propagation comprise of partial derivatives. Such algorithms can not guarantee fast convergence speed, particularly in complex non-convex search space. Also, there are chances to be trapped in a local minimum. Alternately, evolutionary algorithms were attempted in \cite{topalov1996fast} to tune parameters. Nevertheless, the stability of their proposed controller was not confirmed, and the fast response was not ensured. Such constraints can be handled simply by imposing SMC theory to adapt the consequent parameters as witnessed in \cite{ferdaus2019generic}. Inspired by the simplicity of the SMC theory \cite{ferdaus2019generic}, in our work, it has been employed to update the consequent parameters. Like \cite{ferdaus2019generic}, predefined static sliding parameters are replaced with learning rate based self-organizing parameters. It makes the proposed PAC a complete self-adaptive model-free controller.

Arrangement of the remaining sections of this paper is as follows: In Section \ref{sec:Problem statement}, limitations of existing evolving controllers are analyzed. Section \ref{sec:Structure of PALM_C} details the network structures of the PALM-based evolving controller PAC along with the explanation on rule generation and pruning mechanism. Challenges in formulating FW AMV or hexacopter model have asserted in Section \ref{sec:confrontations MAV dynamics}. Experimental results and performance evaluation of the proposed controller are described in Section \ref{sec:Numerical experiments}. The concluding remarks are drawn in Section \ref{sec:Conclusion}.
\begin{center}
	\begin{figure}[t]
		\begin{centering}
			\textsf{\includegraphics[scale=0.3]{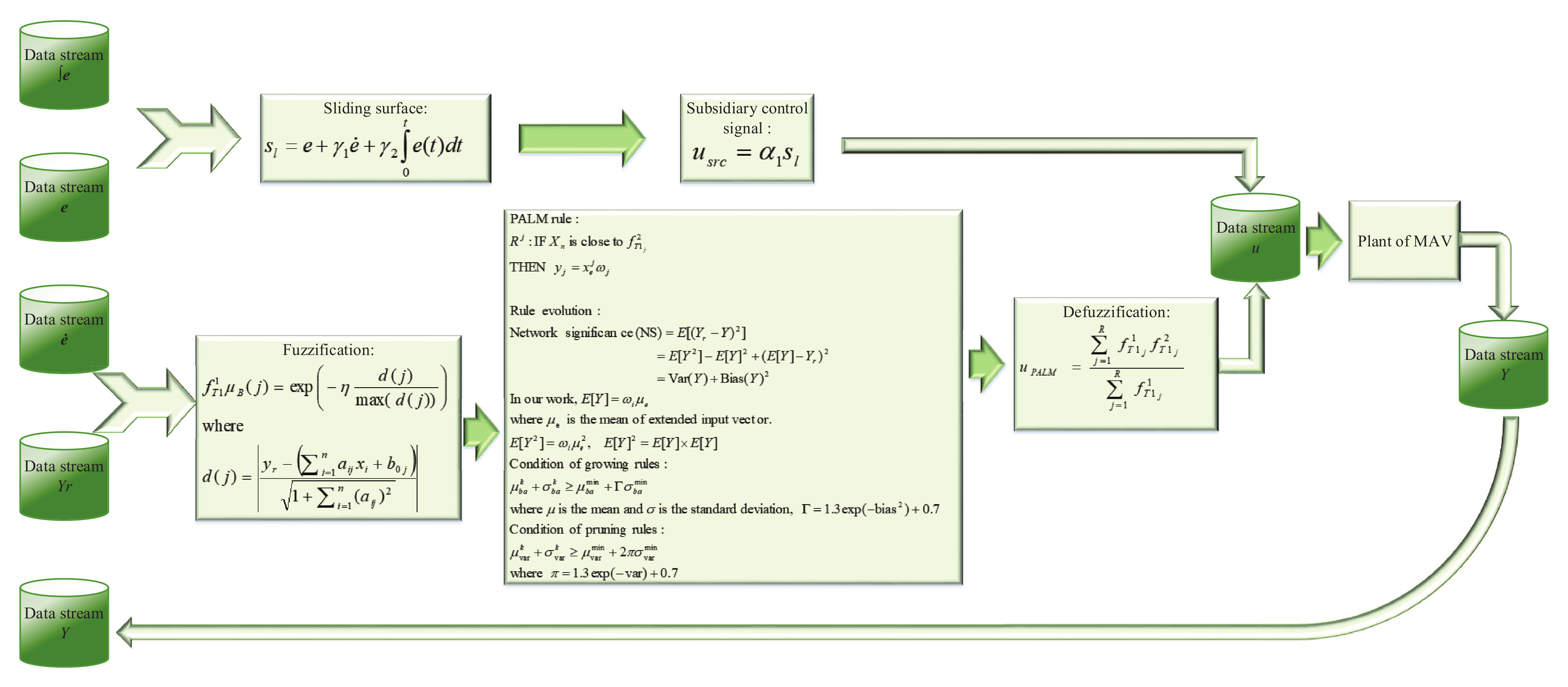}}
			\par\end{centering}
		\caption{Flow of data streams in closed-loop PAC \label{fig:flow of data streams}}
	\end{figure}
	\par\end{center}
\section{Problem statement}\label{sec:Problem statement}     
In a closed-loop control system, data comes in a sequential online manner to the controller, which may contain various uncertainties and disturbances. Such an irregularly distributed sequence of incoming data to the controller can be exposed formally as $e=\left\{ e^{1},e^{1+t},e^{1+2t}...,e^{f}\right\},$ denoted by "Data steam of $e$" in Fig. \ref{fig:flow of data streams}, where $t$ is the time step (in sec) of the closed-loop control system, $f$ is the final time until which plants need to be controlled. Here, $e$ is indicating the data stream of error, which is the difference between the reference trajectory data stream $Y_r$ and observed corresponding out data stream from the plant $Y$. The remaining incoming data streams to the closed-loop system, as displayed in Fig. \ref{fig:flow of data streams}, are data streams of derivative of error ($\dot{e}$) and the integral of error ($\int e$). To tackle irregular sequence of incoming data, the controller should hold some desiring features such as 1) able to work in single-pass mode; 2) deal with various uncertainties of the incoming data; 3) perform with low memory burden and computational complexity to enable real-time deployment under resource-constrained environment. In the realm of NF systems, such learning proficiency is manifested by evolving NF systems \cite{juang1998online}. However, enormous free parameters associated with the evolving NF controllers may cause complex computation. From the perspective of controlling MAV, a swift response from the controller is much expected, where tuning of several premise parameters in the evolving controller is a hindrance to fulfill such expectation. To decipher the involvement of profuse premise parameters in evolving controller, their premise parameter-dependent hyper-spherical or hyper-ellipsoidal clustering techniques can be substituted with premise parameter-free clustering method. From this research gap, a hyper-plane-based clustering method \cite{ferdaus2019palm} is utilized in our work, which is composed from the consequent parameters as explained in \cite{ferdaus2019palm}. Another inadequacy of state of the art evolving controllers is their affiliation to user-defined parameters to evolve their structure. Those parameters claim to alter with respect to the corresponding closed-loop system. Our proposed evolving controller is free from such parameters, and they have been superseded by the bias-variance concept. To get a clearer view, the flow of data streams in our closed-loop system is displayed in Fig. \ref{fig:flow of data streams}. Unlike the PALM \cite{ferdaus2019palm}, in the fuzzification layer of PAC, we always have incoming target/reference data stream $Y_r$ as shown in the fuzzification block of the Fig. \ref{fig:flow of data streams}. Therefore, we have not used the recurrent PALM (rPALM) architecture like \cite{ferdaus2019palm}, since it does not have any access to the target value in the testing phase. To obtain an explicit impression, the self-evolving formation of our proposed  PAC is enumerated in the next section.                                         
                 
\section{Proposed parsimonious controller\label{sec:Structure of PALM_C}}
Our proposed PAC is a three-layered NF system and functioning in tandem. It's evolving architecture is rooted with TS-fuzzy model, where classical hyper-spherical \cite{angelov2004fuzzy}, hyper-ellipsoidal \cite{pratama2014panfis}, or data-cloud based \cite{vskrjanc2014robust} clusters are substituted with hyper-plane-based clusters. Utilization of hyper-planes has removed antecedent parameters as explained in \cite{ferdaus2019palm}, which reduces the number of operative parameters in our controller dramatically. Popular hyper-plane-based clustering techniques procedures like fuzzy C-regression model (FCRM) \cite{hathaway1993switching}, fuzzy C-quadratic shell (FCQS), double FCM \cite{kim1997new}, inter type-2 fuzzy c-regression model (IT2-FCRM) \cite{zou2017ts} are non-incremental in nature, thereupon can not entertain evolving hyper-planes. Additionally, they deployed  hyper-spherical function for instance Gaussian function to accommodate hyper-planes. To mitigate such inadequacies, a new membership function \cite{zou2017ts} is used in PAC. To fetch a vivid overview of the mechanism of PALM itself, the architecture of the PALM network is detailed in the following subsection.

\subsection{Architecture of PAC}
In our PAC's closed-loop control system, the PALM network is fed by three inputs namely error ($e$), the derivative of error ($\dot{e}$) and actual plants output ($y$). Referring to the theory of fuzzy system, these crisp data ($e, \dot{e}, y$) need to be transformed into a fuzzy set, which is the initial  step in the PALM's workflow. This fuzzification process is accomplished by adopting a new membership function that can adopt the hyper-plane-shaped clusters directly, which is framed through the concept of point-to-plane distance. The employed membership function can be expressed as follows:
\begin{equation}
f_{T1}^{1}=\mu_{B}(j)=\exp\left(-\eta\frac{d(j)}{\max\left(d(j)\right)}\right)\label{eq:mu_T1}
\end{equation}
where $\eta$ is a regulating parameter which adjusts the fuzziness of membership grades. Based on the observation in \cite{zou2017ts,ferdaus2019palm}, and empirical analysis with different MAV plants in our work, the range of $\eta$ is fixed as $[1, 100]$. This membership function empowers the utilization of hyper-plane-based clusters directly into the PALM network without any rule parameters except the first-order linear function or hyperplane. In the proposed membership function, the calculation of the distance from a point to plane is not unique, where the compatibility measure is executed using the minimum point to plane distance. $d(j)$ in Eq. \eqref{eq:mu_T1} denotes the distance between the current data point and $j$th hyperplane as with Eq. \eqref{eq:dis_p_hp}. It is determined by following the definition of a point-to-plane distance \cite{pointPlane} and is formally expressed as follows \cite{ferdaus2019palm}:

\begin{equation}
d(j)=\bigg|\frac{y_{r}-(\sum_{i=1}^{n}a_{ij}x_{i}+b_{0j})}{\sqrt{1+\sum_{i=1}^{n}(a_{ij})^{2}}}\bigg|\label{eq:dis_p_hp}
\end{equation} 
where $a_{ij}$ and $b_{0j}$ are consequent parameters of the $j$th rule, $i=1,2,...,n;$ $n$ is the number of input dimension, and $y_r$ is the reference trajectory to the plant. In the closed-loop control perspective, the target value $y_r$ for the plant is always known. However, in the regression or classification problem, it is not possible to acquire the target value in the testing phase. Therefore, in \cite{ferdaus2019palm}, the target-value dependent PALM was advanced to a recurrent structure namely rPALM. Such recurrent architecture is not required in PAC since the target trajectory $y_r$ is continually available up to the end of control performance. It is noteworthy to state that a type-1 fuzzy structure is facilitated in PAC. In light of a MISO system, the IF-THEN rule of PAC can be exposed as follows:
\begin{align}
R^{j}:\quad & \text{IF}\;X_{n}\;\text{is close to}\;f_{T1_{j}}^{2}\;\text{THEN}\;y_{j}=x_{e}^{j}\omega_{j}\label{eq:fuz_R-1}
\end{align}
where $x_{e}$ is the extended input vector and is expressed by inserting the intercept to the original input vector as $x_{e}=[1,e,\dot{e},y]$, $e$ is the error, i.e. the difference between the reference and actual output of the plant, $\dot{e}$ is the error derivative, i.e. the difference between the present and previous state error value, $y_r$ is the reference trajectory for the plant to be controlled, $\omega_{j}$ is the weight vector for the $j$th rule, $y_{j}$ is the consequent part of the $j$th rule. The antecedent part of PAC is simply the hyperplane and does not consist of any premise parameters. From Eq. \eqref{eq:fuz_R-1}, it is observed that the drawback of PAC lies in the high-level fuzzy inference scheme which lowers the transparency of fuzzy rule. The intercept of extended input vector dominates the slope of hyperplanes which eliminates the untypical gradient dilemma.

In PAC, an analogous consequent part alike the basic TS-fuzzy model's rule consequent part $(y_{j}=b_{0j}+a_{1j}x_{1}+...+a_{nj}x_{n})$ is employed. The consequent part for the $j$th hyperplane is calculated by weighting the extended input variable $(x_{e})$ with its corresponding weight vector as follows:
\begin{equation}
f_{T1_{j}}^{2}=x_{e}^{T}\omega_{j}\label{eq:consequent}
\end{equation}
The weight vector in Eq. \eqref{eq:consequent} is updated recursively by the SMC theory-based adaptation laws, which ensures a smooth alteration in the weight value. In the next step, the rule firing strength is normalized and added with the rule consequent to produce the end-output of PALM. The final defuzzified crisp output of the PALM can be expressed as follows:
\begin{equation}
u_{PALM}=\frac{\sum_{j=1}^{R}f_{T1_{j}}^{1}f_{T1_{j}}^{2}}{\sum_{i=1}^{R}f_{T1_{i}}^{1}}\label{eq:y_l}
\end{equation}
The normalization term in Eq. \eqref{eq:y_l} assures the partition of unity where the sum of normalized membership degree is one. The above-described PALM network-based PAC has a similar structure like \cite{ferdaus2019palm}. Nonetheless, a simplified learning mechanism is used in PAC than the original PALM, as explained in the next subsection.
    
\subsection{Automatic constructive mechanism of PAC}
In PALM \cite{ferdaus2019palm}, the self-constructive clustering technique was adopted to grow the rules. The rule-significance was determined by measuring input and output coherence, where the coherence was calculated by investigating the correlation between the existing data samples and the target concept. Again, the computation of correlation has a dependency on finding variance and covariance among different variables. In addition, the PALM's rule growing method was regulated by two predefined thresholds. PALM's rules were merged by measuring the similarity between the hyperplane-shaped fuzzy rules. The similarity among rules was measured by observing the angle and minimum distance among them. The merging strategy was also controlled by predefined thresholds. This clearly shows the high computational cost of PALMs' rule evolving mechanism and made them incompatible in fast response-based control applications. To subjugate such complexity, a simplified rule evolution technique is implemented in PAC bottomed by the network significance method, which is formulated from the concept of bias-variance \cite{pratama2018atl}. Here, no predefined thresholds are required to regulate the generation or pruning of the fuzzy rules. This network significance method based rule growing and deletion modules of PAC are clarified in subsequent paragraphs of this section.

\subsubsection{New Mechanism of rule generation in PAC}
The strength of PAC can be analyzed by considering the tracking error, which can be written in terms of mean square error (MSE) as follows:
\begin{equation}
	MSE=\sum_{t=1}^T \frac{1}{T}\big(y_r(t)-y(t)\big)^2
\end{equation} 
where $y_r(t)$ is expressing the desired trajectory and $y(t)$ is attained output from the plant to be controlled. Such formulation experiences two obstructions in the learning mechanism of an evolving controller, such as 1) it needs to memorize all data points to get a clearer view about the controller's constructive mechanism; 2) though the recursive calculation of $MSE$ by excluding preceding data is possible, it does not investigate the strength of reconstruction for uncertain upcoming data. In simple words, it does not consider the generalization capacity of evolving controllers. To mitigate such hindrance, let us consider that $Y(t)$ is the observed plant's output for the reference input $Y_r(t)$, and $E[Y(t)]$ is the plant's expected output. According to the definition of expectation, if $x$ is continuous, then the expectation of $f(x)$ can be formulated as $E[f(x)]=\int_{-\infty}^\infty f(x)p(x)dx$, where $p(x)$ is the probability density function of $x$. Then the network significance (NS) method can be defined as follows:
\begin{equation}
NS=\int_{-\infty}^\infty \big(Y(t)_r-Y(t)\big)^2p(t)dt
\end{equation}
where $p(t)$ is denoting the probability density function of $t$. In effect, the NS method is the expectation of squared tracking error, and verbalized as follows:
\begin{equation}
NS=E\big[(Y_r(t)-Y(t))^2\big]=E\big[(Y(t)-E[Y(t)]+E[Y(t)]-Y_r(t))^2\big]
\end{equation}    
After several mathematical operations, $NS$ can be formulated with the concept of bias and variance as follows:
\begin{equation}\label{eq:NS_var_bias}
NS=E\big[(Y(t)-E[Y(t)])^2\big]+\big(E[Y(t)]-Y_r(t)\big)^2=\text{Var}(Y(t))+\text{Bias}(Y(t))^2
\end{equation}
In Eq. \eqref{eq:NS_var_bias}, the variance of $Y(t)$ i.e., $\text{Var}(Y(t))$ can be presented as follows:
\begin{equation}
\begin{aligned}
   \text{Var}(Y(t))=E\big[(Y(t)-E[Y(t)])^2\big]&=\int_{-\infty}^\infty \big(Y(t)-E(Y(t))\big)^2 p(t)dt\\
   &=E[Y(t)^2]-E[Y(t)]^2 
   \end{aligned}
\end{equation}

In establishing the NS in PAC, a rule firing strength of unity is assumed. It simplifies the network outcome as $Y(t)=\sum_{i=1}^R x_e \omega_i$, because the expectation of the normalized firing strength of PAC does not have a unique solution. Here $x_e$ is the extended input and can be exposed as $x_e=[1, x_1, x_2,...,x_n]$, where $n$ is the number of inputs to the network; $\omega_i$ is the weight vector of the $i$th rule. Now the expectation of $Y$ can be formulated as $E[Y(t)]=\int_{-\infty}^\infty x_e \omega_i p(t)dt$. Suppose that the normal distribution applies, it leads the expectation to the following expression.           
\begin{equation}\label{eq:N43S_var_bias}
E[Y(t)]= \omega_i \mu_e
\end{equation} 
where $\mu_e$ is mean of the extended input $x_e$. Factually, the integration of $x_e$ over $-\infty$ to $\infty$ originates the mean $\mu_e$. 

Let us call back the formula of variance $\text{Var}(Y(t))=E[Y(t)^2]-E[Y(t)]^2$, where the second term of the right-hand side is simply $E[Y(t)]^2=E[y(t)]\times E[y(t)]$, and the first term can be formulated as $E[Y(t)^2]= \omega_i \mu_e^2$. Incorporation of all these results in NS formula of Eq. \eqref{eq:NS_var_bias} settles the final manifestation of NS. Since the NS consists of both bias and variance, a high value of NS may indicate a high variance (over-complex network with profuse fuzzy rules) or a high bias (oversimplified network) problem. Such phenomenon can not be elucidated simply by system error index. Augmentation of a new rule is inferred to subjugate the high bias dilemma. Nonetheless, such phenomenon is not convenient for high variance context since the addition of rules magnifies the network complexity. To retain a compact network structure with satisfactory tracking performance, the concept of bias-variance trade-off is inserted in our work to calculate the NS \cite{pratama2018autonomous,pratama2018atl}. Therefore, the rule-evolution of our controller has no reliance on user-defined parameters. By confirming the fundamental objective of rule growing procedure to ease the high bias dilemma, the condition of growing rules in our work is expressed as follows:
\begin{equation}\label{eq:add_R}
\mu_{ba}^k+\sigma_{ba}^k\ge \mu_{ba}^{min}+\varGamma \sigma_{ba}^{min}
\end{equation}               
where $\mu_{ba}^k$ is denoting the mean of bias and $\sigma_{ba}^k$ is the standard deviation of bias at the $k$th observation while $\mu_{ba}^{min}$ and $\sigma_{ba}^{min}$ are pointing the minimum of mean and standard deviation up to $k$th time instant. In computing these variables, no preceding data are required. Their values are being updated directly based on the availability of upcoming signals to the PAC. When Eq. \eqref{eq:add_R} is satisfied, the values of $\mu_{ba}^{min}$ and $\sigma_{ba}^{min}$ are to be reset. To perceive an improved tracking performance from the commencing of PAC's control operation, a rapid decay in the bias value is expected. It is retained in formulating the settings of bias in Eq. \eqref{eq:add_R} as long as the plant does not encounter any uncertainties or disturbances. The presence of any disruptions in control system will elevate value of bias, which cannot be addressed directly by adapting the consequent parameter of the PAC. To elucidate such hindrance, Eq. \eqref{eq:add_R} is originated from the adaptive sigma rule, where $\varGamma$ controls the degree of confidence of the sigma rule. In our work, the $\varGamma$ is expressed as $\varGamma=1.3\exp(-\text{bias}^2)+0.7$, which revolves $\varGamma$ between 1 and 2. Consequently, it obtains the level of confidence from around 68\% to 96\%. Such scheme enhances the flexibility in the rule-growing module to adapt to environmental perturbations. It also eliminates the dependency of evolving controller on user-defined problem-dependent parameters. To sum up, a high bias usually signifies an oversimplified network, which is solved by adding rules. However, it is avoided in case of low bias since it may magnify the variance. (\ref{eq:add_R}) follows the statistical process control approach \cite{gama2006decision,gama2014survey} where the modification here exists in the use of dynamic $\varGamma$.

\subsubsection{New rule pruning module of PAC}
The structure of PAC may become highly complex as a result of high variance. On that ground, control of variance is essential to reduce the network complexity by pruning the fuzzy rules. Since a high variance indicates the overfitting condition, the rule pruning scheme initiates from the evaluation of variance. Like the rule growing mechanism of PAC, a statistical process control technique \cite{gama2014survey}  with dynamic confidence level is embraced in rule pruning module to trace the high variance dilemma as follows:
\begin{equation}\label{eq:prune_R}
\mu_{var}^k+\sigma_{var}^k\ge \mu_{var}^{min}+2\pi \sigma_{var}^{min}
\end{equation} 
where $\mu_{var}^k$ is denoting mean and $\sigma_{var}^k$ is the standard deviation of variance at the $k$th observation while $\mu_{var}^{min}$ and $\sigma_{var}^{min}$ are pointing the minimum of mean and standard deviation up to $k$th time instant. Here, the term $\pi$ is adopted as $\pi=1.3\exp(-\text{var})+0.7$, $\pi$ is a dynamic constant and regulating the degree of confidence in the sigma rule. The term 2 is Eq. \eqref{eq:prune_R} halts the direct pruning after growing. Furthermore, $\mu_{var}^{min}$ and $\sigma_{var}^{min}$ are reset when the condition in Eq. \eqref{eq:prune_R} is fulfilled. 

After the execution of Eq. \eqref{eq:prune_R}, the significance of each rule is examined via the idea of network significance, and inconsequential rules are pruned to reduce the overfitting condition. The significance of the $i$th rule is determined by its average activation degree for all possible incoming data samples or its expected values as expressed in Eq. \eqref{eq:NS_var_bias}.
Considering the normal distribution assumption, the importance of $i$th rule can be expressed as $HS_i=\omega_i \mu_e$. A small value of $HS_i$ indicates that the $i$th rule plays an small role to recover the clean input attributes. Therefore, it can be pruned with a very insignificant loss of tracking accuracy. Since the contribution of the $i$th rule is calculated in terms of the expectation $E(Y)$, the least contributing rule having the lowest $HS$ is regarded inactive. When the overfitting condition occurs or Eq. \eqref{eq:prune_R} is satisfied; the rule with the lowest $HS$ is pruned and can be expressed as follows:
\begin{equation}\label{eq:pruning_lowest}
\text{Pruning}\longrightarrow \underset{i=1,...R}{min} HS_i
\end{equation}
The condition in Eq. \eqref{eq:pruning_lowest} targets to mitigate the overfitting situation by deleting the least significant rule. It also indicates that the desired trajectory tracking performance can still be achieved with the rest $R-1$ rules. Furthermore, this rule pruning strategy enhances the generalization power of PAC by reducing its variance, which helps to deal with a variety of disturbances.              
\begin{center}
	\begin{figure}[t]
		\begin{centering}
			\textsf{\includegraphics[scale=0.4]{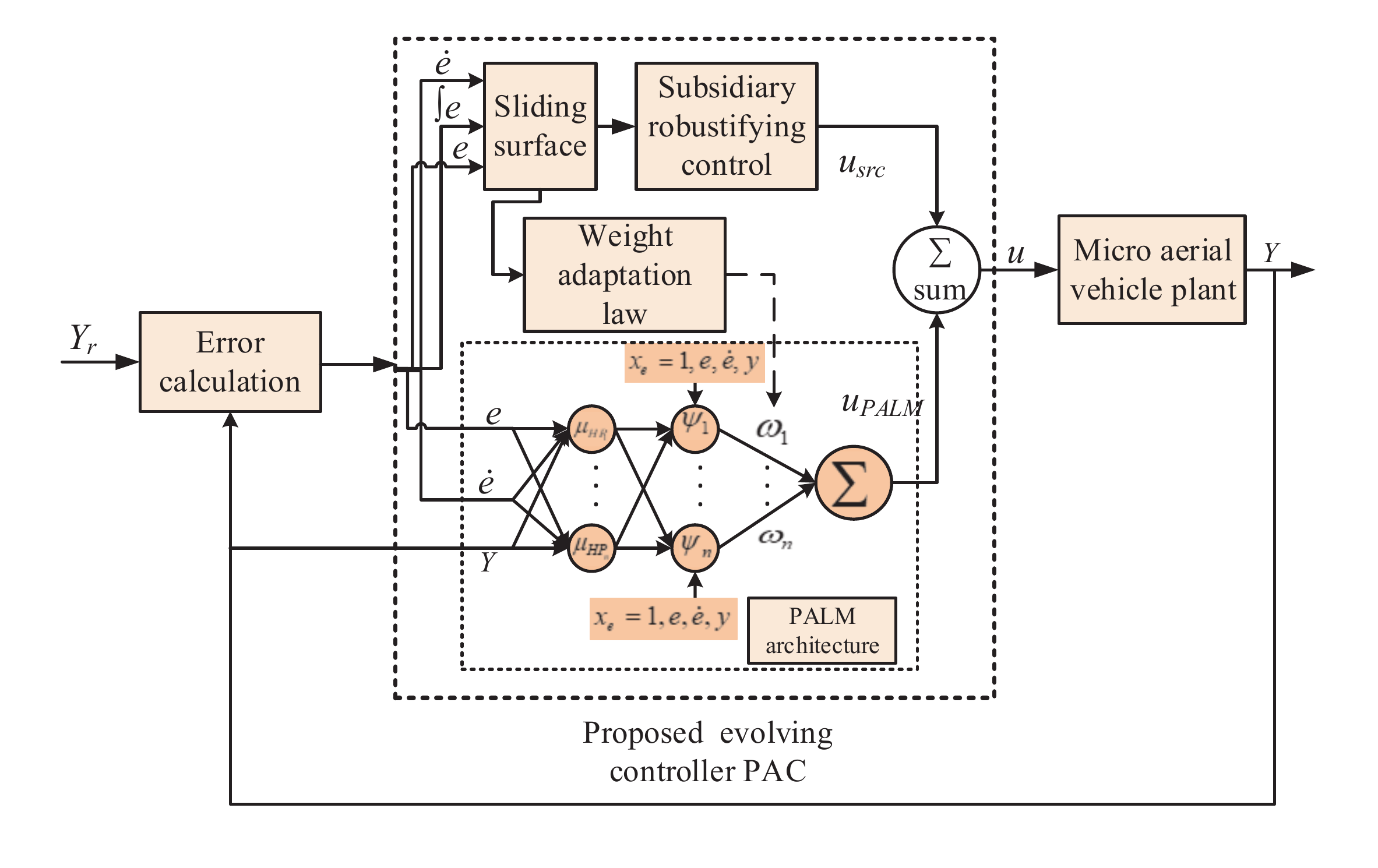}}
			\par\end{centering}
		\caption{Self-adaptive PAC's closed-loop mechanism\label{fig:g-controller-closed-loop}}
	\end{figure}
	\par\end{center}

\subsection{Adaptation of weights in PAC\label{sec:weight-Adaptation} }
Unlike the conventional evolving controllers, PAC does not possess any premise parameter, consequently free from the computation of tuning those parameters. PAC's consequent parameter namely Weight needs to be adjusted to realize the desired control efficacy. Inspired by the smooth employment and regulations of SMC theory in various neuro-fuzzy systems \cite{kayacan2012sliding,utkin2013sliding,kayacan2017type}, in our work, SMC learning theory is functioned to adapt the PAC's weight, which ascertains the boundedness of the tracking error in the closed-loop control system. Besides, it confirms adequate robustness in a system against exterior obstructions, parameter variations, and uncertainties \cite{ferdaus2019generic}. In designing the SMC, a time-varying $\textit{sliding surface}$ that restricts motion of a system to a plane can be exposed as follows:
\begin{equation}
S_{ss}(u_{PALM},u)=u_{src}(t)=u_{PALM}(t)+u(t)
\end{equation}

In this work, the sliding surface for BI-FWMAV and hexacopter plants to be controlled can be expressed as:
\begin{equation}
s_{l}=e+\gamma_{1}\dot{e}+\gamma_{2}\int_{0}^{t}e(t)dt
\end{equation}
where, $\gamma_{1}=\frac{\alpha_{2}}{\alpha_{1}},$ $\gamma_{2}=\frac{\alpha_{3}}{\alpha_{1}}$$,$$\: e$ is the error, i.e., the divergence from the trajectory obtained from the plant to the reference one. Here, the sliding parameters are initiated with small values such as $\alpha_{1}=1\times10^{-2},~\alpha_{2}=1\times10^{-3},~\alpha_{3}\thickapprox0.$ These parameters are further evolved by using dissimilar learning rates. Proper assignment of these rates supports to secure the desired parameters with minimal time, which affirms to gain stability in the closed-loop system swiftly. Engagement of these self-organizing sliding parameters shapes a fully model-free controller. Similar definition as of G-controller \cite{ferdaus2019generic} is maintained in this work as follows:       

$Definition:$ After a specific time $t_{k}$, a sliding motion will be formed on the sliding manifold $S_{ss}(u_{PALM},u)=u_{src}(t)=0$, where the state $S_{ss}(t)\dot{S}_{ss}(t)=u_{src}(t)\dot{u}_{src}(t)<0$ to be convinced for the entire period with some non-trival semi-open sub-interval of time expressed as $[t,\; t_{k})\subset(0,\; t_{k})$.

To enforce the above-mentioned definition of sliding mode condition, the weights of the proposed PAC are adapted accordingly.

In our proposed controller, the reliance of the subsidiary robustifying control term on the sliding surface can be formulated as follows:
\begin{equation}
u_{src}(t)=\alpha_{1}s_{l}\label{eq:u_aux}
\end{equation}
This subsidiary robustifying control term $u_{src}$ may endure high-frequency oscillations in contributing to the control input. Such repulsive occurrence in sliding mode control theory is termed as chattering effect. To suppress this chattering effect, control systems are primarily facilitated with saturation or sigmoid functions. In this work, due to simplicity, a saturation function is used to alleviate those detrimental consequence. 

The outcome from PALM ($u_{PALM}$) in PAC can be expressed as follows:
\begin{equation}
u_{PALM}(t)=\psi^{T}(t)\omega(t)\label{eq:u_g}
\end{equation}

The overall control signal as observed in Fig.~\ref{fig:g-controller-closed-loop} can be declared as follows:
\begin{equation}
u(t)=u_{src}(t)-u_{PALM}(t)\label{eq:u}
\end{equation}

\begin{thm}
The adaptation law to guarantee the boundedness of the tracking error and the consequent parameter, i.e. weights of PAC is expressed as:
\end{thm}

\begin{equation}
\dot{\omega}(t)=-\gamma \textbf{e} P b \psi
\end{equation}
where $\gamma>0$, $P$ is a positive definite matrix as exposed in Eq. \eqref{eq:P}, $b$ is an unknown positive constant, $\textbf{e}=[e ~~\dot{e}]$. These weight adaptation laws assure a stable closed-loop control system.

\subsubsection{Proof of boundedness of error and weights in PAC}
$Proof:$ Let us consider a $n$th order nonlinear system of the form as follows:
\begin{equation}\label{eq:nolinear_system}
    X^{(n)}=F(X,\dot{X},...,X^{(n-1)})+bu,~~ Y=X
\end{equation}
where $F(.)$ is an unknown continuous function, $b$ is an unknown positive constant. $u\in \Re$ is the input and $Y\in \Re$ is the output of the system. We have considered that the state vector $\bold{X}=(X_1, X_2,...,X_n)^T=(X, \dot{X},...,X^{(n-1)})^T\in \Re^n$ is available for measurement. Our control objective in this work is to push $Y$ to track a given reference trajectory $Y_r$. To be specific, the control objectives can be summarized as follows:
\begin{itemize}
    \item [i)] Parameters of the closed-loop control system should be uniformly ultimately bounded to confirm the global stability of that closed-loop system. In this work, $|\bold{X}(t)|\le M_{\bold{X}} < \infty$, $|\omega(t)|\le M_{\omega}<\infty$, and $|u(\bold{X}|\omega)|\le M_u< \infty$ for all $t\ge 0$, where $M_{\bold{X}}$, $M_{\omega}$, and $M_u$ are pre-defined design parameters. 
    \item [ii)] The closed-loop tracking error $e=Y_r-Y$ should be as small as possible by satisfying the conditions in i. 
\end{itemize}

Our proposed PAC should be able to achieve these control objectives. To show that, let us consider $\bold{e}=(e,\dot{e},...,e^{(n-1)})^T$ and $\bold{k}=(k_1,...,k_n)^T \in \Re^n$ carry such features that all roots of the polynomial $h(p)=p^n+k_1p^{(n-1)}+...+k_n$ will be in the open left-half plane. If we know about the function $F(.)$ and constant $b$, then the optimal control law for PAC can be expressed as follows:
\begin{equation}\label{eq:u_star1}
    u^*=\frac{1}{b}\big[-F(\bold{X}+Y_r^{(n)}+\bold{k}^T\bold{e})\big]
\end{equation}

Now, applying the optimal control law of Eq. \eqref{eq:u_star1} in Eq. \eqref{eq:nolinear_system} yields the follows:
\begin{equation}\label{eq:en}
    e^{(n)}+k_1e^{(n-1)}+...+k_ne=0
\end{equation}
From Eq. \eqref{eq:en}, it is obvious that $\underset{t\rightarrow\infty}{\lim}e(t)=0$, which is the desired control objective from the proposed controller. Since both $F(.)$ and $b$ are unknown, the optimal control law cannot be implemented. In such circumstance, PALM is used to approximate the optimal control law. Now utilizing Eq. \eqref{eq:u} in Eq. \eqref{eq:nolinear_system}, the following is obtained:
\begin{equation}\label{eq:non_eq2}
    X^{(n)}=F(\bold{X})+b\big[u_{src}-u_{PALM}\big]
\end{equation}

By following the approach in \cite{wang1993stable}, $bu^*$ is added and substructed to Eq. \eqref{eq:non_eq2}. After several modification, the error equation governing the closed-loop system can be exposed as follows:
\begin{equation}
    e^n=-k^T\bold{e}+b\big[u_{src}-u_{PALM}\big]
\end{equation}
By following the approach of \cite{wang1993stable,rysdyk1998fault}, the error dynamics in our work can be presented as follows:
\begin{equation}\label{eq:bold_e}
    \dot{\textbf{e}}=A\textbf{e}+b(u_{PALM}-\epsilon)
\end{equation}
where $\textbf{e}=[e ~~\dot{e}]^T$, $A=\left[\begin{array}{cc}
0 & 1\\
-\alpha_{1} & -\alpha_{2}
\end{array}\right]$, and $b=[0~~1]^T$. Let us denote an ideal weight as $\omega^*$ by defining the corresponding weight error as $\tilde{\omega}=\omega-\omega^{*}$. Now, the error dynamics can be rewritten as:
\begin{equation}
    \dot{\textbf{e}}=A\textbf{e}+b\psi^{T}\tilde{\omega}+b(\psi^{T}\omega^{*}-\epsilon)
\end{equation}
Assume that, in the domain of interest, the ideal weight brings the term $\psi^{T}(t)\omega^{*}(t)$ to within a $\Delta$-neighbourhood of the error $\epsilon$. It is bounded by 
\begin{equation}
\Delta^{*}\equiv\underset{z}{\text{sup}}\text{\ensuremath{\left|\psi^{T}(z)\omega^{*}-\epsilon(z)\right|}}
\end{equation}
where $z$ is the vector that contains all the variables of the inversion error. The term $\text{\ensuremath{\left|\psi^{T}(z)\omega^{*}-\epsilon(z)\right|}}$ represents a residual inversion error which is not modeled by the PAC. Therefore, $\Delta^{*}$ is defined as the worst-case difference between the error and its best approximation.  

In this work, the considered candidate Lyapunov function is as follows:
\begin{equation}\label{eq:V}
    V=\frac{1}{2}\textbf{e}^TP\textbf{e}+\frac{1}{2\gamma}\tilde{\omega}^{T}\tilde{\omega}
\end{equation}
where $\gamma>0$. For $\alpha_1>0$ and $\alpha_2>0$, $A$ in Eq. \eqref{eq:bold_e} is Hurwitz, $P$ is a positive definite matrix and satisfying the following equation:
\begin{equation}\label{eq:Q}
    A^TP+PA=-Q
\end{equation}
where $Q>0$. For all $Q>0$, the solution of Eq. \eqref{eq:Q} is $>0$. For $Q=I_2$, it is implying the following value of $P$:
\begin{equation}\label{eq:P}
P=\left[\begin{array}{cc}
\frac{\alpha_{2}}{\alpha_{1}}+\frac{1}{2\alpha_{2}} & \frac{1}{2\alpha_{1}}\\
\frac{1}{2\alpha_{1}} & \frac{1+\alpha_{1}}{2\alpha_{1}\alpha_{2}}
\end{array}\right]
\end{equation}

Now, the time derivative of the Lyapunov function $V$ with the substitution of Eq. \eqref{eq:bold_e} and Eq. \eqref{eq:Q} can be expressed as:
\begin{equation}\label{eq:vdot}
    \dot{V}=-\frac{1}{2}\textbf{e}^TQ\textbf{e}+\textbf{e}^TPb(\psi^{T}\omega^{*}-\epsilon)+\tilde{\omega}^{T}(t)\big(\textbf{e}^TPb+\frac{1}{\gamma}\dot{\tilde{\omega}}\big)
\end{equation}
The third term of Eq. \eqref{eq:vdot} is suggesting a design of the adaptation law as follows:
\begin{equation}
   \dot{\tilde{\omega}}=\dot{\omega}=-\gamma\textbf{e}^TPb \psi
\end{equation}
After employing the above adaptation law to Eq. \eqref{eq:vdot}, it can be reduced as:
\begin{equation}\label{eq:vdot2}
    \dot{V}=-\frac{1}{2}\textbf{e}^TQ\textbf{e}+\textbf{e}^TPb(\psi^{T}\omega^{*}-\epsilon)\\
    \le -\frac{1}{2}||\textbf{e}||_2^2+\Delta^*|\textbf{e}^TPb|
\end{equation}
Utilizing the inequality exposed in \cite{rysdyk1998fault}, we can write:
\begin{equation}\label{eq:inequality}
    \textbf{e}^TP\textbf{e}\le \overline{\lambda}(P)||\textbf{e}||_2^2
\end{equation}
Using Eq. \eqref{eq:inequality}, the following is obtained from Eq. \eqref{eq:vdot2}:
\begin{equation}\label{eq:vdot3}
    \dot{V}\le -\frac{\textbf{e}^TP\textbf{e}}{2\overline{\lambda}(P)}+\Delta^*\sqrt{\textbf{e}^TP\textbf{e}}\sqrt{\overline{\lambda}(P)}
\end{equation}
Equation \eqref{eq:vdot3} is strictly negative when:
\begin{equation}
    \sqrt{\textbf{e}^TP\textbf{e}}> 2\Delta^*(\overline{\lambda}(P))^{3/2}
\end{equation}
According to the LaSalle-Yoshizawa theorem, since $\dot{V}$ is strictly negative, it is sufficient to prove that $\textbf{e}$ and $\omega(t)$ will be remained bounded. In addition, if $\Delta^*=0$, there will be no approximation error, then $e(t)\rightarrow0$ as $t\rightarrow\infty$. It is guaranteeing the asymptotic stability of the system.

Unlike the conventional neuro-fuzzy systems, consequent parameters, namely weights of PAC, are also utilized in the antecedent part as exposed in Eq. \eqref{eq:dis_p_hp}. Therefore, it is important to confirm the boundedness of the weights while they were used in the antecedent part. In this work, the weights are initialized with small values (less than one). Then, they are updated recursively, where their boundedness is confirmed and explained in the above paragraphs. Absolute value is considered in the distance formula since the distance cannot be negative. The calculated distance from the weights is employed in the fuzzification layer. In Eq. \eqref{eq:mu_T1}, $\eta$ is a positive constant, the highest value for the ratio between the distance and maximum distance is one and positive. Therefore, the values for the exponent operator always remains negative and bounded by $\eta$, which confirms the stability of the antecedent part of the PAC.  

\section{Plant dynamics of MAVs \label{sec:confrontations MAV dynamics}}
Both hexacopter and BI-FWMAV plants were developed at UAV laboratory of the UNSW Canberra, where development of the BI-FWMAV plant was inspired by the work of \cite{kok2015low,wang2007effect}. These plants are used in our work to evaluate the proposed PAC's performance. In this section, we are initiating with hexacopter's nonlinear complex plant dynamics.
\subsection{Dynamics of hexacopter plant}
Rigid body dynamics of our hexacopter plant is formulated using Newton's second law of motion, where we have calculated correlations between the forces and moments acting on the hexacopter and translational and rotational accelerations. We have assumed our hexacopter plant as a traditional mass distribution, where the plane of symmetry was XZ plane. Such consideration makes the cross product of moments of inertia in YZ and XY plane zeros. After this simplified implementation, the equations of forces ($F_x, F_y, F_z$) and moments ($L, M, N$) in X, Y and Z axes are exposed in Eq. \eqref{eq:F_M}. For further clarifications, readers can go through \cite{nelson1998flight}, where equations in \eqref{eq:F_M} are derived elaborately.
\begin{equation}
\begin{array}{ccc}
F_{x} & = & m(\dot{u}+qw-rv)\\
F_{y} & = & m(\dot{v}+ru-pw)\\
F_{z} & = & m(\dot{w}+pv-qu)\\
L & = & I_{x}\dot{p}-I_{xz}\dot{r}+qr(I_{z}-I_{y})-I_{xz}pq\\
M & = & I_{y}\dot{q}+rp(I_{x}-I_{z})+I_{xz}(p^{2}-r^{2})\\
N & = & -I_{xz}\dot{p}+I_{z}\dot{r}+pq(I_{y}-I_{x})+I_{xz}qr
\end{array}\label{eq:F_M}
\end{equation}
where 
\begin{equation}
\begin{array}{ccc}
I_{x} & = & \int\int\int(y^{2}+z^{2})dm\\
I_{y} & = & \int\int\int(x^{2}+z^{2})dm\\
I_{z} & = & \int\int\int(x^{2}+y^{2})dm\\
I_{xy} & = & \int\int\int xydm\\
I_{xz} & = & \int\int\int xzdm\\
I_{yz} & = & \int\int\int yzdm
\end{array}
\end{equation}
where $m$ is the body mass in kg; $I_x, I_y, I_z$ are hexacopter's mass moments of inertia with regard to x, y, and z-axis respectively in $kgm^{2}$; $I_{xz}$ is the product of inertia. In our simulated plant, the practiced values of the above parameters are as: $m=3~ kg$, $I_x=0.04~ kgm^{2}$, $I_y=0.04~ kgm^{2}$, $I_z=0.06~ kgm^{2}$, $I_{xz}=0~ kgm^{2}$.       
\subsection{Complexities in hexacopter's over-actuated plant dynamics}
Unlike the conventional hexacopter plant with 6 degrees of freedom (DOF) rigid body dynamics, 8 DOF over-actuated hexacopter plant with medium fidelity is considered in this work. Two surplus DOF are induced from two moving masses. The masses can slide along their own rail aligned in lengthwise and sideways consecutively. To get a synopsis about the over-actuated simulated Hexacopter plant, its high-level diagram is shown in Fig. \ref{fig:Top-level-diagram-hexacopter}. The roll and pitch command to the "control mixing" block of Fig. \ref{fig:Top-level-diagram-hexacopter} is driven by "attitude controller". In the attitude control mechanism, the inner loop is controlled by a linear PID controller and the outer loop is regulated by PAC. The thrust command of "control mixing" is provided by PAC. Moving of mass to the longitudinal direction shifts the center of gravity (CoG) to X-axis, which is denoted by $CG_X$ in Fig. \ref{fig:Top-level-diagram-hexacopter}, and $CG_Y$ is expressing the shift of CoG to Y-axis due to the movement of mass to the lateral direction. Both the movements $CG_X$ and $CG_Y$ are controlled by PAC. In control mixing block, a simple linear mixing composition is utilized to convert the roll, pitch, yaw, and thrust commands to the required speed of motors. These signals are used to calculate the required thrust and torque of individual rotor based on the relative airflow faced by each of them and the commanded motor speed. Afterward, the total vertical force and yawing torque of the plant are calculated by summing up the thrust and torque of the individual rotor. The product of thrust to a single rotor and moment arm yields the rolling and pitching moments acting on the hexacopter. Finally, the controlled thrust along with the yawing torque, and rolling and pitching moments are fed to the rigid body dynamics to update the body state accordingly. Hexacopter's detailed rigid body dynamics, nonlinear aerodynamics along with associated complexities are discoursed in the supplementary document.
\begin{center}
	\begin{figure*}[t]
		\begin{centering}
			\includegraphics[scale=0.3]{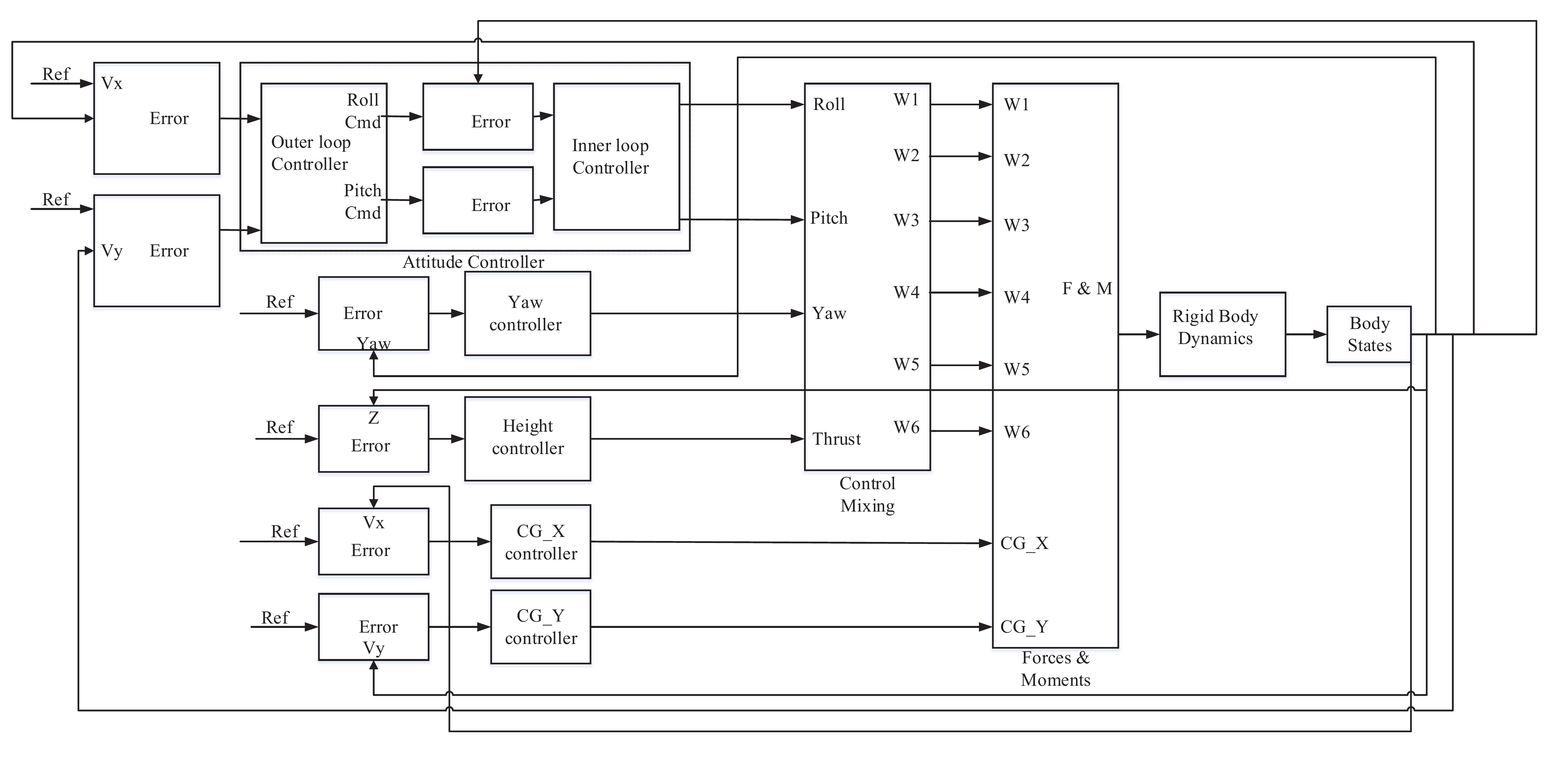}
			\par\end{centering}
		\caption{High-level presentation of the over-actuated simulated Hexacopter plant diagram\label{fig:Top-level-diagram-hexacopter} }
	\end{figure*}
	\par\end{center} 

A user-friendly graphical mask for the rigid body dynamics block allows users to alter the mass, moments of inertia and initial states on demand. During experimentation, errors are obtained by measuring the difference from actual to reference altitude and attitude. These errors, their derivatives, and reference trajectories are intakes for our proposed PAC, which is free from any plant parameters. Therefore, it can operate in a complete model-free manner. Finally, it is important to note that, after manifold simplification as explained in the supplementary document, the employed hexacopter dynamics is still highly nonlinear, complex with numerous parameters. Efficient controlling of such plant is difficult for model-based nonlinear controllers or model-free but the parameter-dependent conventional evolving controllers, whereas the premise parameter-free PAC exposes an improved control performance.   

\subsection{Dynamics of BI-FWMAV plant}              
Dynamics of the BI-FWMAV plant is highly nonlinear and expresses higher complexity than the hexacopter. It is mainly due to its lightweight and smaller size. The top-level diagram of the simulated BI-FWMAV plant is shown in Fig. \ref{fig:High-level-architecture_flap}. 
\begin{center}
	\begin{figure*}[t]
		\begin{centering}
			\includegraphics[scale=0.3]{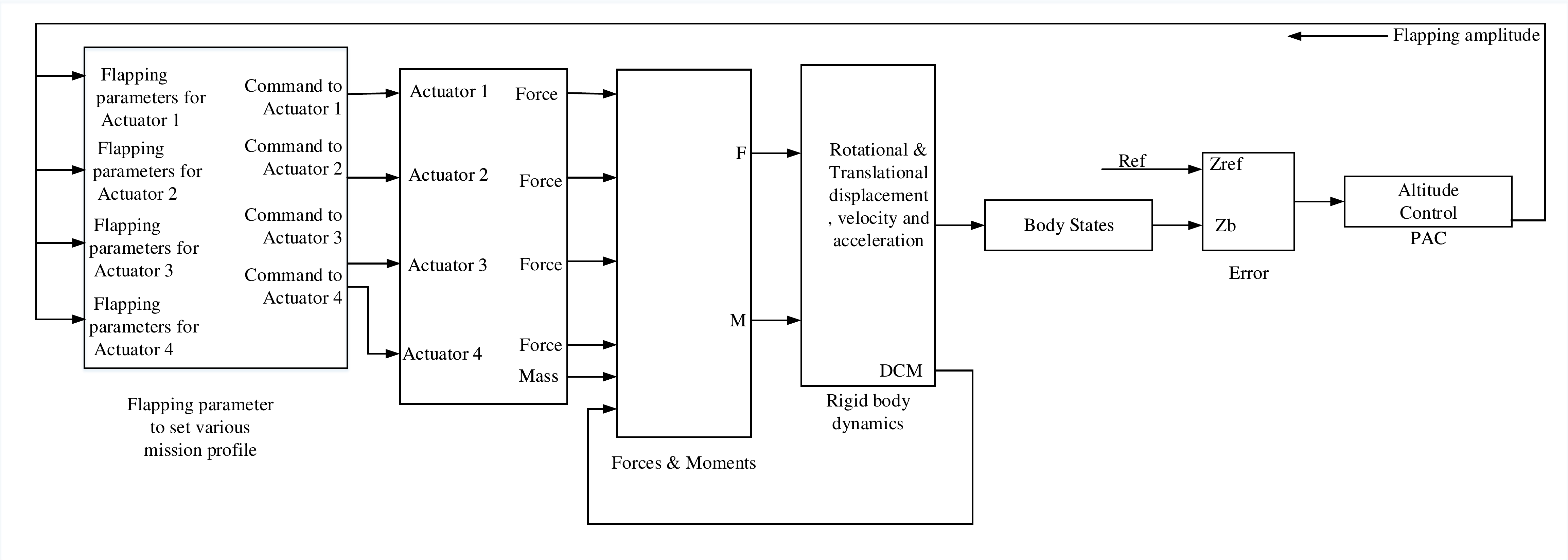}
			\par\end{centering}
		\caption{Top level framework of the BI-FWMAV plant\label{fig:High-level-architecture_flap}}
	\end{figure*}
	\par\end{center}

Four wings of the BI-FWMAV is operated by four actuators as exhibited in Fig. \ref{fig:High-level-architecture_flap}. From the analysis on dragonfly flight in \cite{kok2014systems}, the influence of seven different flapping parameters on the wings and actuators as well, are considered in developing the BI-FWMAV plant. These parameters are namely stroke plane angle ($Fp_{spa}$)(in rad) , flapping frequency ($Fp_{ff}$) (in Hz), flapping amplitude ($Fp_{fa}$) (in rad), mean angle of attack ($Fp_{aoa}$) (in rad), amplitude of pitching oscillation ($Fp_{po}$) (in rad), phase difference between the pitching and plunging motion ($Fp_{pd}$), time step ($Fp_{ts}$) (in sec). Exploring a variety of combinations of these parameters, the BI-FWMAV can carry out take-off, rolling, pitching, and yawing as explained in \cite{ferdaus2019generic}. Since our proposed controller is used to regulate the altitude of the BI-FWMAV, it is essential to know the dominant parameter in determining the altitude. After a successful parametric analysis, the flapping amplitude has turned out to be the dominant one in tracking the altitude. Individual forces and moments of actuators are combined to provide the demanded force and moment to the rigid body dynamics based on the relative airflow acting on each wing and the commanded actuator speed. The combined force utilized in our work can be formulated as follows:
\begin{equation}\label{eq:Ft}
F_T=F_{a_{1}}+F_{a_{2}}+F_{a_{3}}+F_{a_{4}}+(mg\times DCM)
\end{equation}               
where $m$ is the mass, $g$ is the acceleration due to gravity, $DCM$ is the direction cosine matrix, $F_{a_{i}}~(\text{where}~ i=1,2,3,4)$ is the force provided by the individual actuator. Similarly, the total moment necessary for the rigid body can be demonstrated as follows:
\begin{equation}\label{eq:Mt}
M_T=M_{a_{1}}+M_{a_{2}}+M_{a_{3}}+M_{a_{4}}
\end{equation}
where $M_{a_{i}}~(\text{where}~ i=1,2,3,4)$ is presenting the individual momentum of each wing and can be articulated as 
\begin{equation}\label{eq:M1}
M_{a_{i}}=F_{a_{i}}\times(CG-CP_{i})
\end{equation}
where $i=1,2,3,4$; $CG=[0\;0\;0];$ and $CP_{1}=[0.08\;0.05\;0]$; $CP_{2}=[0.08\;0.05\;0]$; $CP_{3}=[0.08\;-0.05\;0];$ $CP_{4}=[-0.08\;-0.05\;0]$; and $\times$ is presenting $(3\times3$) cross product. Finally, the accumulated force and moment are transformed into the body coordinate system, and all the required body states like three-dimensional angular displacements ($\phi, \theta, \psi$), angular velocities$\,$($\omega_{bx},\:\omega_{by},\:\omega_{bz}$) and accelerations$\,$($\alpha_{bx}=\frac{d\omega_{bx}}{dt},\:\alpha_{by}=\frac{d\omega_{by}}{dt},\:\alpha_{bz}=\frac{d\omega_{bz}}{dt}$) and linear displacements ($X_{b},$ $Y_{b}$, $Z_{b}$), linear velocities ($v_{bx},\: v_{by},\: v_{bz}$) and accelerations ($a_{bx}=\frac{dv_{bx}}{dt},\: a_{by}=\frac{dv_{by}}{dt},\: a_{bz}=\frac{dv_{bz}}{dt}$) are acquired, and the BI-FWMAV states are updated. Complex nonlinear wing dynamics and aerodynamics module of the BI-FWMAV has been clarified in Section 2 of the supplementary document. 

From the above discussion, and description of the supplementary document, it is obvious that the BI-FWMAV plant associates profuse parameters with high nonlinearity, though we have omitted some complexity in revealing precise wing kinematics. Deriving a precise mathematical model of such highly nonlinear, complex, and over-actuated plant is exceptionally laborious, where inclusion of uncertainties and uncharted disruption is more complicated or unfeasible in some cases. These perspectives necessitate a controller that performs well with a minimum or no knowledge about the system. Being model-free and self-evolving, our developed PAC is a suitable candidate. More importantly, the impediment of conventional evolving controllers, i.e. involvement of numerous free parameters is resolved here since our controller do not have any premise parameters and only depends upon the consequent parameters namely weights of the network. With such simplistic evolving structure, our proposed controller provides comparable and satisfactory tracking performance. 
                        
\section{Numerical experiments\label{sec:Numerical experiments}}
In our work, the proposed evolving PAC is used to regulate an over-actuated hexacopter and BI-FWMAV plant, where various altitude and attitude trajectories are tracked for both the MAVs. To be specific, PAC is appraised in tracking altitude of six different trajectories for BI-FWMAV. On the other hand, in hexacopter, the performance of PAC is witnessed both for tracking six different altitudes and sinusoidal attitude, which is detailed in upcoming subsections.

\begin{center}
	\begin{table}
		\caption{Measured features of various controllers in operating the BI-FWMAV (RT: rise time, ST: settling time, CH: constant height, VH: variable height, SS: sum of sine, RMSE: root mean square error, ms: millisecond, m: meter, MA: maximum amplitude, PSW: periodic square wave)}\label{tab:difr_all}
		\centering{}%
		\begin{tabular}{>{\raggedright}p{2.5cm}>{\raggedright}p{1.8cm}>{\centering}p{1cm}>{\centering}p{1cm}>{\centering}p{1cm}>{\centering}p{1.1cm}>{\centering}p{1.1cm}}
			\hline 
			\multirow{2}{2.5cm}{\textbf{Desired trajectory}} & \multirow{2}{1.8cm}{\textbf{Measured features}} & \multicolumn{5}{c}{\textbf{Control methods}}\tabularnewline
			\cline{3-7} 
			&  & \textbf{PID} & \textbf{FFNN} & \textbf{TS-Fuzzy}& \textbf{G-control} & \textbf{PAC}\tabularnewline
			\hline 
			\multirow{4}{2.5cm}{CH (MA 10 m)} & RMSE & \textbf{0.6460} & 0.7108 & 0.6693 & 0.6631 & 0.6668\tabularnewline
			& RT (ms) & 50.772 & 55.828 & 44.629& \textbf{41.208} & 47.207\tabularnewline
			& ST (ms) & 560.98 & 415.90 & 222.06& \textbf{127.15} & 147.09\tabularnewline
			& Peak (m) & 12.246 & 11.572 & 10.451& 10.813 & \textbf{10.306}\tabularnewline
			\hline 
			\multirow{4}{2.5cm}{VH with sharp change (MA 9 m)} & RMSE & \textbf{0.3303}& 0.4078 & 2.4951 & 0.3324 & 0.3561\tabularnewline
			& RT (ms) & 23.931 & 48.943& 43.949 & 50.892 & \textbf{13.728}\tabularnewline
			& ST (ms) & 8176.4 & 8386.3 & 8329.4& \textbf{8133.2} & 8166.5\tabularnewline
			& Peak (m) & 9.3732 & 9.6740 & 9.3010& \textbf{9.0069} & 9.2265\tabularnewline
			\hline 
			\multirow{4}{2.5cm}{VH with smooth change (MA 13 m)} & RMSE & 0.0895& 0.0556 & 0.0368 & \textbf{0.0228} & 0.0523\tabularnewline
			& RT (ms) & 8.8573 & 11.231 & 1.6537 & 4.187 & \textbf{0.1314}\tabularnewline
			& ST (ms) & 9884.3 & \textbf{9857.7} & 9871.1 & 9870.5 & 9872.1\tabularnewline
			& Peak (m) & 13.006 & 13.009& \textbf{13.004} & 13.019 & 13.006\tabularnewline
			\hline 
			\multirow{4}{2.5cm}{SS function (MA 11 m)} & RMSE & 0.4730 & 0.5356 & \textbf{0.4631} & 0.4963 & 0.5018\tabularnewline
			& RT (ms) & 21.109 & 31.092 & 20.616& \textbf{18.675} & 19.459\tabularnewline
			& ST (ms) & 9960.1 & 9960.7 & 9960.4 & \textbf{9960.1} & 9960.5\tabularnewline
			& Peak (m) & 11.468 & 11.502 & 11.455& \textbf{11.431} & 11.462\tabularnewline
			\hline 
			\multirow{4}{2.5cm}{PSW function (MA 11 m)} & RMSE & 2.7771& 3.3185 & 499.19 & \textbf{2.5098} & 2.5115\tabularnewline
			& RT (ms) & 548.93 & 474.36 & 2124.3 & 61.563 & \textbf{59.017}\tabularnewline
			& ST (ms) & 9924.1 & 9911.7 & 9940.7& \textbf{9603.2} & 9634.1\tabularnewline
			& Peak (m) & 12.794 & 12.573 & 1030.8& \textbf{11.007} & 11.294\tabularnewline
			\hline 
			\multirow{4}{2.5cm}{Staircase function (MA 12 m)} & RMSE & 0.3073 & 0.3791& 2.1796 & \textbf{0.2885} & 0.3072\tabularnewline
			& RT (ms) & 5996.0 & 4060.7 & \textbf{4015.7}& 5998.4 & 6000.3\tabularnewline
			& ST (ms) & 8384.2 & 8259.1 & 8094.4 & 8056.6 & \textbf{8055.1}\tabularnewline
			& Peak (m) & 12.453 & 12.458 & 12.074 & \textbf{12.007} & 12.198\tabularnewline
			\hline 
		\end{tabular}
	\end{table}
	\par\end{center}

\begin{center}
	\begin{figure}
		\begin{centering}
			\subfloat[]{\includegraphics[scale=0.13]{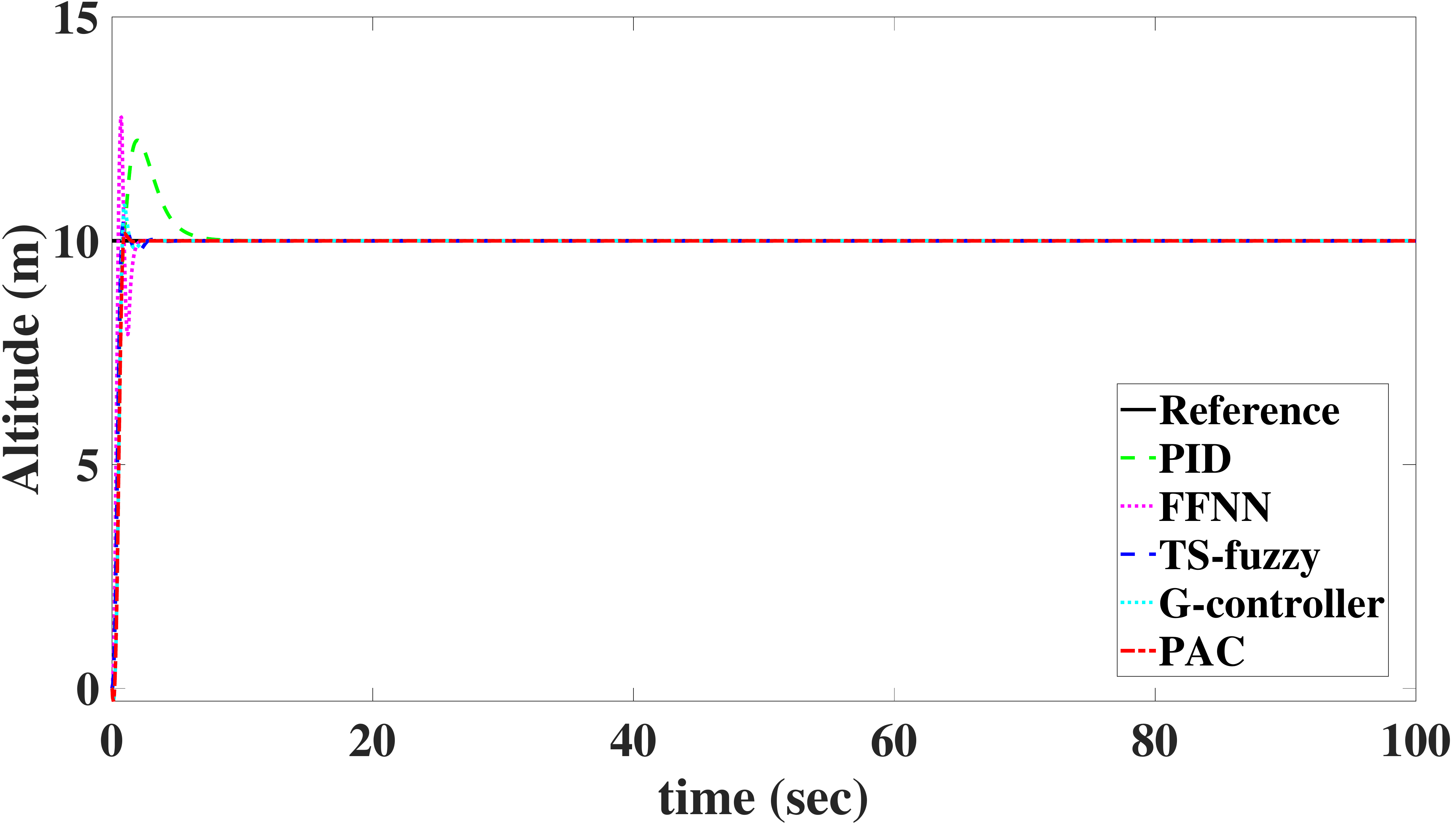}
				
			}\subfloat[]{\includegraphics[scale=0.13]{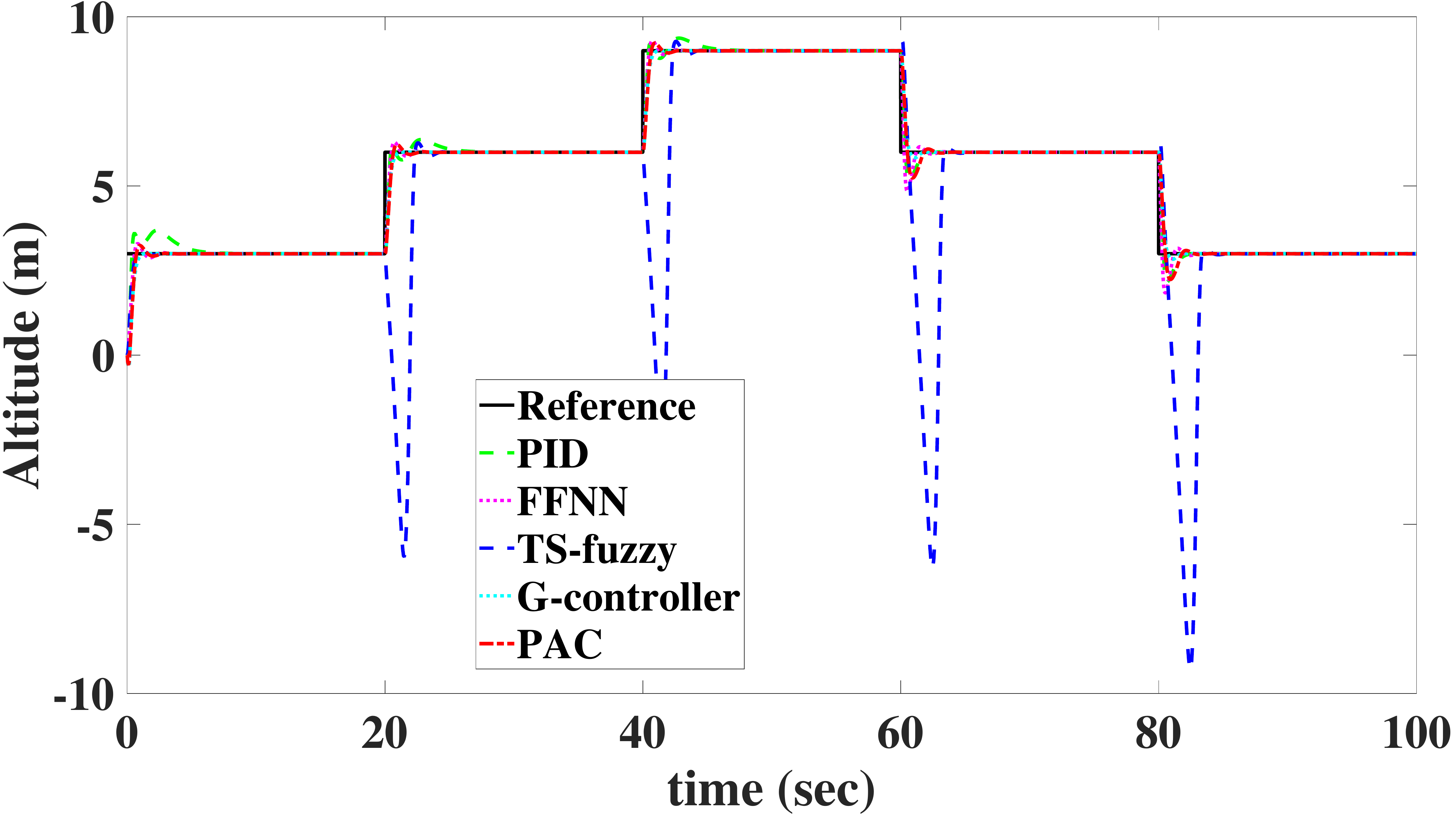}
				
			}
			\par\end{centering}
		\begin{centering}
			\subfloat[]{\includegraphics[scale=0.121]{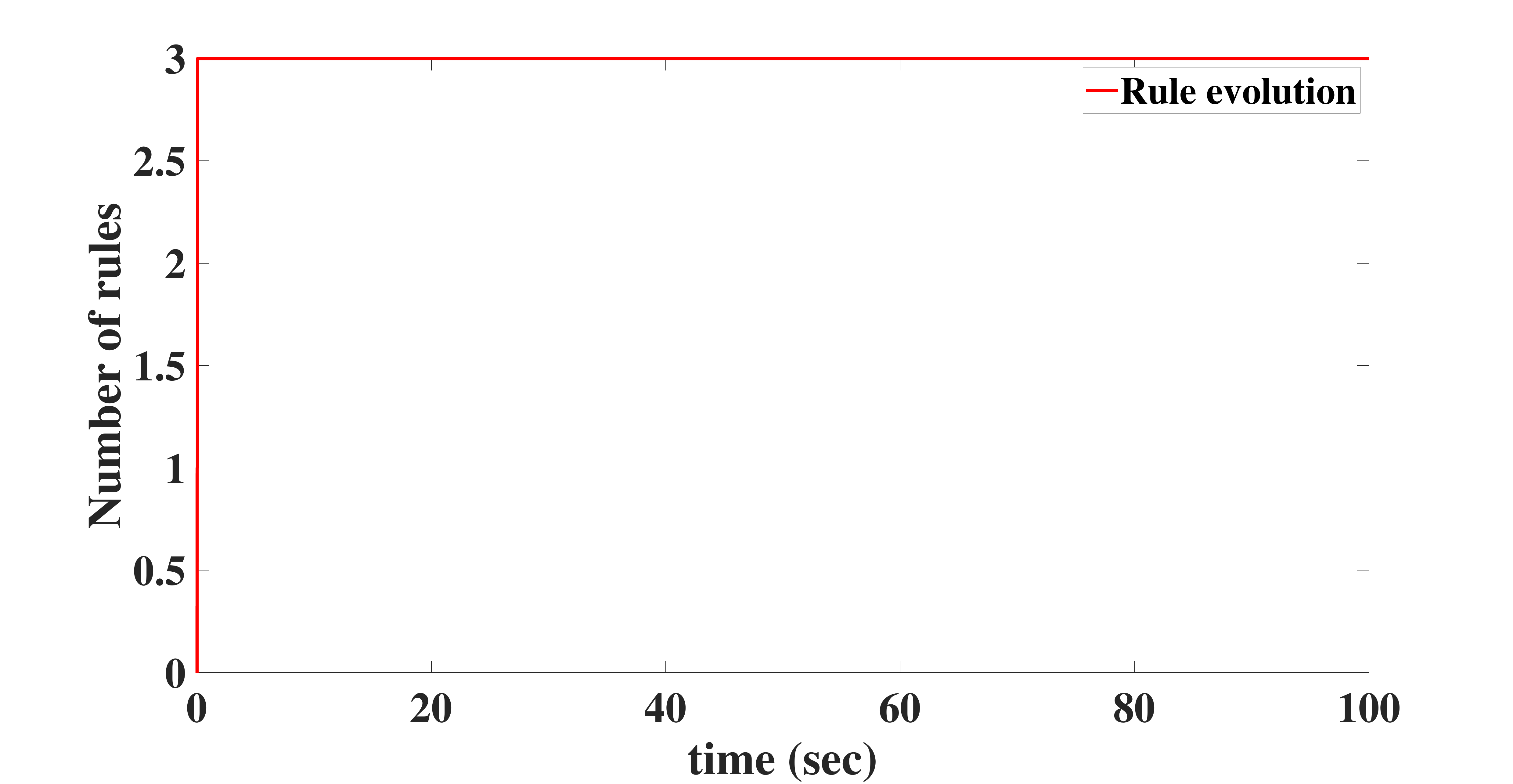}
				
			}\subfloat[]{\includegraphics[scale=0.121]{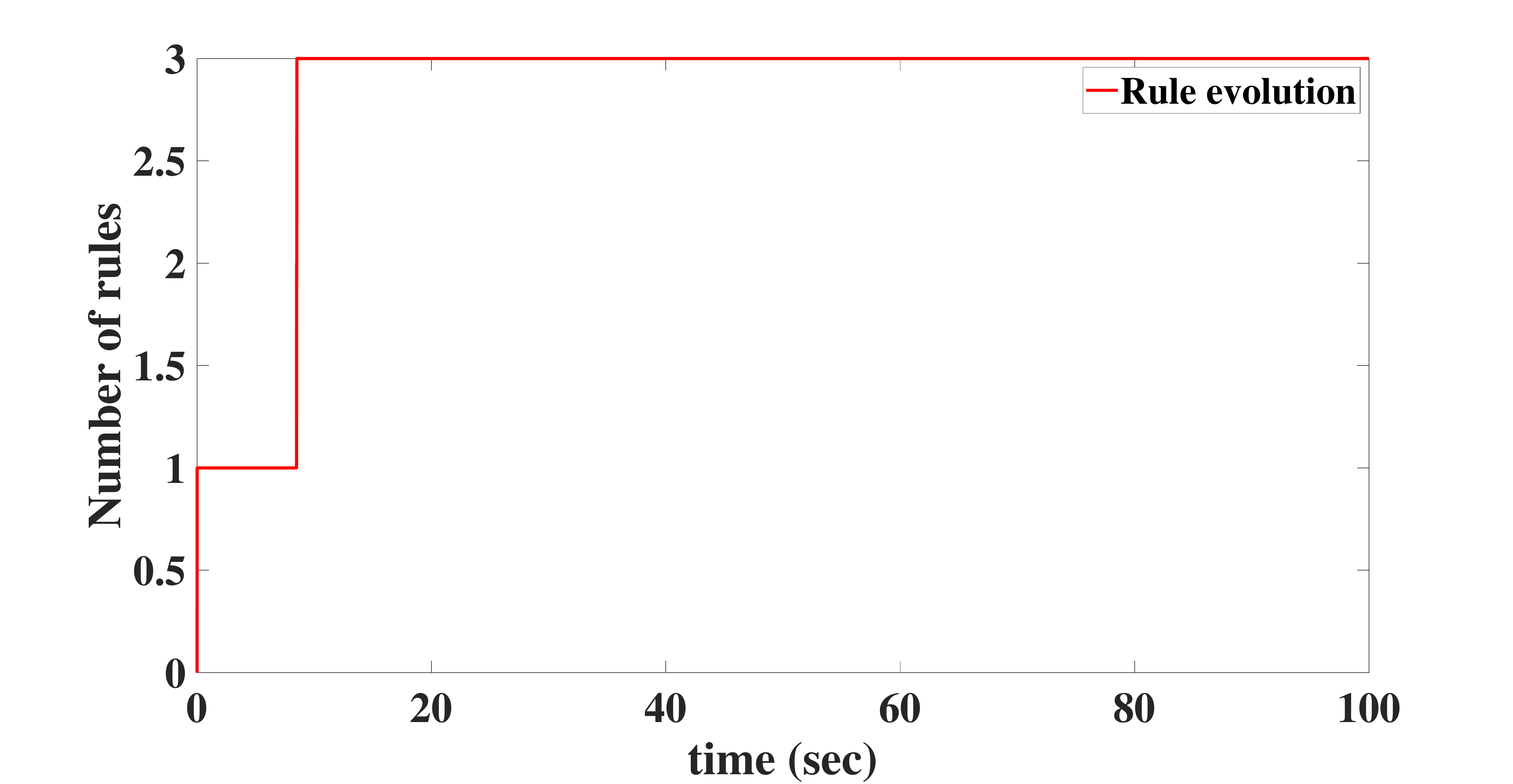}
				
			}
			\par\end{centering}
		\begin{centering}
			\subfloat[]{\includegraphics[scale=0.13]{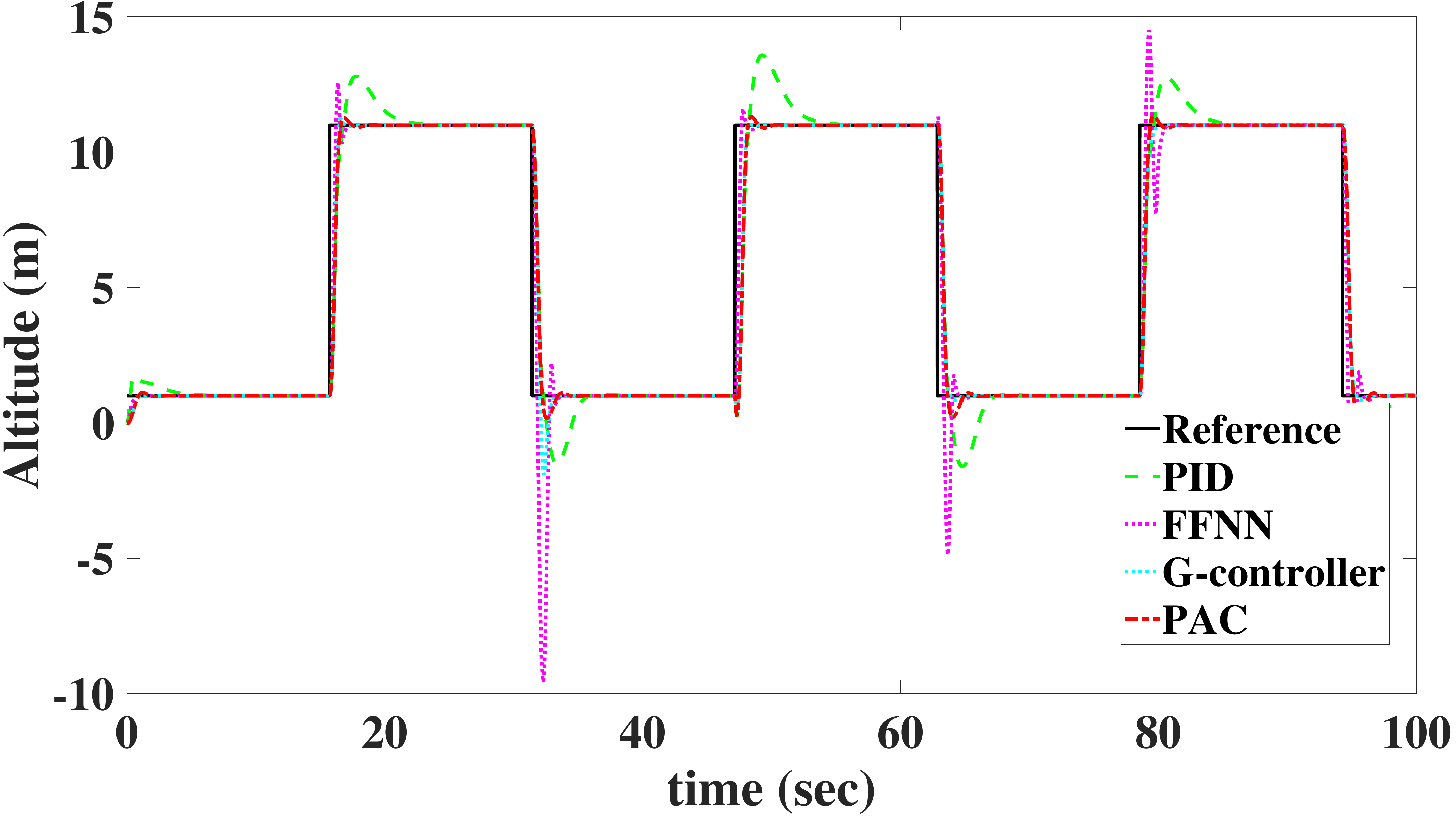}
				
			}\subfloat[]{\includegraphics[scale=0.13]{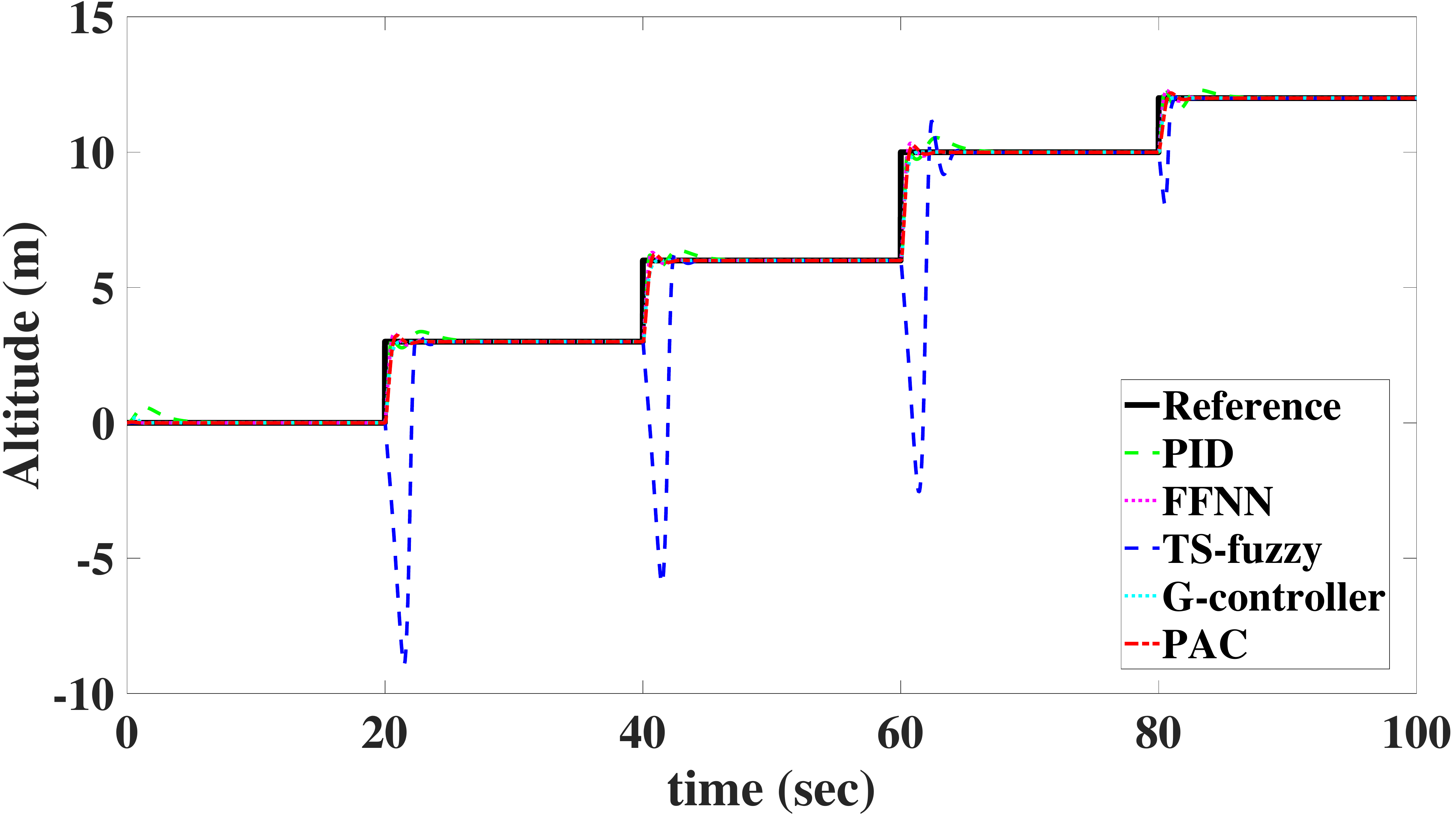}
				
			}
			\par\end{centering}
		\caption{Performance observation of different controllers in tracking altitude of BI-FWMAV, when the trajectories are (a) constant hovering, (b) variable heights with sharp edges, (e) periodic square wave function, (f) staircase function, rule evolution corresponding to (c) constant hovering, (d) variable heights with sharp edges}\label{fig: difr_all}
	\end{figure}
	\par\end{center}
\subsection{Experimental results from BI-FWMAV}
Our proposed evolving PAC was inspected for numerous tracking signals, and their consequent outcomes were contrasted with a recently developed evolving controller called G-controller \cite{ferdaus2019generic}, a Feed-Forward Neural Network (FFNN) based nonlinear adaptive controller, a Takagi-Sugeno (TS) fuzzy controller, and a PID controller. Our developed PAC's code was written in C programming language and made openly accessible in \cite{PAC_Code} along with the hexacopter plant to support reproducible research. The performance of all these controllers was observed in BI-FWMAV plant for 100 seconds, where the characteristics of six separate altitude trajectories were as: 1) an unaltered height of 10 meters, can be exposed as $Y_{r}(t)=10~m$; 2) variable heights with sharp edges, where the heights were altering from 3 m to 6 m after 20 seconds, and then from 6 m to 9 m for another 20 seconds and vice versa, afterward for another 40 seconds. The duration of hovering at a particular height was 20 seconds; 3) variable heights with smooth edges, i.e., the change from one height to another height is not so sharp as the previous trajectory; 4) sum of sines function, which was an amalgamation of a sine wave with a frequency of 0.3 $radsec^{-1}$, an amplitude of 4 m, and a bias of 6 m, and a cosine wave having a frequency of 0.5 $radsec^{-1}$, an amplitude of 3 m and a bias of 3 m; 5) a periodic square wave pulse, where the amplitude was varying between 1 m to 11 m, and its frequency is 0.2 $radsec^{-1}$; 6) a staircase function, where each step had a duration of 20 seconds. Individual heights of the first three steps are the same as a value of 3 m, which is 2 m in the last step. In these numerical experiments, FFNN based nonlinear adaptive controller operates better than its counterpart PID and TS-fuzzy controllers. Again, both the evolving controllers manifested better tracking performance than the FFNN controller. To acquire a deeper understanding of these manifestations, some of their desired features like root mean squared error (RMSE), rising time (RT) in milliseconds, settling time (ST) in milliseconds, and peak value of the overshoot are captured and tabulated in Table \ref{tab:difr_all}. All these anticipations are pictured in Fig. \ref{fig: difr_all} and detailed in the next paragraph.

In Fig. \ref{fig: difr_all} (a), controllers were facilitated to track a 10 m height trajectory, where both from PID and FFNN controller, higher overshoot with peak values more than 12 m were attested. Better performance with peak overshoots of less than 7\% of the height was noticed from the TS-fuzzy, and self-adaptive controllers. With regards to peak overshoot, the lowest values were exhibited by evolving controllers in all six different scenarios of Fig. \ref{fig: difr_all}, which is evidently signifying the superiority of their evolving structure. Unlike the benchmark evolving G-controller, our proposed one does not consist of any premise parameters. A network with fewer parameters supports PAC to procure prompter settlement, which was witnessed from the lowest settling time of 84.7 seconds in Fig. \ref{fig: difr_all} (a). It was considerably faster than the benchmark evolving G-controller since it demands 127.15 seconds to settle. In most cases of Fig. \ref{fig: difr_all}, lowest or very comparative settling time was observed from our proposed controller. The tracking accuracy of the PAC in terms of RMSE and rising time was not always the lowest one. Nonetheless, their achievements were still comparable and sometimes surpassed benchmark controllers.

\subsection{Experimental results from hexacopter plant}
All the controllers were assessed to track both the altitude and attitude (in terms of rolling and pitching) of the over-actuated hexacopter plant. Six separate trajectories of hexacopter's altitude were as follows: 1) a constant height with a value of 4 m; 2) altering heights with sharp edges, where the peak height was of 9 m; 3) altering heights with smooth edges, where the peak height was of 13 m; 4) a step function, which can be expressed as $3u(t-3)$, where $u(t)$ is a unit step function; and 5) a staircase function with a peak of 12 m; 6) sum of sines function, which was an amalgamation of a sine wave with a frequency of 0.3 $radsec^{-1}$, an amplitude of 4 m and a bias of 6 m, and a cosine wave having a frequency of 0.5 $radsec^{-1}$, an amplitude of 3 m and a bias of 3 m. In all conditions, a higher overshoot was perceived from the PID controller at each sharp changes as depicted in Fig. \ref{fig: hex_all}. The peak of this overshoot was lessened while the PID controller was replaced with the nonlinear adaptive FFNN controller. However, FFNN's performance was not consistent in all cases. Especially in dealing with the square wave trajectory, performance deteriorates significantly as portrayed in Fig.\ref{fig: hex_all} (c). This issue was managed by the evolving controllers effectively owing to self-adaptive architecture. Quick settlements were also testified from evolving controllers as recorded in Table \ref{tab:hex_all}. 
\begin{center}
	\begin{figure}
		\begin{centering}
			\subfloat[]{\includegraphics[scale=0.13]{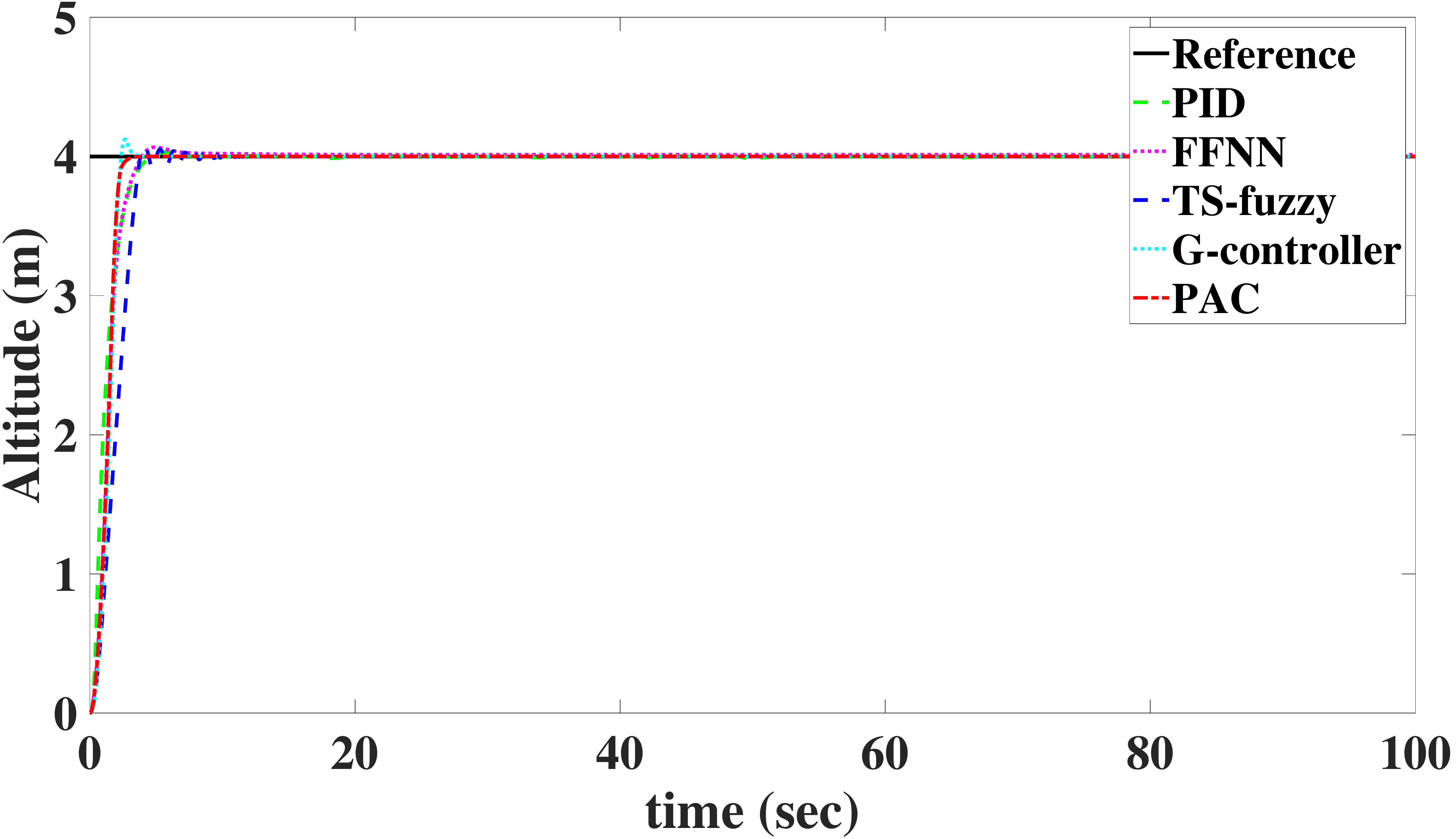}
				
			}\subfloat[]{\includegraphics[scale=0.13]{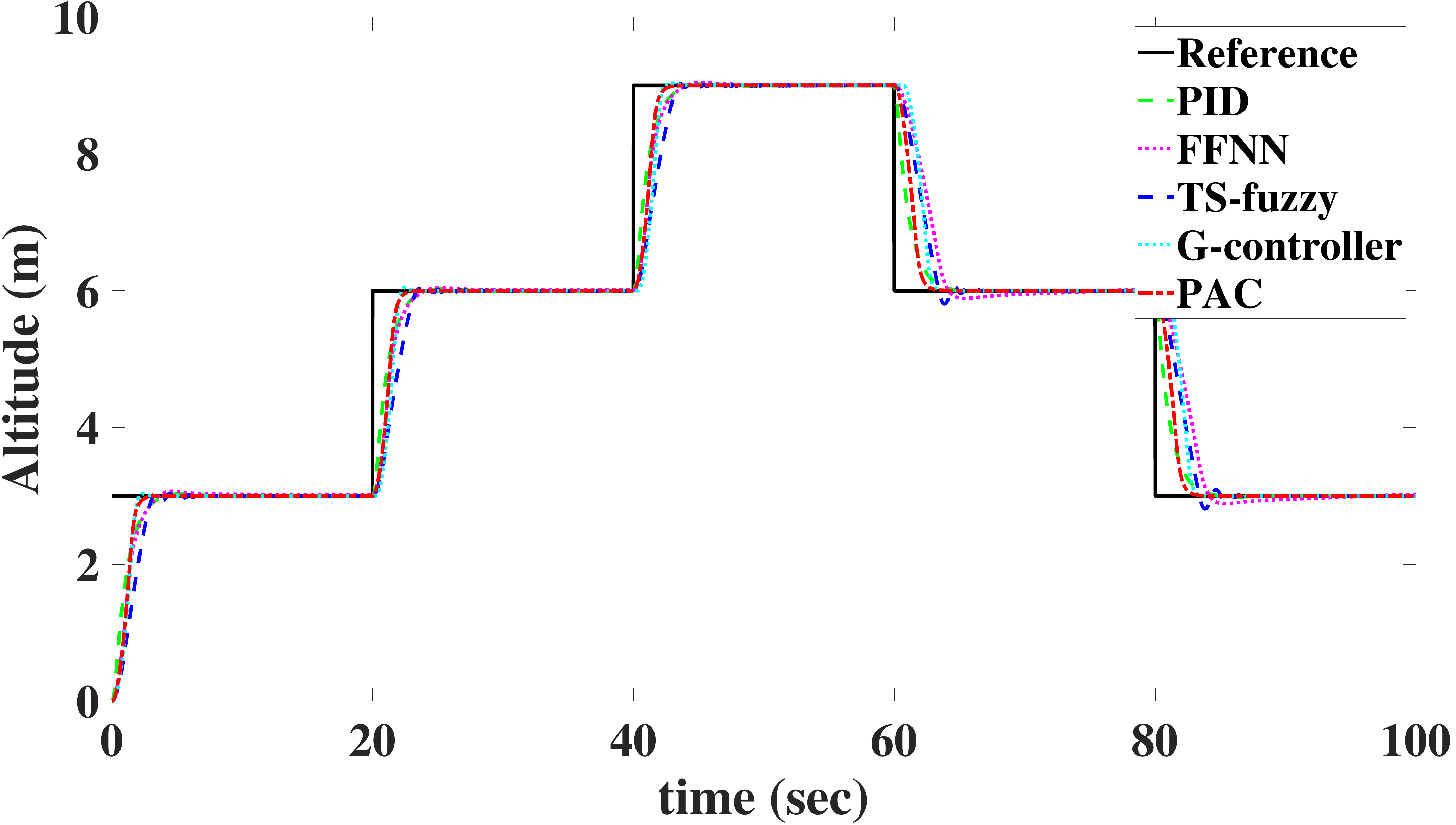}
				
			}
			\par\end{centering}
					\begin{centering}
			\subfloat[]{\includegraphics[scale=0.13]{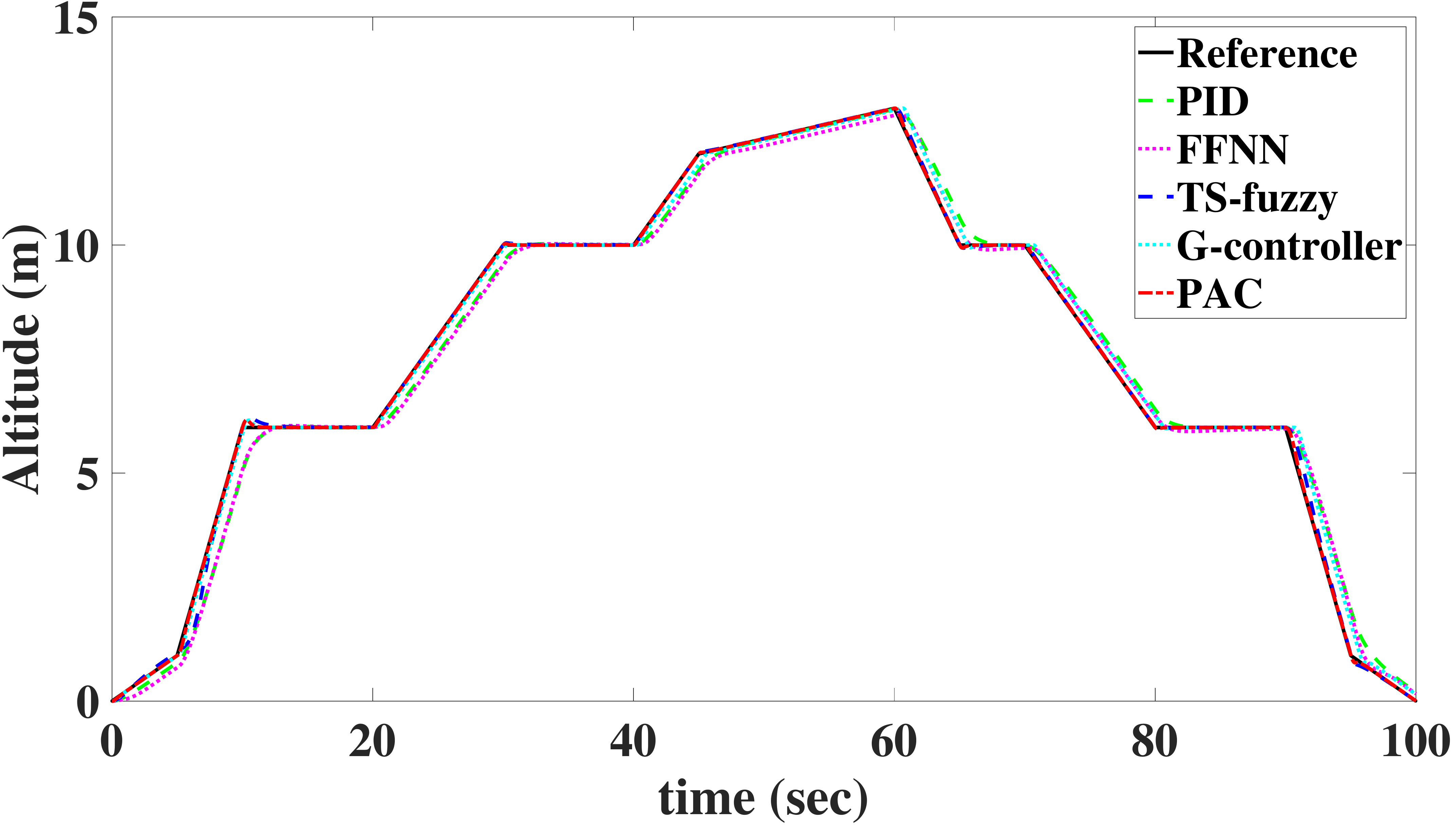}
				
			}\subfloat[]{\includegraphics[scale=0.13]{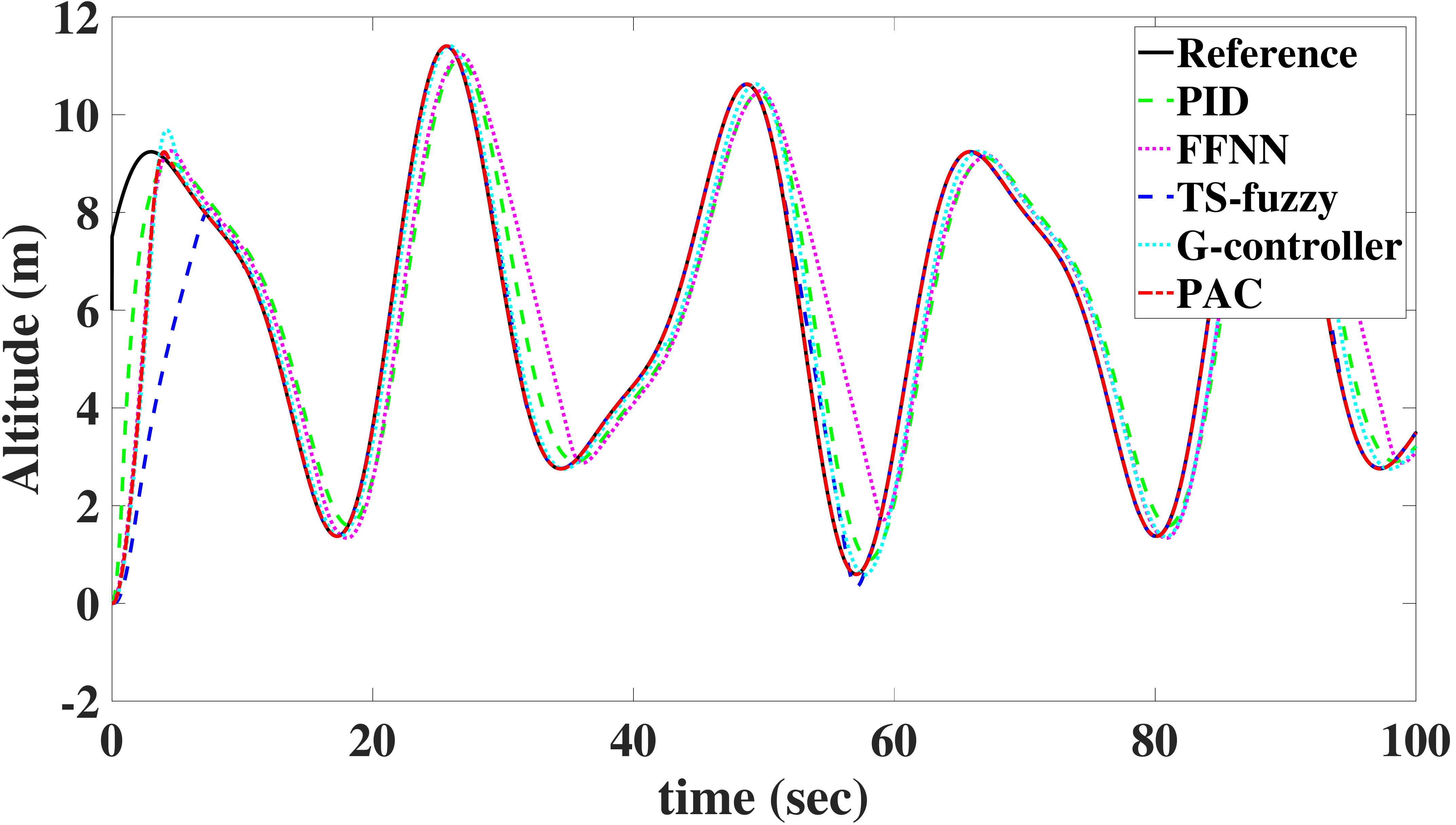}
				
			}
			\par\end{centering}
		\begin{centering}
			\subfloat[]{\includegraphics[scale=0.121]{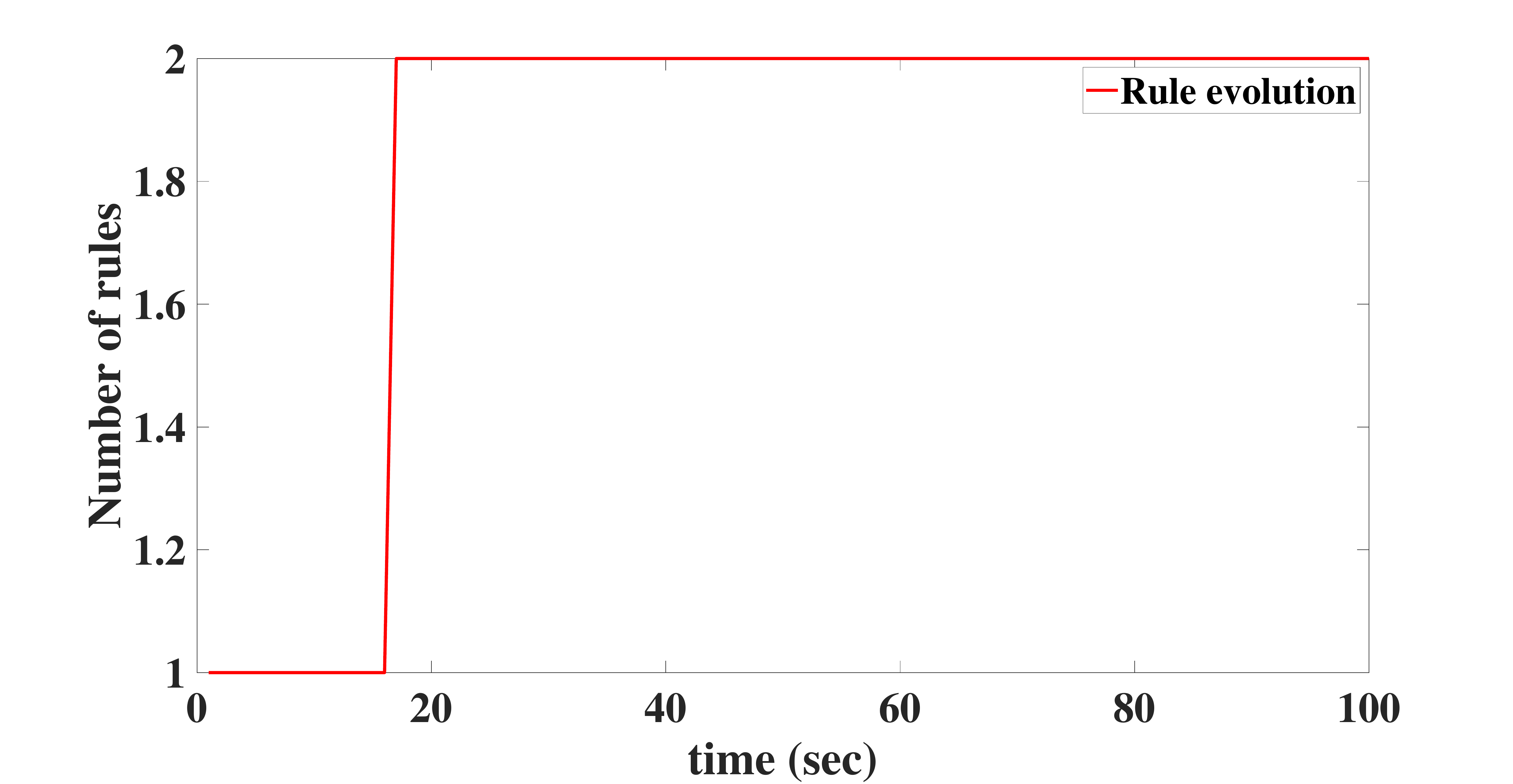}
				
			}\subfloat[]{\includegraphics[scale=0.13]{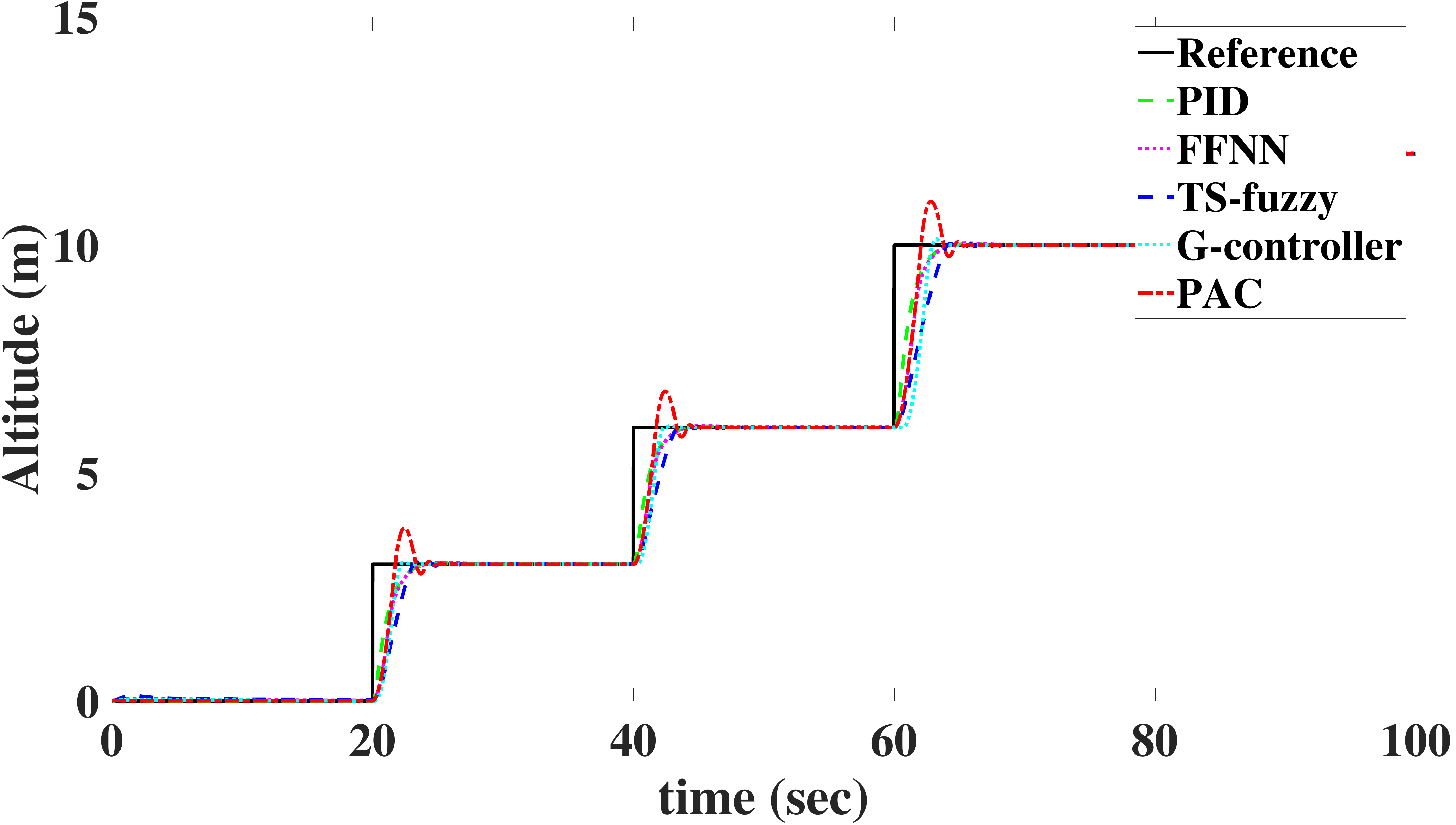}
				
			}
			\par\end{centering}
		\caption{Performance observation of different controllers in tracking altitude of hexacopter, when the trajectories are (a) constant hovering, (b) variable heights with sharp edges, (d) staircase function, (e) variable heights with smooth edges, (f) sum of sines function, and (c) evolution of rules corresponding to constant hovering}\label{fig: hex_all}
	\end{figure}
	\par\end{center}

\begin{center}
	\begin{table}
		\caption{Measured features of various controllers in regulating the hexacopter (RT: rise time, ST: settling time, CH: constant height, VH: variable height, ms: millisecond, m: meter, MA: maximum amplitude, PSW: periodic square wave, rad: radian)}\label{tab:hex_all}
		\centering{}%
		\begin{tabular}{>{\raggedright}m{2.5cm}>{\raggedright}p{1.8cm}>{\centering}p{1cm}>{\centering}p{1cm}>{\centering}p{1cm}>{\centering}p{1.1cm}>{\centering}p{1.1cm}}
			\hline 
			\multirow{2}{2.5cm}{\textbf{Desired trajectory}} & \multirow{2}{1.8cm}{\textbf{Measured features}} & \multicolumn{5}{c}{\textbf{Control method}}\tabularnewline
			\cline{3-7} 
			&  & \textbf{PID} & \textbf{FFNN} & \textbf{TS-fuzzy} & \textbf{G-control} & \textbf{PAC}\tabularnewline
			\hline 
			\multirow{4}{2.5cm}{CH (MA 4 m)} & RMSE & \textbf{0.3551} & 0.4221 & 0.4771 & 0.4239 & 0.4204\tabularnewline
			& RT (ms) & 208.97 & 199.15 & 259.03 & \textbf{141.66} & 144.63\tabularnewline
			& ST (ms) & 372.82 & 364.22 & 368.31 & 274.51 & \textbf{247.52}\tabularnewline
			& Peak (m) & 4.0272 & 4.0704 & 4.0714 & 4.0909 & \textbf{4.0015}\tabularnewline
			\hline 
			\multirow{4}{2.5cm}{VH with sharp change (MA 9 m)} & RMSE & \textbf{0.5574} & 0.7588  & 0.7607 & 0.6491 & 0.6537\tabularnewline
			& RT (ms) & 205.77 & 197.21 & 209.19 & \textbf{122.12} & 127.84\tabularnewline
			& ST (ms) & 8368.9 & 8649.9 & 8412.3 & \textbf{8249.2} & 8279.0\tabularnewline
			& Peak (m) & 9.0281 & 9.0406 & 9.0216 & 9.0022 & \textbf{9.0010}\tabularnewline
			\hline 
			\multirow{4}{2.5cm}{VH with smooth change (MA 13 m)} & RMSE & 0.3642 & 0.3651  & 0.1013 & 0.0273 & 0.0268\tabularnewline
			& RT (ms) & 122.21 & 144.33 & 6.7804 & 5.2776 & \textbf{2.4948}\tabularnewline
			& ST (ms) & 9932.5 & 9938.4 & \textbf{9869.4} & 9929.0 & 9927.0\tabularnewline
			& Peak (m) & 12.987 & 12.868 & 13.007 & \textbf{12.999} & 13.002\tabularnewline
			\hline 
			\multirow{4}{2.5cm}{Step function (MA 3 m)} & RMSE & \textbf{0.2420} & 0.2795 & 0.3078 & 0.2842 & 0.2834\tabularnewline
			& RT (ms) & 203.94 & 197.22 & 215.14 & 121.83 & \textbf{112.14}\tabularnewline
			& ST (ms) & 445.12 & 432.87 & 396.36 & \textbf{300.50} & 451.29\tabularnewline
			& Peak (m) & 3.0289 & 3.0676 & 3.0394 & \textbf{3.0040} & 3.6242\tabularnewline
			\hline 
			\multirow{4}{2.5cm}{Staircase function (MA 12 m)} & RMSE & \textbf{0.5221} & 0.6151 & 0.6959 & 0.5999 & 0.6237\tabularnewline
			& RT (ms) & 6019.9 & 6030.9 & 5970.0 & 6004.2 & \textbf{4227.9}\tabularnewline
			& ST (ms) & 8245.3 & 8285.4 & 8212.7 & \textbf{8179.5} & 8391.5\tabularnewline
			& Peak (m) & 12.026 & 12.039 & 12.017 & \textbf{11.998} & 12.587\tabularnewline
			\hline
			\multirow{4}{2.5cm}{Sum of sine function (MA 11 m)} & RMSE & 1.2270 & 1.7091 & 1.2636 & 1.0956 & \textbf{1.0856}\tabularnewline
			& RT (ms) & \textbf{59.638} & 114.04 & 165.62 & 125.94 & 122.53\tabularnewline
			& ST (ms) & 10050 & 10009 & \textbf{9955.6} & 10004 & 10210\tabularnewline
			& Peak (m) & \textbf{11.129} & 11.235 & 11.426 & 11.413 & 11.409\tabularnewline
			\hline
			\multirow{4}{2.5cm}{Pitching} & RMSE & 0.3513 & 0.0451 & N/A & 0.0466 & \textbf{0.0109}\tabularnewline
			& RT (ms) & 14.907 & 14.686 & N/A & 65.469 & \textbf{10.135}\tabularnewline
			& ST (ms) & 10057 & 10057 & N/A & \textbf{9982.9} & 10053\tabularnewline
			& Peak (rad) & 0.5615 & 0.5758 & N/A & \textbf{0.5398} & 0.5469\tabularnewline
			\hline 
			\multirow{4}{2.5cm}{Rolling} & RMSE & 0.1673 & N/A & N/A & 0.0290 & \textbf{0.0259}\tabularnewline
			& RT (ms) & 166.116 & N/A & N/A & 118.75 & \textbf{91.596}\tabularnewline
			& ST (ms) & 10037 & N/A & N/A & \textbf{9978.9} & 9979.9\tabularnewline
			& Peak (rad) & 0.3907 & N/A & N/A & 0.3513 & 0.4852\tabularnewline
			\hline 
		\end{tabular}
	\end{table}
	\par\end{center}

Furthermore, the rolling and pitching position (in rad) was observed with a sum of sine trajectory, which was a fusion of a sine wave with a frequency of 0.3 $radsec^{-1}$ and an amplitude of 0.3 m, and a cosine wave possessing a frequency of 0.5 $radsec^{-1}$and an amplitude of 0.5 m. The amplitude of the cosine wave was substituted with 0.4 m in tracking the rolling trajectory, where a better tracking was witnessed from the PAC than PID. Interestingly, both the adaptive FFNN controller and TS-fuzzy controller failed to track the rolling trajectory, which is the reason for their absence in Fig. \ref{fig: hex_roll_pitch} (a). Insertion of PAC yielded better tracking of pitching position, which is obvious from the lowest RMSE of 0.01. The captured RMSE for G-controller was higher, nearly 0.04 as recorded in Table \ref{tab:hex_all}.  
\begin{center}
	\begin{figure}
		\begin{centering}
			\subfloat[]{\includegraphics[scale=0.13]{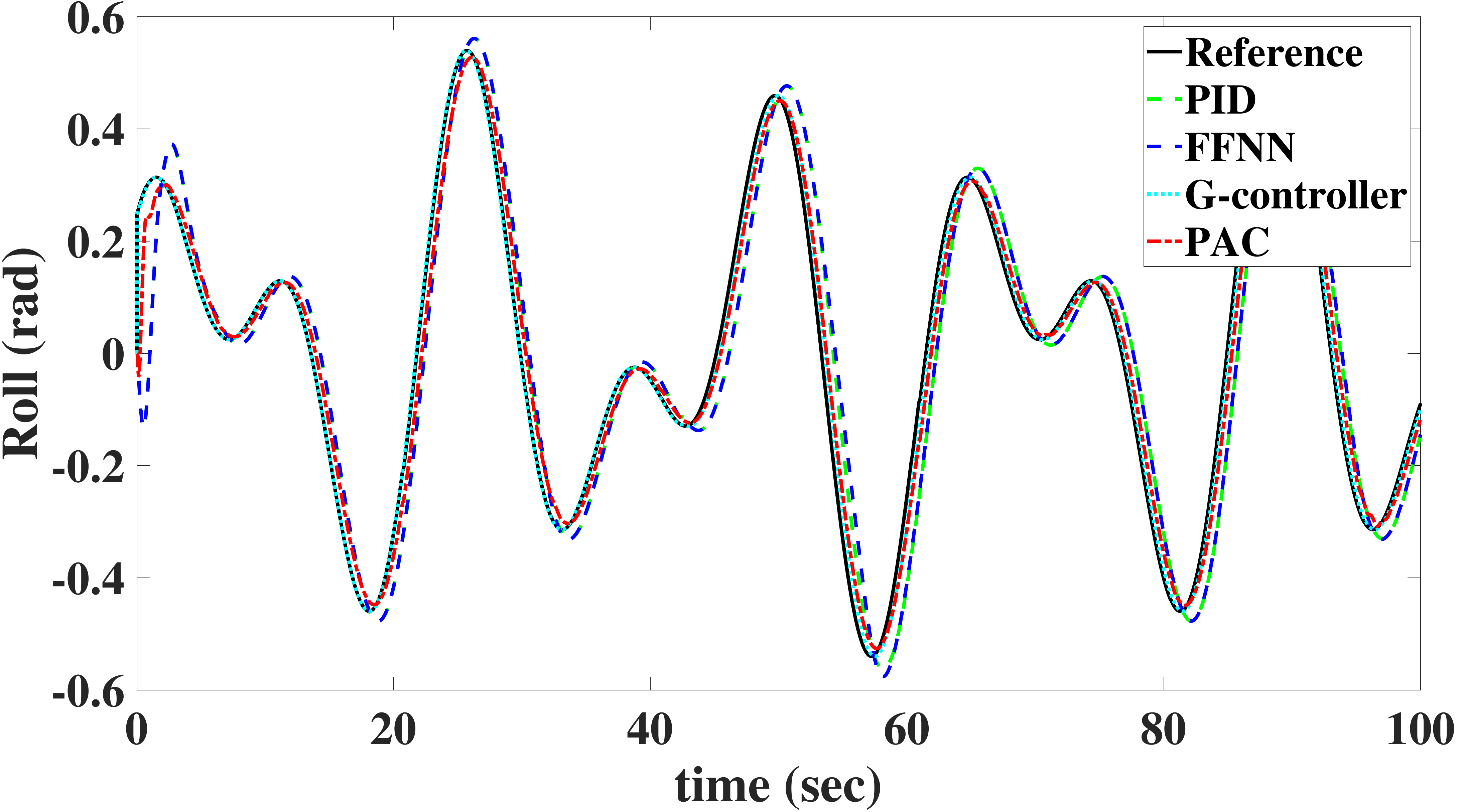}
				
			}\subfloat[]{\includegraphics[scale=0.13]{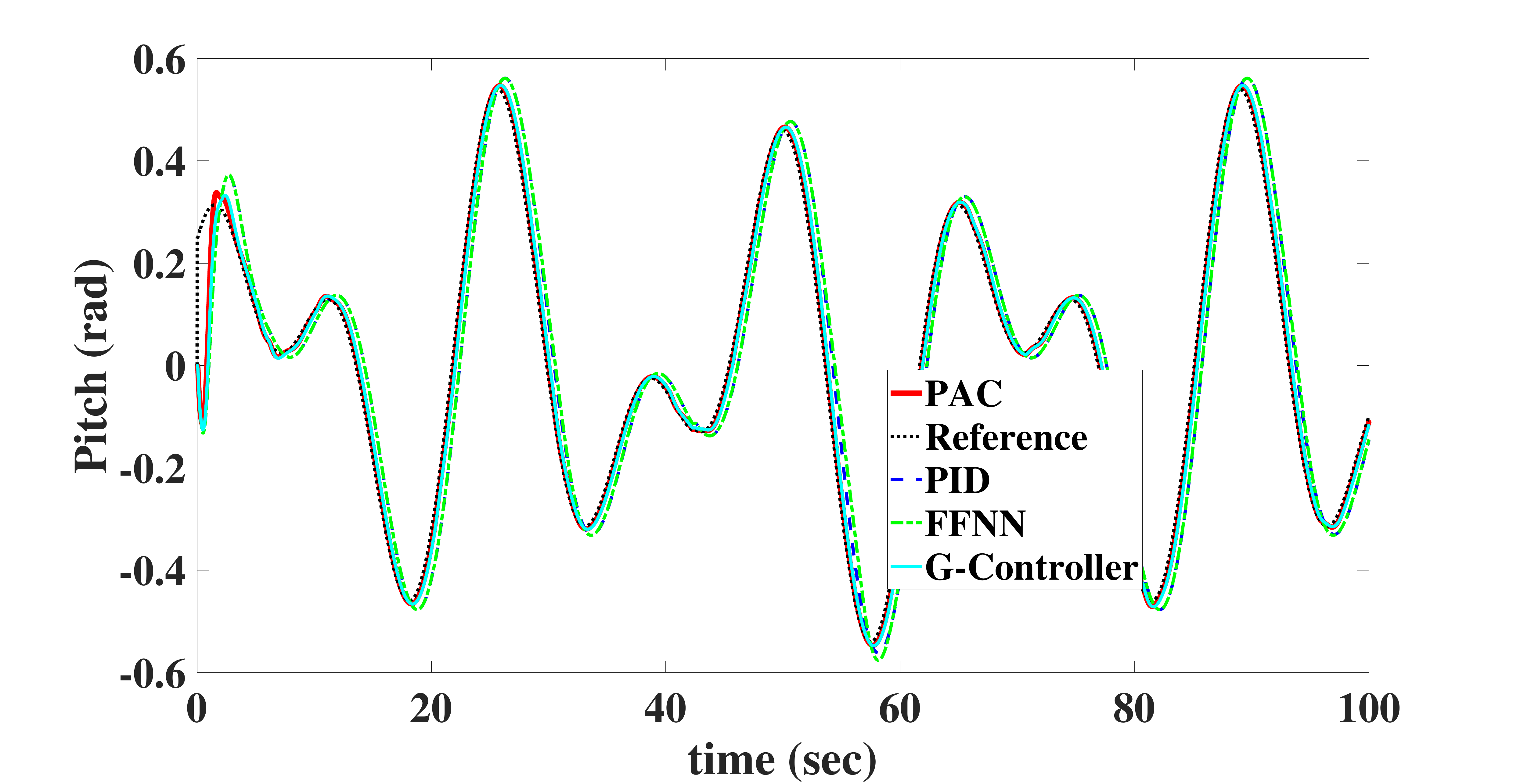}
				
			}
			\par\end{centering}
		\begin{centering}
			\subfloat[]{\includegraphics[scale=0.13]{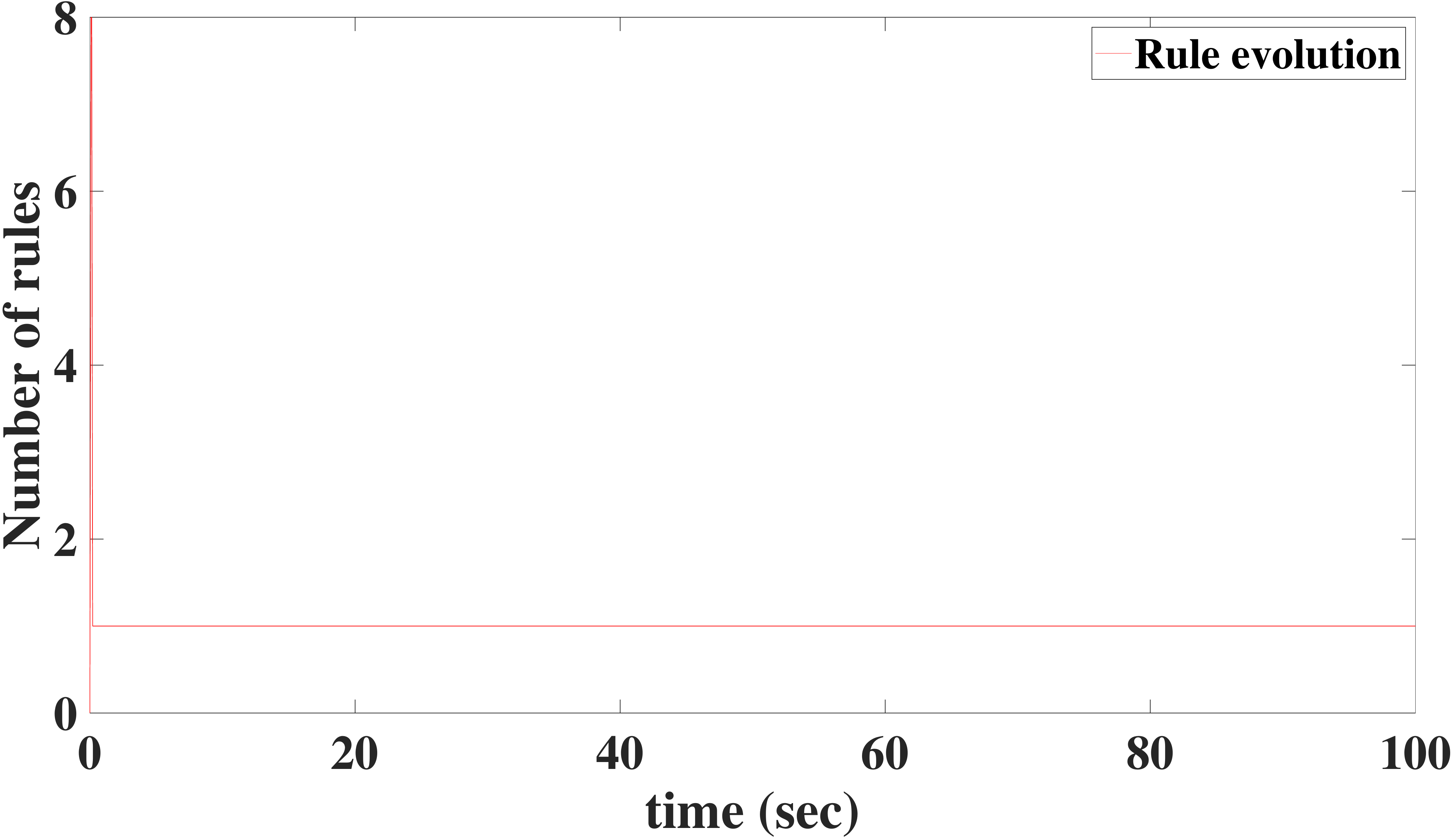}
				
			}\subfloat[]{\includegraphics[scale=0.13]{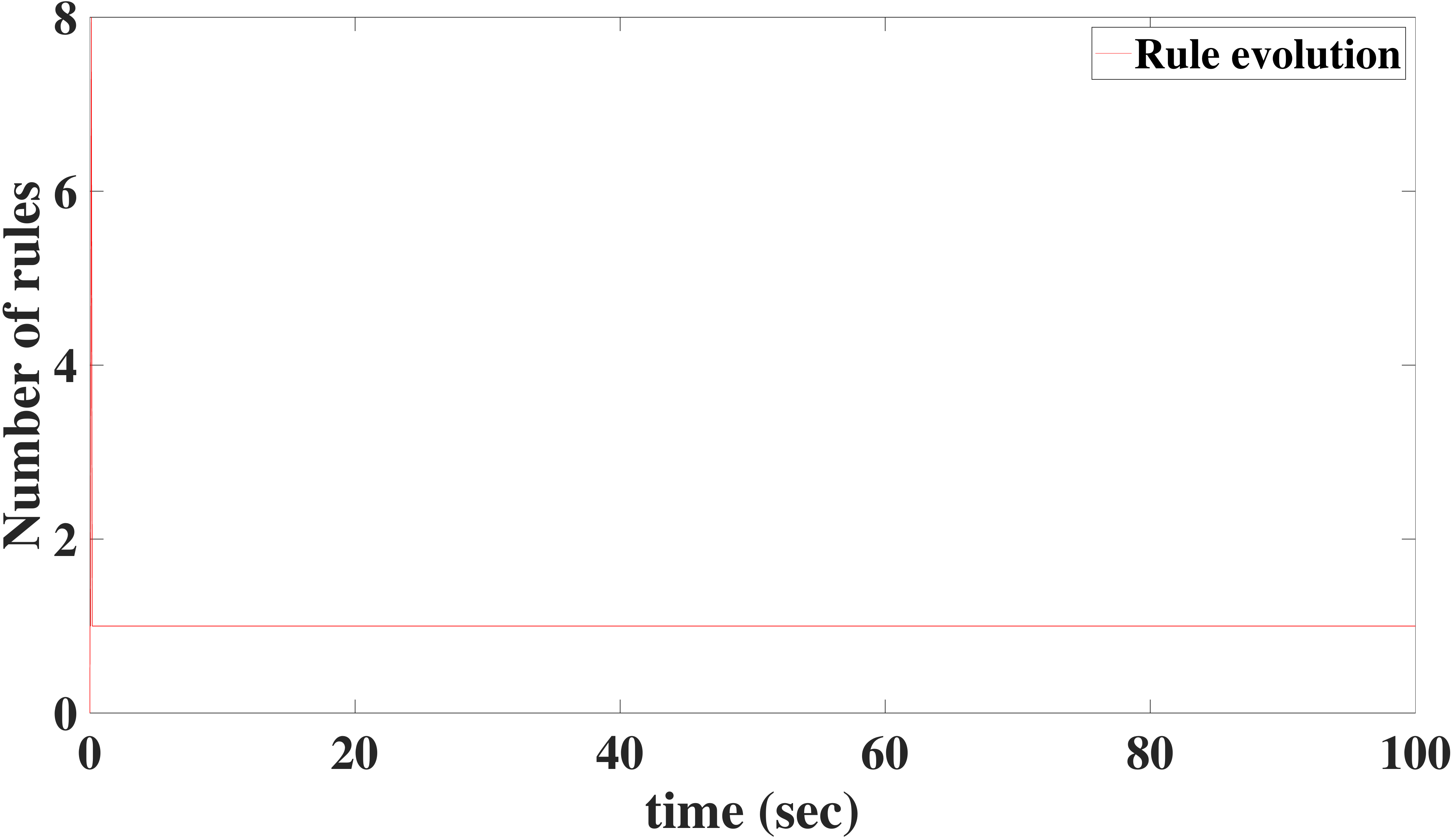}
				
			}
			\par\end{centering}
		\caption{Performance observation of different controllers in tracking desired (a) rolling, (b) pitching of the hexacopter MAV, (c) evolution of rules in tracking rolling, and (d) pitching in hexacopter}\label{fig: hex_roll_pitch}
	\end{figure}
	\par\end{center}

To sum up, superior or comparative tracking of trajectories were witnessed from our proposed evolving controller. Additionally, faster responses were obtained than the benchmark evolving G-controller, testifying the benefits of having an evolving structure with minimal network parameters.

\subsection{Robustness against uncertainties and noise}
In this work, a variety of disturbances were inserted in both BI-FWMAV and hexacopter's plant to verify controllers robustness against those disturbances. For instance, in the plant dynamics of BI-FWMAV, a sudden noise with a peak of 7 m for 0.1 seconds, and the discrete wind gust model Matlab block with a wind velocity 4 $ms^{-1}$ immediately after 2 seconds was embedded in the plant. The mathematical representation of the discrete gust is as follows:
\begin{equation}
V_{wind}=\begin{cases}
\begin{array}{c}
0\\
\frac{V_{m}}{2}\left(1-\cos\left(\frac{\pi x}{d_{m}}\right)\right)\\
V_{m}
\end{array} & \begin{array}{c}
x<0\\
0\le x\leq d_{m}\\
x>d_{m}
\end{array}\end{cases}
\end{equation}
here $V_m$ is the amplitude of the gust, $d_{m}$ is denoting the length, $x$ is the traveled-distance, and $V_{wind}$ is the resultant wind velocity in the body axis frame.

Effects of both wind gust and sudden peak noise were observed for all six different altitude trajectories of BI-FWMAV, which are depicted in Fig. \ref{fig: difr_all_gust}. From a closer view, obvious performance degradation in dealing with disturbances was witnessed from the non-adaptive PID controller in all cases. In the FFNN controller, due to the adaptation of the network parameters, it performed better than the PID. In some cases, the TS-fuzzy controller performs better than PID. However, its performance was not consistent for all the trajectories. Both FFNN and TS-fuzzy controllers performance deteriorates in tracking trajectories with sharp changes because of the absence of structure adaptation mechanism. On the other hand, sharper settlement and recovery from the adverse effect of gust were sighted from evolving controllers in our experiments. At the same time, in rejecting sudden peak noise, both the PAC and the G-controller dominated other benchmark controllers, where the lowest peaks were witnessed from them. Such accomplishments were possible due to their evolving structure with an adaptation of fewer parameters than the benchmark controllers.

In the hexacopter dynamics, a sharp peak noise with an amplitude of 2 m for 0.1 seconds was implanted to observer robustness of the controllers. Effects of disturbance were witnessed for four different altitude trajectories of hexacopter. RMSE, settling time, rise time and peak overshoot values for all those trajectories were tabulated for all benchmark and proposed controller in Table \ref{tab:hex_all_noise}. Such perturbation was handled effectively by evolving controllers than their static counterparts as attested in Fig. \ref{fig: hex_all_gust}. A high peak and slow settlement were detected in PID, TS-fuzzy and FFNN controllers. On the contrary, a negligible overshoot with a rapid settlement was inspected from the evolving controllers. For example, after closely observing the constant altitude trajectory in Fig. \ref{fig: hex_all_gust} (c), recorded values of rising time were less than 6 ms from the evolving controllers, which was more than 100 ms in FFNN and PID controllers. A similar phenomenon was observed in remaining trajectories, which is evidently declaring the improved robustness against uncertainties of the PAC in contrast with the benchmark static controller.     
\begin{center}
	\begin{figure}
		\begin{centering}
			\subfloat[]{\includegraphics[scale=0.13]{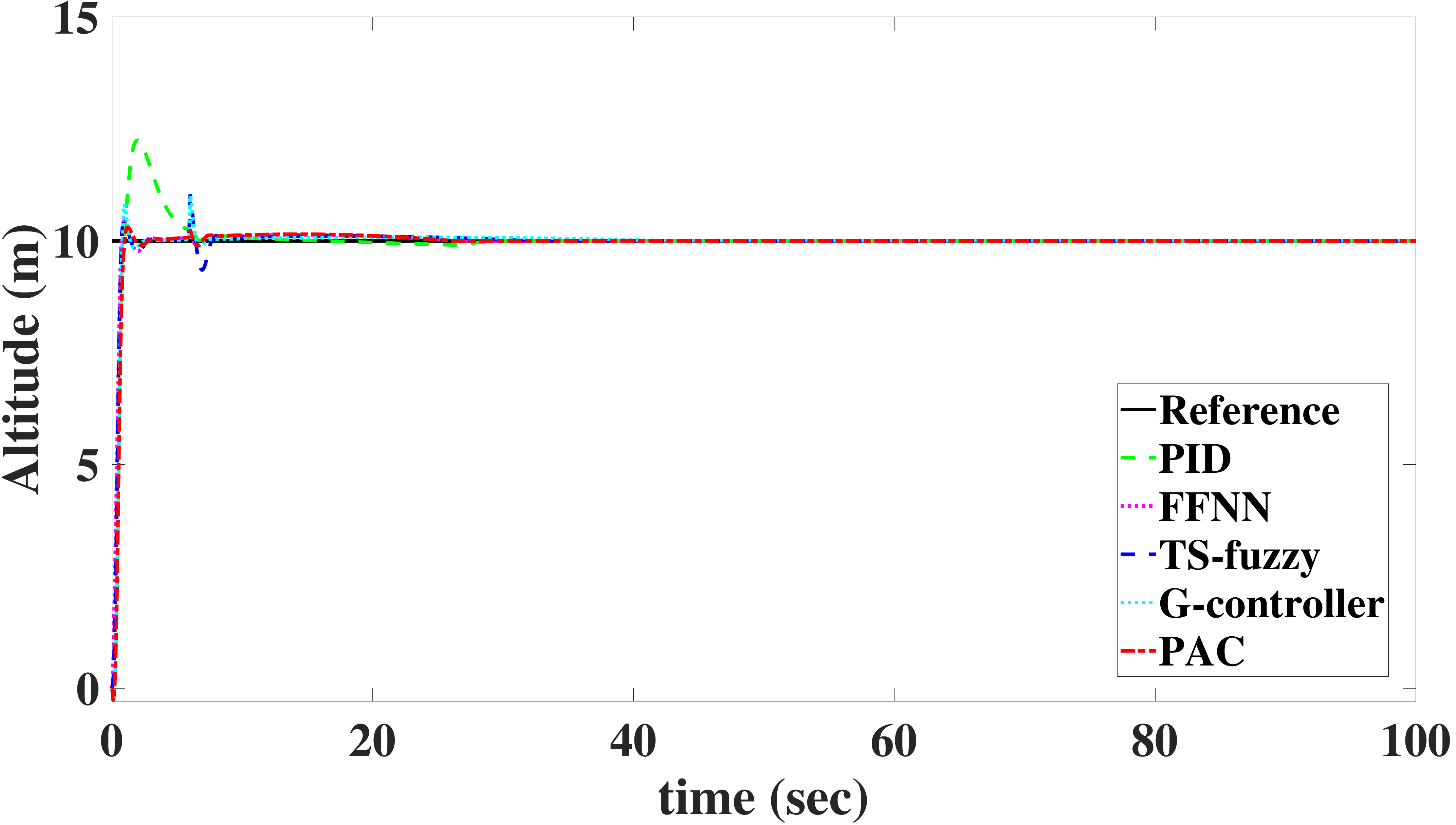}
				
			}\subfloat[]{\includegraphics[scale=0.13]{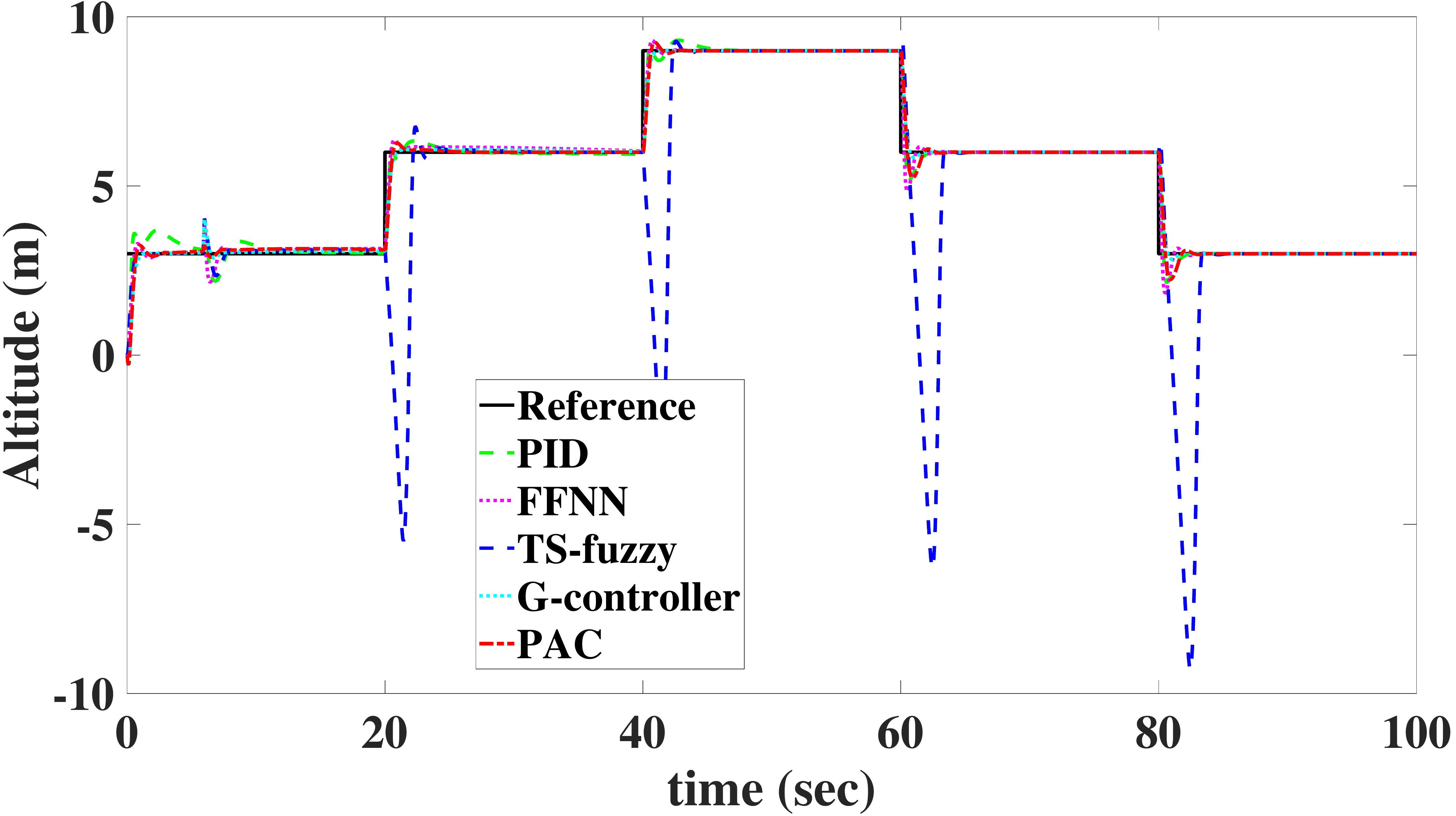}
				
			}
			\par\end{centering}
		\begin{centering}
			\subfloat[]{\includegraphics[scale=0.13]{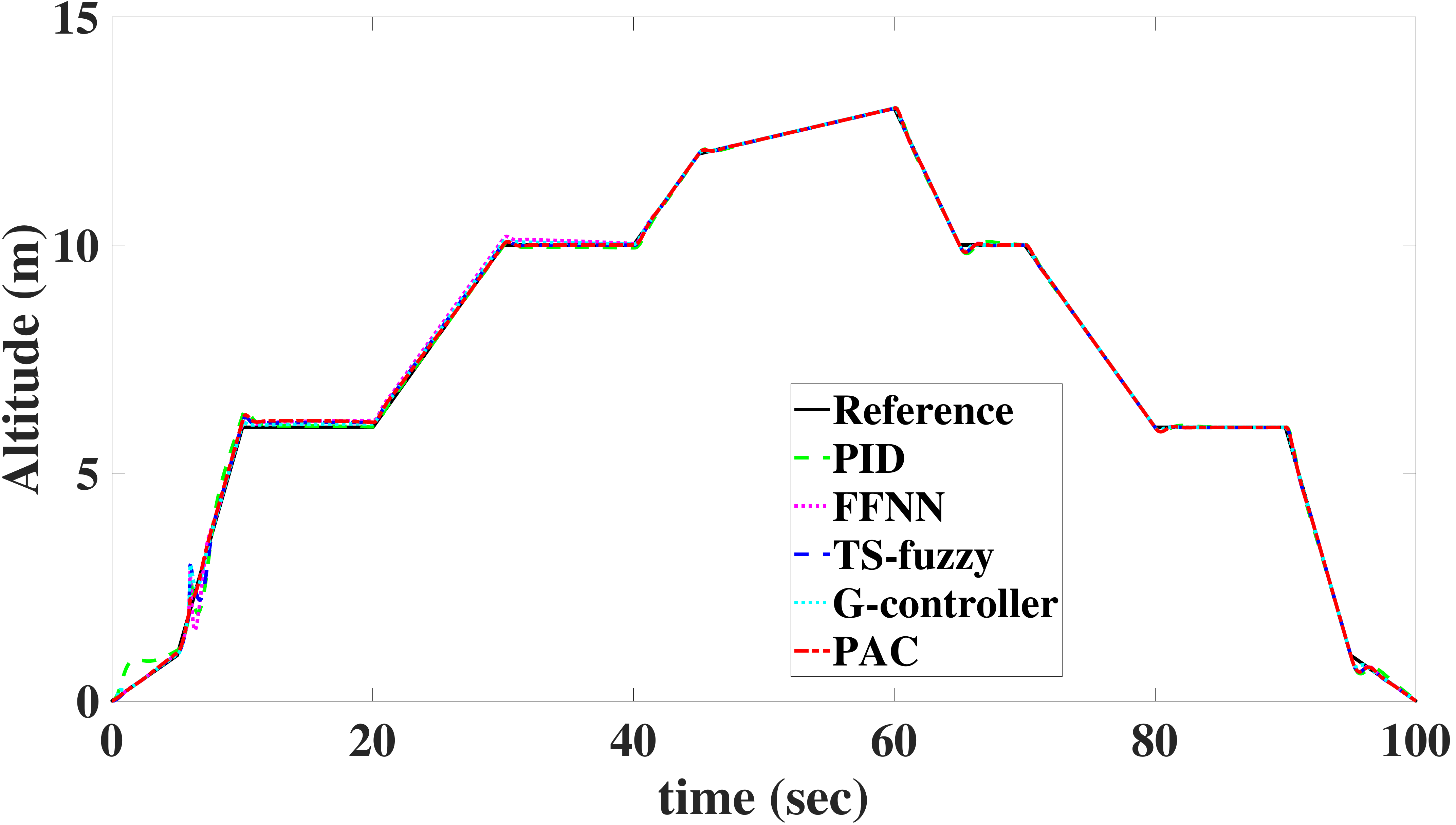}
				
			}\subfloat[]{\includegraphics[scale=0.13]{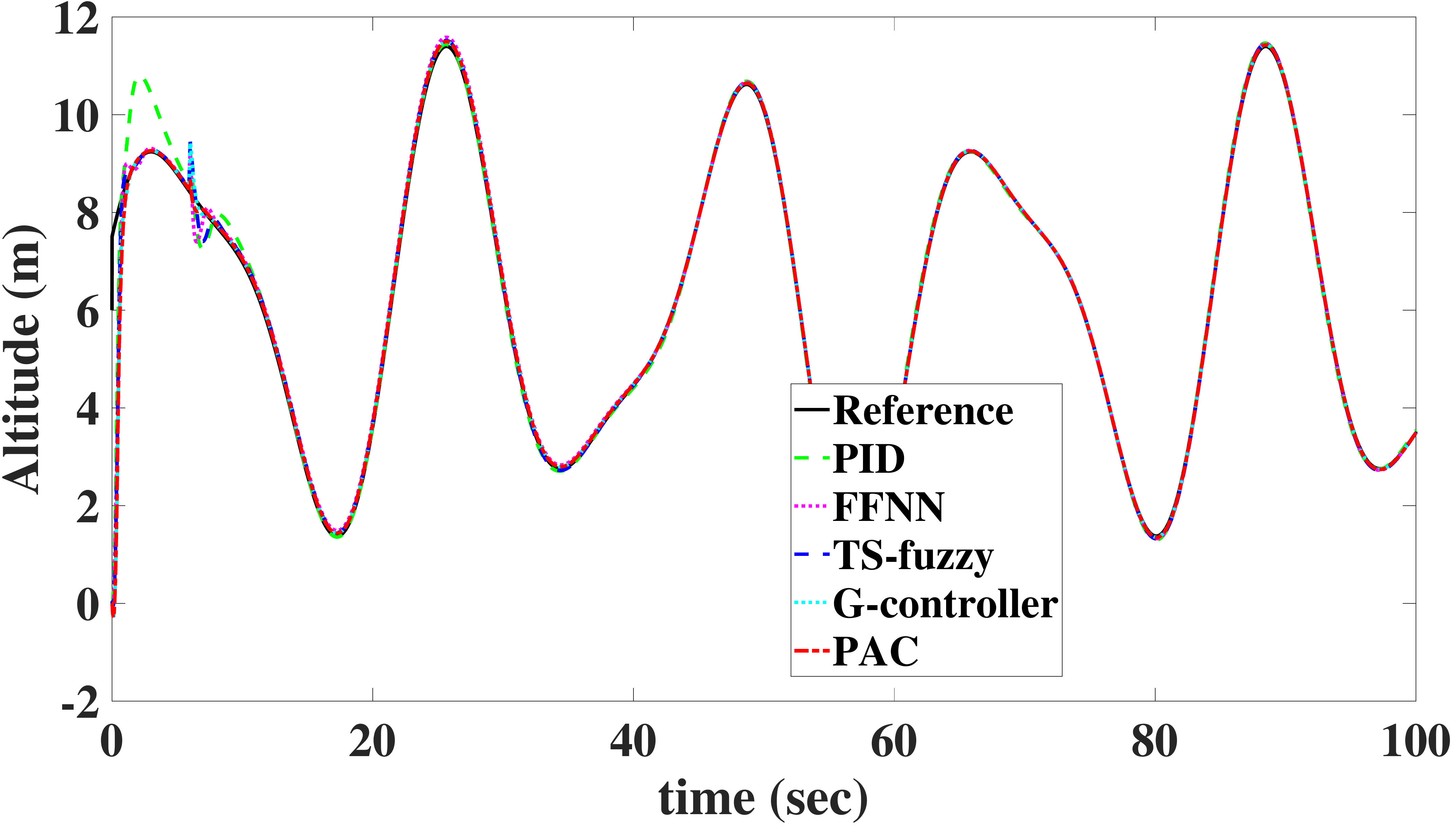}
				
			}
			\par\end{centering}
		\begin{centering}
			\subfloat[]{\includegraphics[scale=0.13]{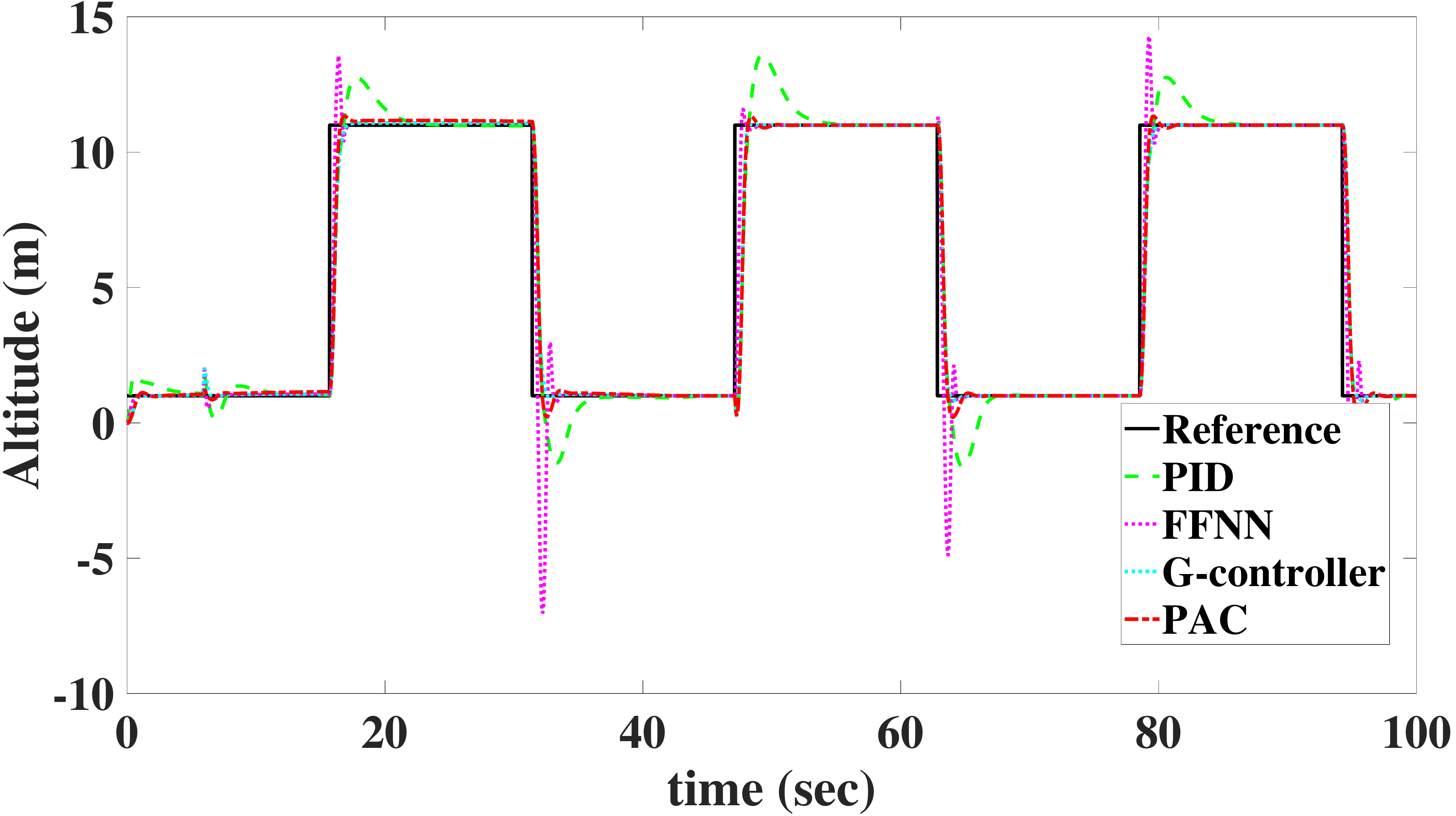}
				
			}\subfloat[]{\includegraphics[scale=0.13]{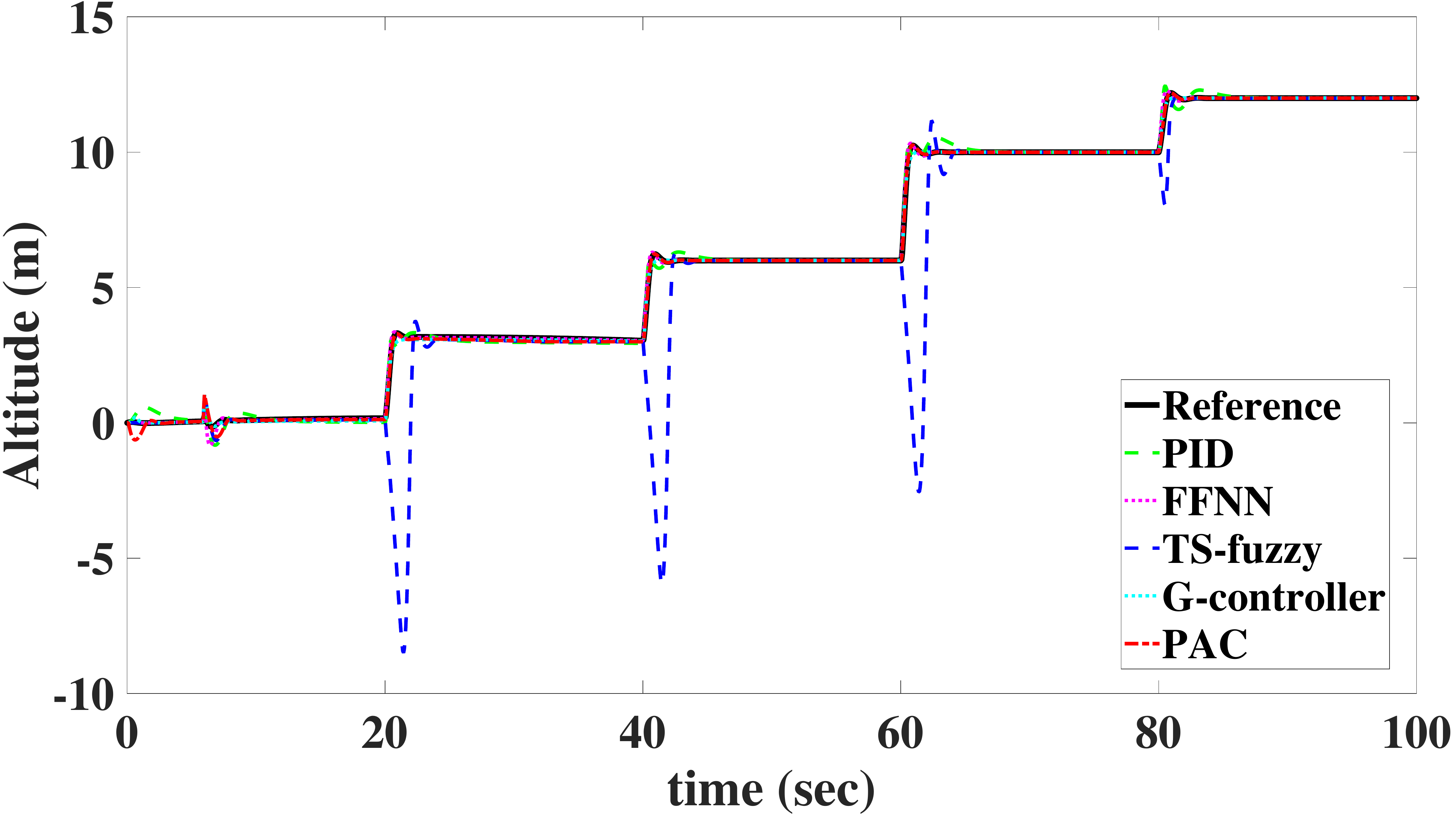}
				
			}
			\par\end{centering}
		\caption{Performance observation of different controllers in tracking altitude of BI-FWMAV by considering a sudden noise, and wind gust uncertainty, when the trajectories are (a) constant hovering, (b) variable heights with sharp edges, (c) variables height with smooth edges, (d) sum of sine function, (e) periodic square wave function, and (f) staircase function}\label{fig: difr_all_gust}
	\end{figure}
	\par\end{center}

\begin{center}
	\begin{figure}
		\begin{centering}
			\subfloat[]{\includegraphics[scale=0.13]{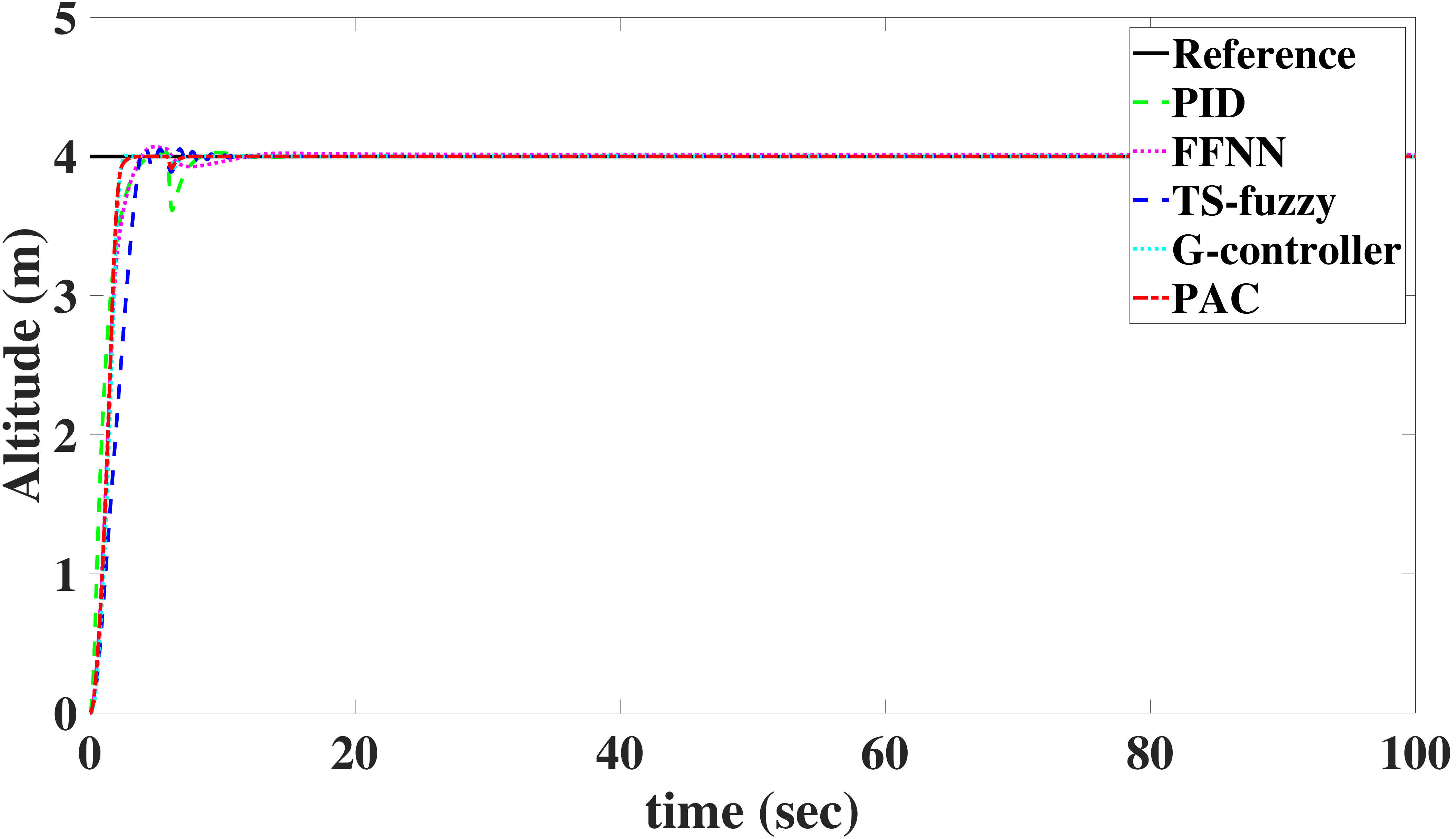}
				
			}\subfloat[]{\includegraphics[scale=0.13]{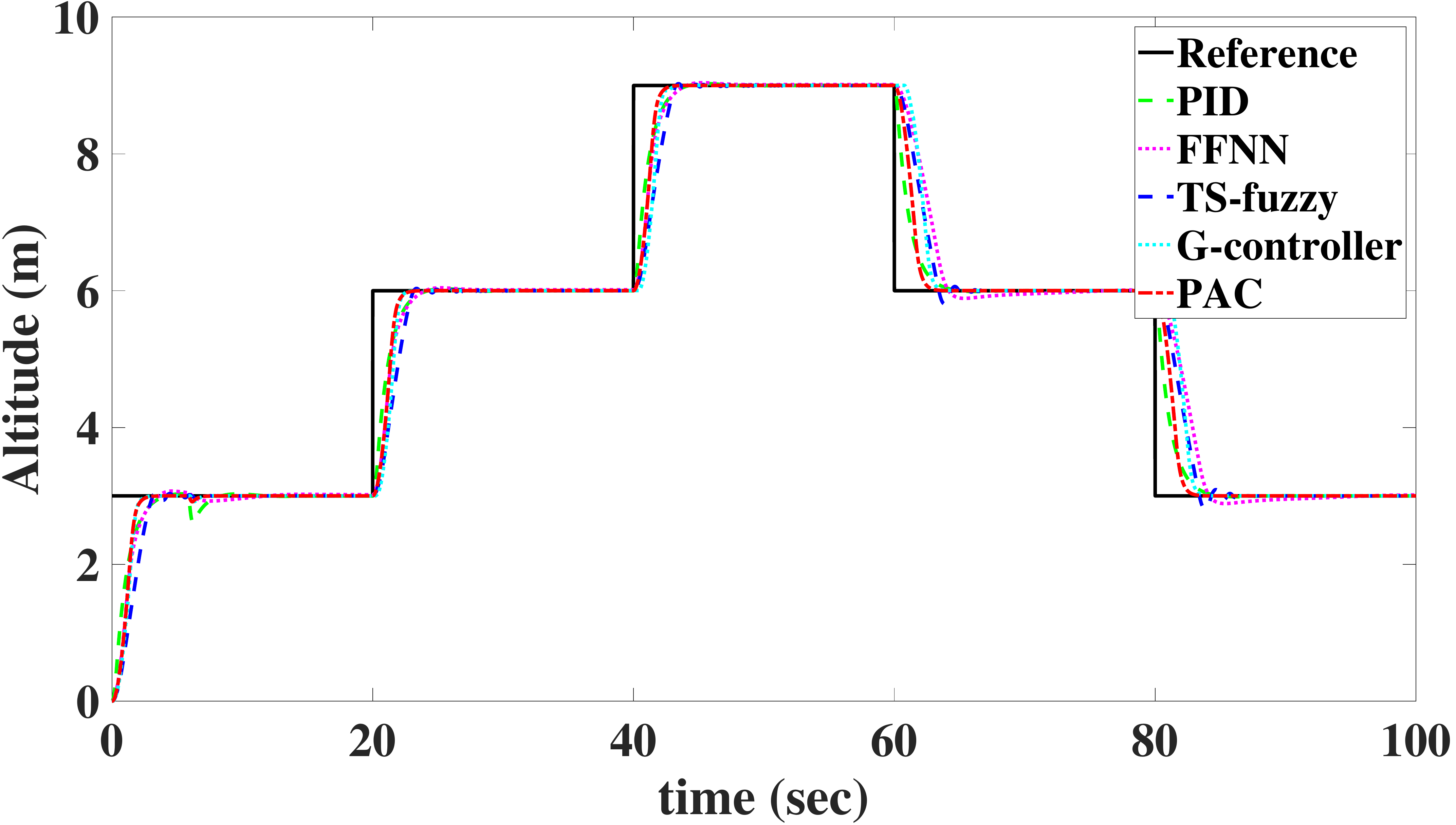}
				
			}
			\par\end{centering}
			\begin{centering}
			\subfloat[]{\includegraphics[scale=0.13]{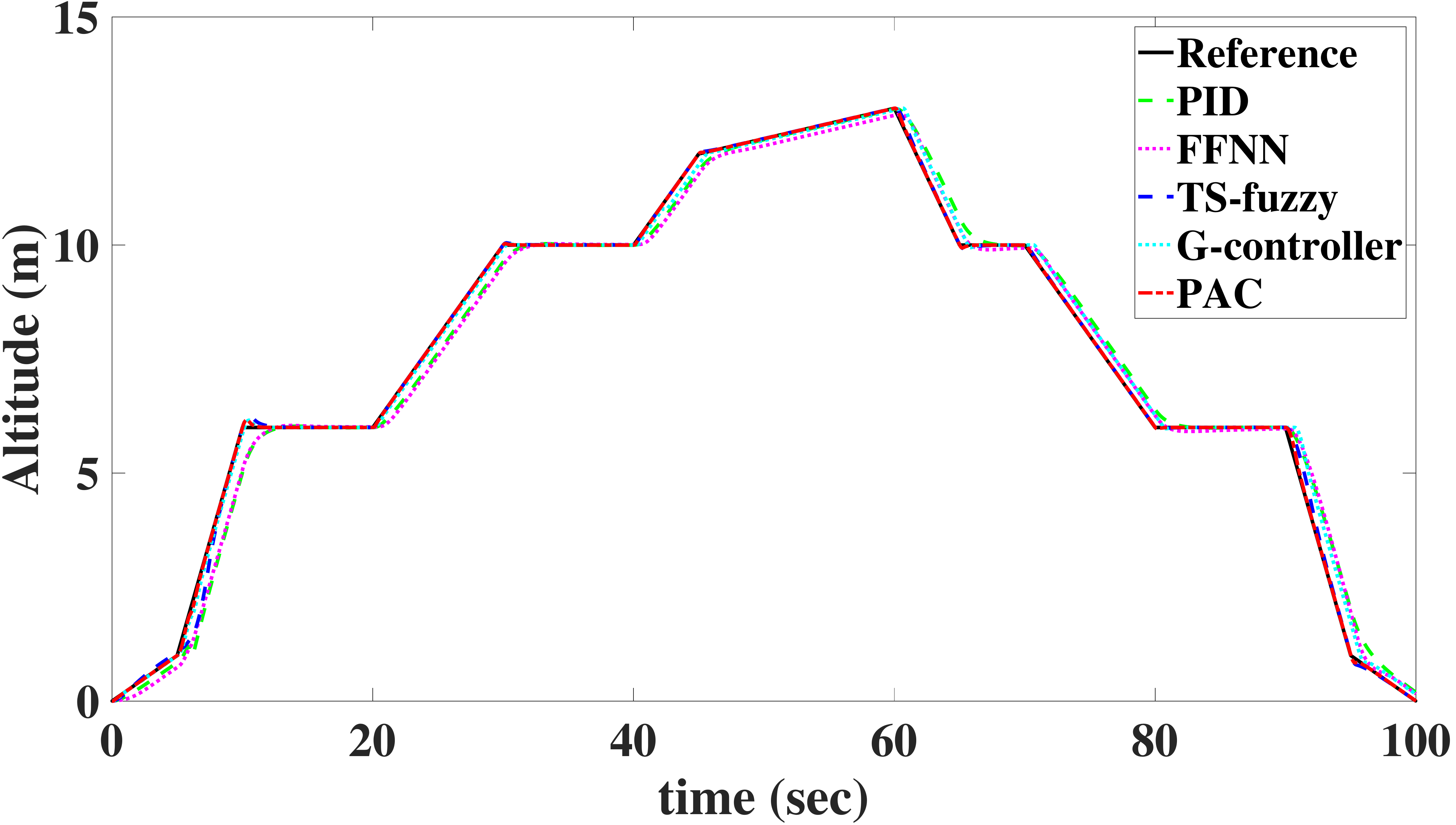}
				
			}\subfloat[]{\includegraphics[scale=0.13]{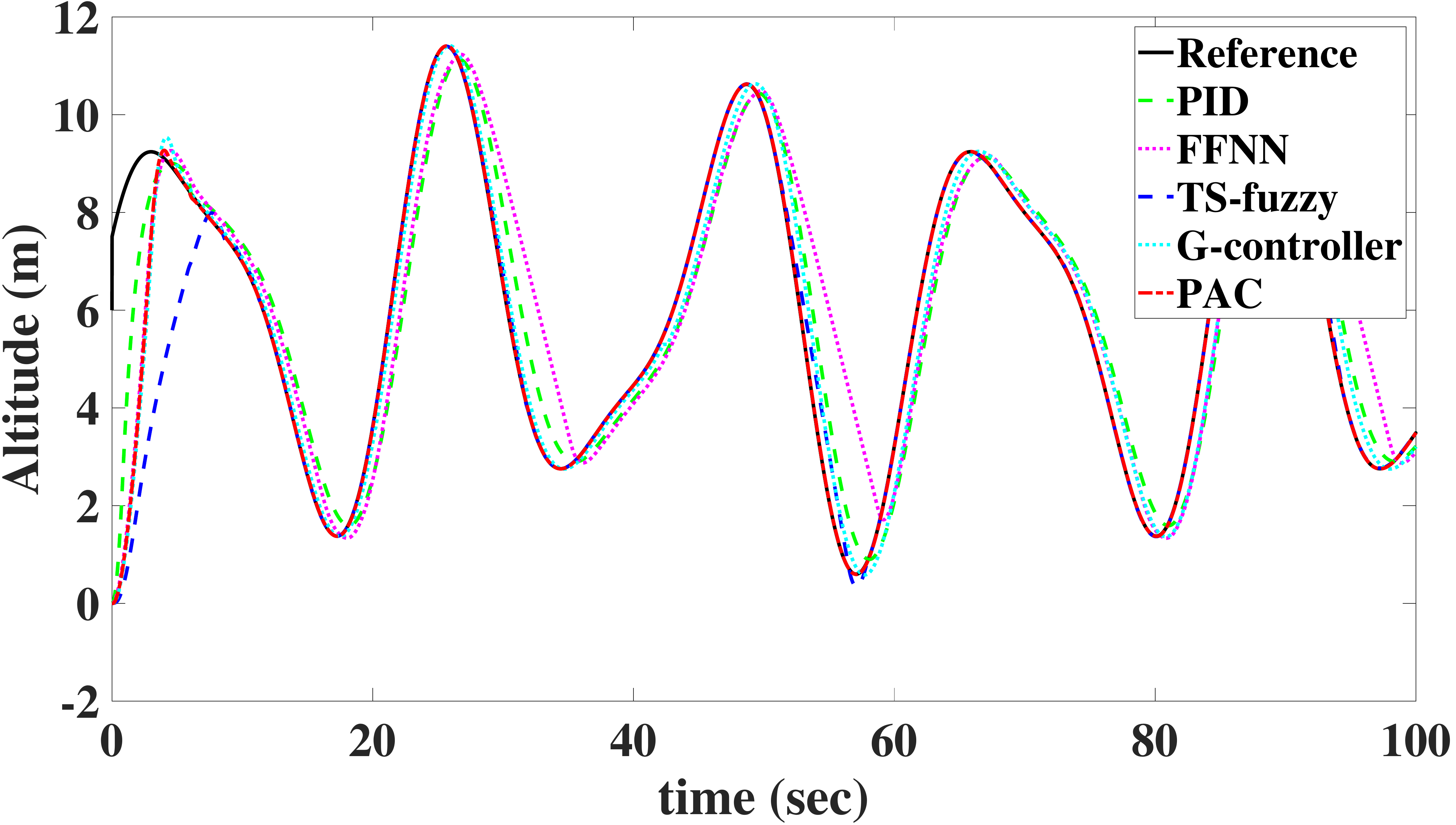}
				
			}
			\par\end{centering}
		\begin{centering}
			\subfloat[]{\includegraphics[scale=0.13]{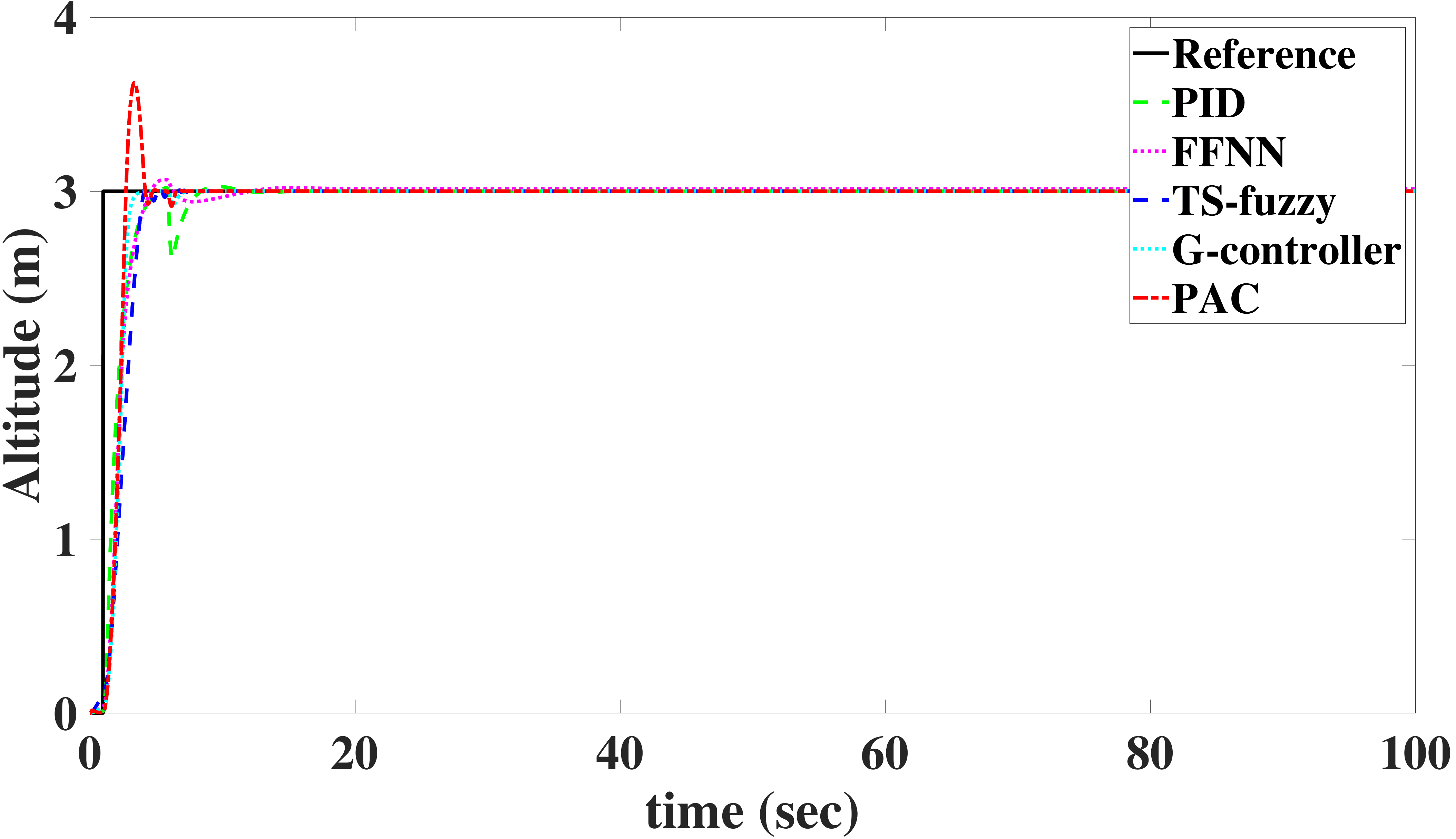}
				
			}\subfloat[]{\includegraphics[scale=0.13]{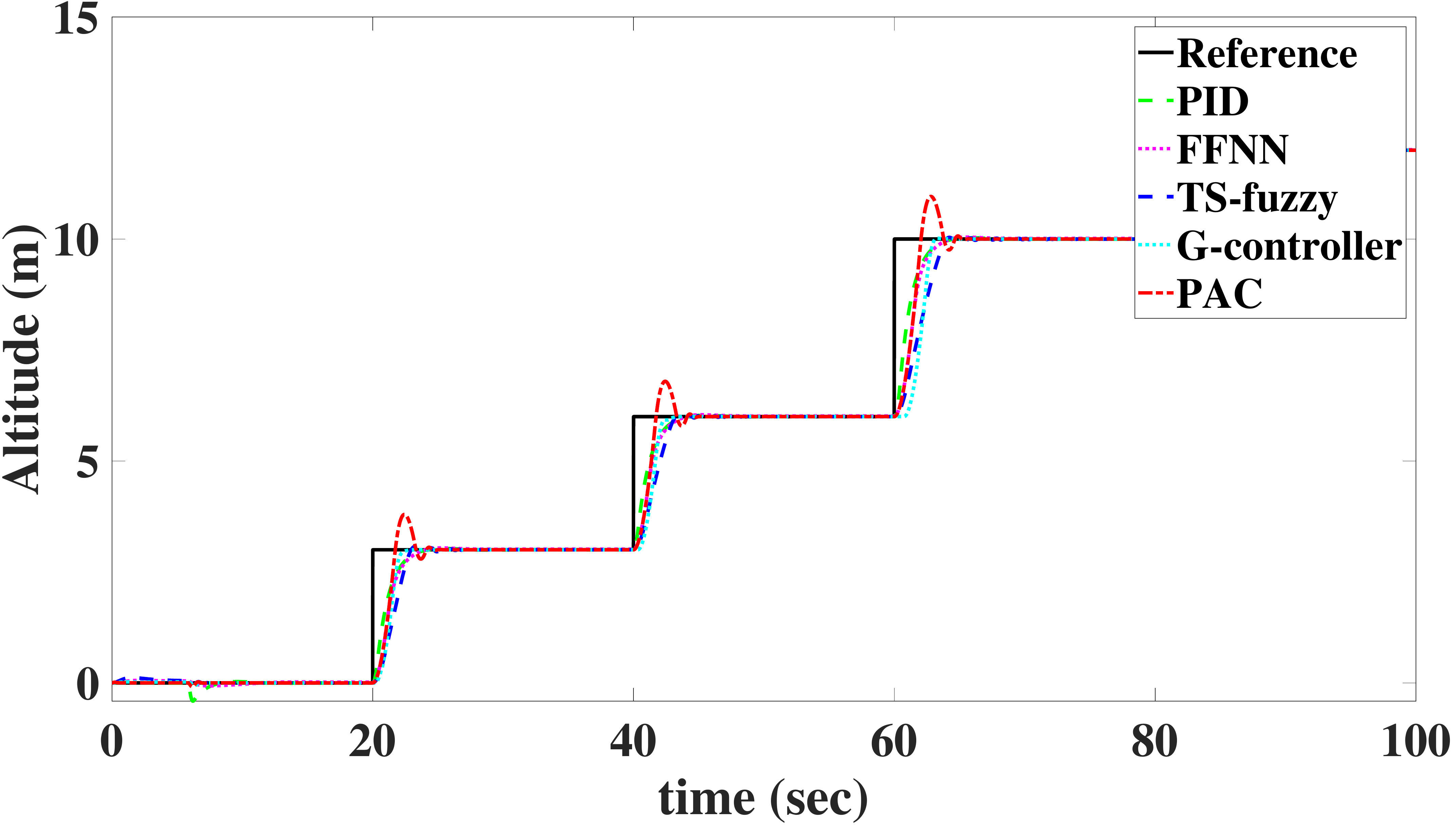}
				
			}
			\par\end{centering}
		\caption{Performance observation of different controllers in tracking altitude of hexacopter considering sudden noise, when the trajectories are (a) constant hovering, (b) variable heights with sharp edges, (c) variable heights with smooth edges, (d) sum of sines function, (e) step function, and (f) staircase function}\label{fig: hex_all_gust}
	\end{figure}
	\par\end{center}

\begin{center}
	\begin{table}
		\caption{Measured features of various controllers in operating the BI-FWMAV by considering a noise of sudden peak amplitude, and wind gust disturbance (RT: rise time, ST: settling time, CH: constant height, VH: variable height, SS: sum of sine, ms: millisecond, m: meter, MA: maximum amplitude, PSW: periodic square wave)}\label{tab:difr_all_noise}
		\centering{}%
		\begin{tabular}{>{\raggedright}p{2.5cm}>{\raggedright}p{1.8cm}>{\centering}p{1cm}>{\centering}p{1cm}>{\centering}p{1cm}>{\centering}p{1.1cm}>{\centering}p{1.1cm}}
			\hline 
			\multirow{2}{2.5cm}{\textbf{Desired trajectory}} & \multirow{2}{1.8cm}{\textbf{Measured features}} & \multicolumn{5}{c}{\textbf{Control method}}\tabularnewline
			\cline{3-7} 
			&  & \textbf{PID} & \textbf{FFNN} &\textbf{TS-fuzzy}& \textbf{G-control} & \textbf{{PAC}}\tabularnewline
			\hline 
			\multirow{4}{2.5cm}{CH (MA 10 m)} & RMSE & 0.6536 & 0.7268 & \textbf{0.5746} & 0.6657 & 0.6712\tabularnewline
			& RT (ms) & 50.772 & 55.828 & 44.629& \textbf{41.208} & 47.207\tabularnewline
			& ST (ms) & 1025.9 & 2707.2 & 743.39 & \textbf{635.53} & 747.43\tabularnewline
			& Peak (m) & 12.247 & 11.573 & 11.035 & 11.022 & 11.035\tabularnewline
			\hline 
			\multirow{4}{2.5cm}{VH with sharp change (MA 9 m)} & RMSE & 0.3430 & 0.4237 & 2.4603 & \textbf{0.3351} & 0.3611\tabularnewline
			& RT (ms) & 23.931 & 48.943 & 43.949 & 50.892 & \textbf{13.728}\tabularnewline
			& ST (ms) & 8176.2 & 8386.3 & 8329.4 & \textbf{8133.2} & 8166.5\tabularnewline
			& Peak (m) & 9.3731 & 9.6742 & 9.3010 & \textbf{9.0073} & 9.2270\tabularnewline
			\hline 
			\multirow{4}{2.5cm}{VH with smooth change (MA 13 m)} & RMSE & 0.1258 & 0.1541 & 0.0823 & \textbf{0.0613} & 0.0899\tabularnewline
			& RT (ms) & 8.8573 & 11.231 & 1.6537 & 4.1881 & \textbf{0.1314}\tabularnewline
			& ST (ms) & 9884.3 & \textbf{9857.5} & 9871.1 & 9870.5 & 9872.1\tabularnewline
			& Peak (m) & 13.007 & 13.009 & \textbf{13.004} & 13.019 & 13.006\tabularnewline
			\hline 
			\multirow{4}{2.5cm}{SS function (MA 11 m)} & RMSE & 0.4832 & 0.5565 & \textbf{0.4685} & 0.4998 & 0.5075\tabularnewline
			& RT (ms) & 21.109 & 31.092 & 20.616 & 18.675 & 19.459\tabularnewline
			& ST (ms) & 9960.1 & \textbf{9959.7} & 9960.1 & 9959.8 & 9960.1\tabularnewline
			& Peak (m) & \textbf{11.468} & 11.712 & 11.518 & 11.489 & 11.534\tabularnewline
			\hline 
			\multirow{4}{2.5cm}{PSW function (MA 11 m)} & RMSE & 2.7739 & 3.2660 & 421.39 & \textbf{2.5112} & 2.5122\tabularnewline
			& RT (ms) & 546.47 & 472.35 & 1040.5 & \textbf{57.067} & 59.508\tabularnewline
			& ST (ms) & 9923.9 & 9911.7 & 9972.2 & \textbf{9603.2} & 9634.1 \tabularnewline
			& Peak (m) & 12.794 & 12.667 & 736.93 & \textbf{12.067} & 12.073\tabularnewline
			\hline 
			\multirow{4}{2.5cm}{Staircase function (MA 12 m)} & RMSE & 0.3205 & 0.3960 & 2.1397 & \textbf{0.2916} & 0.3131\tabularnewline
			& RT (ms) & 5996.0 & 4067.3 & \textbf{4024.4} & 5999.4 & 6002.2\tabularnewline
			& ST (ms) & 8370.8 & 8156.2 & 8094.8 & 8056.5 & \textbf{8055.1}\tabularnewline
			& Peak (m) & 12.453 & 12.458 & 12.072 & \textbf{12.007} & 12.198\tabularnewline
			\hline 
		\end{tabular}
	\end{table}
	\par\end{center}

\begin{center}
	\begin{table}
		\caption{Measured features of various controllers in regulating the hexacopter by considering noise of sudden peak amplitude (RT: rise time, ST: settling time, CH: constant height, VH: variable height, ms: millisecond, m: meter, MA: maximum amplitude)}\label{tab:hex_all_noise}
		\centering{}%
		\begin{tabular}{>{\raggedright}m{2.5cm}>{\raggedright}p{1.8cm}>{\centering}p{1cm}>{\centering}p{1cm}>{\centering}p{1cm}>{\centering}p{1.1cm}>{\centering}p{1.1cm}}
			\hline 
			\multirow{2}{2.5cm}{\textbf{Desired trajectory}} & \multirow{2}{1.8cm}{\textbf{Measured features}} & \multicolumn{5}{c}{\textbf{Control method}}\tabularnewline
			\cline{3-7} 
			&  & \textbf{PID} & \textbf{FFNN} & \textbf{TS-fuzzy} & \textbf{G-control} & \textbf{PAC}\tabularnewline
			\hline 
			\multirow{4}{2.5cm}{CH (MA 4 m)} & RMSE & \textbf{0.3383} & 0.4067 & 0.4772 & 0.4237 & 0.4048\tabularnewline
			& RT (ms) & 208.28 & 197.14 & 259.03 & \textbf{141.66} & 144.67\tabularnewline
			& ST (ms) & 746.30 & 831.62 & 634.67 & \textbf{274.51} & 623.97\tabularnewline
			& Peak (m) & 4.0293 & 4.0704 & 4.0714 & 4.0909 & \textbf{4.0015}\tabularnewline
			\hline 
			\multirow{4}{2.5cm}{VH with sharp change (MA 9 m)} & RMSE & \textbf{0.5367} & 0.7586  & 0.7479 & 0.6509 & 0.6391\tabularnewline
			& RT (ms) & 203.76 & 195.22 & 209.19 & \textbf{122.12} & 125.84\tabularnewline
			& ST (ms) & 8315.9 & 8583.9 & 8411.8 & 8257.2 & \textbf{8232.0}\tabularnewline
			& Peak (m) & 9.0281 & 9.0406 & 9.0258 & 9.0022 & \textbf{9.0010}\tabularnewline
			\hline 
			\multirow{4}{2.5cm}{VH with smooth change (MA 13 m)} & RMSE & 0.3788 & 0.3666  & 0.1189 & 0.0322 & 0.0281\tabularnewline
			& RT (ms) & 110.21 & 142.33 & 6.7867 & 5.2776 & \textbf{2.4948}\tabularnewline
			& ST (ms) & 9879.5 & 9877.4 & \textbf{9869.2} & 9940.0 & 9871.0\tabularnewline
			& Peak (m) & 12.987 & 12.868 & 13.005 & 13.008 & \textbf{13.002}\tabularnewline
			\hline 
			\multirow{4}{2.5cm}{Step function (MA 3 m)} & RMSE & \textbf{0.2439} & 0.2795 & 0.3078 & 0.2841 & 0.2834\tabularnewline
			& RT (ms) & 203.94 & 197.23 & 215.13 & 121.83 & \textbf{112.14}\tabularnewline
			& ST (ms) & 763.38 & 912.24 & 396.34 & \textbf{300.48} & 631.1784\tabularnewline
			& Peak (m) & 3.0285 & 3.0676 & 3.0394 & \textbf{3.0042} & 3.6242\tabularnewline
			\hline 
			\multirow{4}{2.5cm}{Staircase function (MA 12 m)} & RMSE & \textbf{0.5074} & 0.6000 & 0.6961 & 0.5998 & 0.6078\tabularnewline
			& RT (ms) & 5993.9 & 5980.9 & 5973.7 & 6004.0 & \textbf{4144.9}\tabularnewline
			& ST (ms) & 8204.3 & 8222.4 & 8212.7 & \textbf{8183.5} & 8254.5\tabularnewline
			& Peak (m) & 12.028 & 12.039 & 12.021 & \textbf{11.998} & 12.587\tabularnewline
			\hline
			\multirow{4}{2.5cm}{Sum of sine function (MA 11 m)} & RMSE & 1.2123 & 1.7003 & 1.4325 & 1.0954 & \textbf{1.0787}\tabularnewline
			& RT (ms) & \textbf{59.638} & 114.03 & 181.41 & 125.94 & 122.53\tabularnewline
			& ST (ms) & 9945.9& 9944.2 & \textbf{9959.2} & 10004 & 9959.7\tabularnewline
			& Peak (m) & \textbf{11.130} & 11.235 & 11.400 & 11.413 & 11.409\tabularnewline
			\hline
			\multirow{4}{2.5cm}{Pitching} & RMSE & 0.3513 & 0.0451 & N/A & 0.0466 & \textbf{0.0109}\tabularnewline
			& RT (ms) & 14.907 & 14.686 & N/A & 65.469 & \textbf{10.135}\tabularnewline
			& ST (ms) & 10057 & 10057 & N/A &  \textbf{9982.9}& 10053\tabularnewline
			& Peak (rad) & 0.5615 & 0.5758 & N/A & \textbf{0.5398} & 0.5469\tabularnewline
			\hline 
			\multirow{4}{2.5cm}{Rolling} & RMSE & 0.1673 & N/A & N/A & 0.0290 & \textbf{0.0259}\tabularnewline
			& RT (ms) & \centering{}166.116 & N/A & N/A & 118.75 & \textbf{91.596}\tabularnewline
			& ST (ms) & 10037 & N/A & N/A & \textbf{9978.9} & 9979.9\tabularnewline
			& Peak (rad) & 0.3907 & N/A & N/A & 0.3513 & 0.4852\tabularnewline
			\hline 
		\end{tabular}
	\end{table}
	\par\end{center}

\subsection{Self-adaptive mechanism of PAC}
Based on the bias-variance concept explained in section \ref{sec:Structure of PALM_C}, rules of the PAC have been evolved dynamically in different experiments. Before analyzing the evolution of the structure of our proposed controller, we have tried to summarize shortfalls of the benchmark controllers used in this work. The linear PID controller's realization is based upon three gain parameters, namely proportional, integral, and differential gain. They are typically denoted as $K_{p}$, $K_{i}$, and $K_{D}$. It requires to set values for those parameters in offline before utilizing in control operation, which may oblige repetitious efforts. Besides, those parameters can not be tuned online. Before performing the control operation, both the adaptive FFNN and TS-fuzzy controllers require offline training encouraged by the PID controller's input-output datasets. Though they adapt their network parameters during operation, they have a fixed structure with a hidden layer consists of ten fixed nodes and five rules. To attain more robustness against disturbances, the benchmark G-controller can adapt both the network and parameters. Nonetheless, it needs to deal with lots of free parameters in both antecedent and consequent parts. Besides, the evolution of G-controller is regulated by some user-defined thresholds. Sometimes, it is causing a large settling time. On the contrary, our proposed evolving controller has no premise parameters. The only parameter that needs to be adapted is weight, which is adapted here using SMC theory. Unlike the benchmark evolving controllers, PAC is free from predefined problem-dependent parameter for regulating its structure. 

The structure evolution in terms of added or pruned rules for some trajectories of BI-FWMAV and hexacopter was disclosed graphically in Fig. \ref{fig: difr_all} (c) and (d), Fig. \ref{fig: hex_all} (c), and in Fig. \ref{fig: hex_roll_pitch} to get a vivid insight into the evolution of rules in PAC. For further clarification, the fuzzy rule extracted by PAC in controlling the BI-FWMAV can be uttered as follows:
	\begin{align}\label{eq:T1rule_eg}
	R^{1}: \text{IF}~X_n~\text{is close to}~\bigg([1,e,\dot{e},y_r]\times[0.0121,0.0909,0.4291,0.6632]^{T}\bigg),\\\nonumber~\text{THEN}~y=0.0121+0.0909e+0.4291\dot{e}+0.6632 y_r
	\end{align}
where $e$ is the error, i.e., the difference between the reference and actual output of the plant, $\dot{e}$ is the error derivative, i.e., the difference between the present and previous state error value, $y_r$ is the reference for the plant to be controlled. Since PAC is targeted to minimize the tracking error to zero or very close to zero, it needs information about the error as an input to the closed-loop system. It is also witnessed in PAC's rule as exposed in \eqref{eq:T1rule_eg}. When PAC was controlling the BI-FWMAV in tracking a constant altitude of 10 m, it generated 3 rules within 1 second at the beginning of control operation. Since the reference was unaltered and stability of the plant was achieved, PAC did not add or prune any extra rule later on as witnessed from the Fig. \ref{fig: difr_all} (c). While BI-FWMAV was following a variable height trajectory with sharp changes at edges, the PAC started operating by producing only one rule. After 8 seconds it added two more rules and achieved system stability. After that, the changes in trajectories were handled by PAC through the tuning of weights only. It does not need any further structure evolution as observed in Fig. \ref{fig: difr_all} (d). The successful evolution of rules by confirming system stability was also achieved by PAC in controlling the hexacopter plant. For instance, PAC supported hexacopter to track a constant altitude of 4 m with two rules as displayed in Fig. \ref{fig: hex_all} (c). While PAC was regulating the rolling of hexacopter for a sum of sine trajectory, it started operating with only one rule. Immediately after 17 seconds, it added another two rules, however, one of them was pruned at 19 seconds to minimize the overfitting phenomenon, while maintained system stability with two rules only. With these two rules, it tracked the trajectory efficiently through the SMC based weight adaptation. A similar scenario was witnessed in controlling the pitching position of the hexacopter since the same trajectory was employed. In sum up, by adapting both the structure and weights, the PAC was controlling the MAVs effectively to follow the desired trajectories very closely and by preserving the stability of the closed-loop system.

\subsection{PAC's performance analysis using the Wilcoxon signed-rank test}
As a simple statistical test, Wilcoxon signed-rank tests had been performed among various controllers. In these tests, the residual error among the reference trajectories and plant's actual trajectories are calculated for all the utilized controllers. To do the Wilcoxon signed-rank test, the Matlab function, namely $\textit{signrank}$ has been utilized as follows:
\begin{equation}\label{eq:wilcox}
    [p,h] = signrank(A,B)
\end{equation}

where $A$ is indicating the dataset of residual error from our proposed PAC, and $B$ is denoting the dataset of residual error from the benchmark controllers. The test expressed in Eq. \eqref{eq:wilcox} is returning a logical value that indicates the test decision. Theoretically, $h=1$ is expressing a rejection of the null hypothesis, whereas $h=0$ is pointing the opposite at the 5\% significance level. In other words, when $A$ differs statistically from $B$, then $h=1$. When $h=0$, $A$ and $B$ are statistically similar to each other.

In this work, from the Wilcoxon signed-rank test between the PAC (considered as $A$) and a benchmark controller (considered as $B$), we have observed the value of $h$. It gives us better insight regarding our proposed controller's performance, which is explained in this paragraph. We already have recorded the RMSE of all the controllers for various trajectory tracking performance. When the RMSE of the PAC is lower the benchmark one, it is indicating the superiority of PAC over benchmark controllers. It is supported by the logical value of $h=1$ since the residual error dataset of PAC is statistically different and lower than the benchmark controllers. On the other hand, when the RMSE is lower in a benchmark controller than the PAC, the logical value of $h$ is 0. It is indicating a comparable performance of the PAC with regards to the benchmark ones since both of their residual error dataset are very similar. To get a clearer view, the constant height trajectory tracking performance with the Wilcoxon signed-rank test for both the plants are tabulated in \tablename{~\ref{tab:stat_1}}. 

The observation from \tablename{~\ref{tab:stat_1}} can be summarized in three different cases as follows: 1) In case of superior performance from the PAC, i.e., when the RMSE is lower in PAC than the benchmark one, the logical value of $h$ is 1. It indicates that the residual error set of PAC is different from the benchmark controller. Therefore, different residual error set with lower RMSE is clearly indicating the improved performance of the PAC; 2) When the performance of the benchmark controllers are better than the PAC by having a slightly lower RMSE, the value of $h$ is 0. It is indicating similar residual error sets from both the benchmark controller and PAC. Thus, the performance of PAC is comparable with the benchmark one; 3) In case, the performance of PAC is slightly better than the benchmark controllers. However, the value of $h$ is 0, which is stating a comparable performance from the benchmark controllers. The Wilcoxon signed-rank test results for all other trajectories are attached in the supplementary document.          

\begin{center}
\begin{table}
\caption{Wilcoxon signed rank test between controllers in regulating the altitude of MAVs}\label{tab:stat_1}
\begin{centering}
\begin{tabular}{|c|c|c|c|c|}
\hline 
\textbf{Plant} & \textbf{Desired trajectory} & \textbf{Controllers} & \textbf{RMSE} & \textbf{h}\tabularnewline
\hline 
\multirow{8}{*}{BI-FWMAV} & \multirow{8}{*}{CH (MA 10 m)} & PID & 0.6460 & \multirow{2}{*}{0}\tabularnewline
\cline{3-4} 
 &  & PAC & 0.6668 & \tabularnewline
\cline{3-5} 
 &  & TS-Fuzzy & 0.6693 & \multirow{2}{*}{1}\tabularnewline
\cline{3-4} 
 &  & PAC & 0.6668 & \tabularnewline
\cline{3-5} 
 &  & FFNN & 0.7108 & \multirow{2}{*}{1}\tabularnewline
\cline{3-4} 
 &  & PAC & 0.6668 & \tabularnewline
\cline{3-5} 
 &  & G-controller & 0.6631 & \multirow{2}{*}{{0}}\tabularnewline
\cline{3-4} 
 &  & PAC & 0.6668 & \tabularnewline
\hline 
\multirow{8}{*}{Hexacopter} & \multirow{8}{*}{CH (MA 4 m)} & PID & 0.3551 & \multirow{2}{*}{0}\tabularnewline
\cline{3-4} 
 &  & PAC & 0.4204 & \tabularnewline
\cline{3-5} 
 &  & TS-Fuzzy & 0.4771 & \multirow{2}{*}{1}\tabularnewline
\cline{3-4} 
 &  & PAC & 0.4204 & \tabularnewline
\cline{3-5} 
 &  & FFNN & 0.4221 & \multirow{2}{*}{0}\tabularnewline
\cline{3-4} 
 &  & PAC & 0.4204 & \tabularnewline
\cline{3-5} 
 &  & G-controller & 0.4239 & \multirow{2}{*}{0}\tabularnewline
\cline{3-4} 
 &  & PAC & 0.4204 & \tabularnewline
\hline 
\end{tabular}
\par\end{centering}
\end{table}
\par\end{center}

\subsection{Comparison against closely related work}
To revile the strength of the proposed controller, namely PAC against its closely related recent variants, a comparison among them is addressed in this subsection. The mechanism of growing and pruning rules and control laws of these controllers are compared here. To get a clearer overview, they are tabulated in \tablename{~\ref{tab:Comparison against closely}}.
\begin{table}
\caption{Comparison against closely related work}\label{tab:Comparison against closely}
\centering{}%
\begin{tabular}{>{}p{1.6cm}>{}p{2.9cm}>{}p{2.9cm}>{}p{2.9cm}}
\hline 
\textbf{Algorithm} & \textbf{Rule-growing method} & \textbf{Rule-pruning method} & \textbf{Clustering technique and control laws}\tabularnewline
\hline 
G-controller \cite{ferdaus2019generic} &  The datum significance method & The extended rule significance method & Multivariate Gaussian membership function is utilized in the antecedent part, which creates hyper-ellipsoidal clusters. SMC-based adaptation laws are used.\tabularnewline
\hline 
eTS controller \cite{angelov2004fuzzy} & Spatial proximity between a particular data point and all other points, namely potential is used to add rules. & No rule pruning mechanism. & The Gaussian membership function is used in the antecedent part, which creates hyper-spherical clusters. Recursive least-square-based adaptation laws are used.\tabularnewline
\hline 
Adaptive fuzzy neural controller \cite{gao2003online} & $\epsilon-$completeness and system error are measured to add rules. & Error reduction ratio (ERR) is utilized to prune a rule. & The Gaussian membership function is used in the antecedent part, which creates hyper-spherical clusters. Tracking error-based parameter adaptation laws are employed. \tabularnewline
\hline 
Interval type-2 fuzzy brain emotional learning control (T2FBELC) \cite{le2018self}& Maximum membership grade of input is measured to add rules & Minimum membership grade is calculated to delete rules. &The Gaussian membership function is used in the antecedent part, which creates hyper-spherical clusters. The steepest descent gradient approach is used to derive adaptation laws. \tabularnewline
\hline 
\end{tabular}
\end{table}

In closely-related evolving controllers, mostly univariate or multivariate Gaussian membership functions are employed in the antecedent part, where they need to tune the antecedent parameters like centers and widths. Unlike them, hyperplane-based clustering technique is used in PAC, where it does not have any antecedent parameters. Therefore, PAC is free from the computation of tuning of such parameters. Besides, all the above-mentioned controllers have a dependency on predefined thresholds to grow or prune rules. To get rid of such dependence on predefined values, in our PAC, the concept of bias and variance has been incorporated.

\section{Conclusion\label{sec:Conclusion}}
Based on the research gap in controlling MAVs in cluttered environments, a completely model-free evolving controller, namely PAC is proposed in this work. A bottleneck of the existing evolving controllers is the utilization of numerous free parameters and their online adaptation. Such inadequacy has been mitigated in our PAC since it has no premise parameters. The only parameter used in our evolving controller to acquire the desired tracking is the weight. For instance, in our experiment, the benchmark evolving G-controller with three rules requires 48 network parameters, whereas PAC needs only 12 parameters with three rules. Apart from that, conventional evolving controllers adhere to user-defined problem-based thresholds to shape their structure. In PAC, rather than predefined parameters, the bias-variance concept based network significance method is utilized to determine it's structure. The PAC has been verified by implementing them in various MAV plants namely BI-FWMAV and hexacopter to track diverse trajectories. All the achievements are contrasted with a commonly utilized PID controller, an adaptive nonlinear FFNN controller, a TS-fuzzy controller, an evolving controller namely G-controller. Furthermore, the controller's robustness against uncertainties and disruptions is ascertained by injecting a wind gust and sudden peak to the MAVs dynamics. In controlling both plants with uncertainties, lower or comparable overshoot and settling time were observed from PAC with a simplified evolving structure, which is testifying its robustness against uncertainties and compatibility in regulating MAVs. Inspired by PAC's efficient performance in controlling the simulated BI-FWMAV and hexacopter plant, it will be employed in MAVs hardware in the future. Besides, we will be continuing to develop evolving fuzzy controllers with fewer network parameters by maintaining satisfactory performance.

\section*{Acknowledgment}
The authors would like to thank the unmanned aerial vehicle laboratory of the UNSW Canberra for supporting with the BI-FWMAV and hexacopter plants, and the computational support from the Computational Intelligence Laboratory of Nanyang Technological University (NTU) Singapore. The 5th author acknowledges the support by the COMET-K2 Center of the Linz Center of Mechatronics (LCM) funded by the Austrian federal government and the federal state of Upper Austria. This work was financially supported by the NTU start-up grant and MOE Tier-1 grant.

\bibliographystyle{unsrt}  
\bibliography{references}

\end{document}